\documentclass[10pt,twocolumn,letterpaper]{article}

\usepackage{iccv}              

\definecolor{iccvblue}{rgb}{0.21,0.49,0.74}
\usepackage[pagebackref,breaklinks,colorlinks,allcolors=iccvblue]{hyperref}

\def\paperID{10116} 
\def\confName{ICCV}
\def\confYear{2025}

\title{Automated Model Evaluation for Object Detection\\via Prediction Consistency and Reliability}

\author{Seungju Yoo$^{1}$\qquad Hyuk Kwon$^{1}$\qquad Joong-Won Hwang$^{2}$\qquad Kibok Lee$^{1}$\\
${}^{1}$Yonsei University\qquad ${}^{2}$ETRI\\
${}^{1}${\tt\small \{seungju\_yoo, kh12043, kibok\}@yonsei.ac.kr}\qquad ${}^{2}${\tt\small jwhwang@etri.re.kr}
}

\usepackage[absolute,overlay]{textpos}
\usepackage{multirow}
\usepackage{makecell}
\usepackage{pifont}
\usepackage{stfloats}
\usepackage{caption}
\usepackage{wrapfig}
\usepackage[accsupp]{axessibility}

\begin{document}
\maketitle
\begin{abstract}
Recent advances in computer vision have made training object detectors more efficient and effective; however, assessing their performance in real-world applications still relies on costly manual annotation. To address this limitation, we develop an automated model evaluation (AutoEval) framework for object detection. We propose Prediction Consistency and Reliability (PCR), which leverages the multiple candidate bounding boxes that conventional detectors generate before non-maximum suppression (NMS). PCR estimates detection performance without ground-truth labels by jointly measuring 1) the spatial consistency between boxes before and after NMS, and 2) the reliability of the retained boxes via the confidence scores of overlapping boxes. For a more realistic and scalable evaluation, we construct a meta-dataset by applying image corruptions of varying severity. Experimental results demonstrate that PCR yields more accurate performance estimates than existing AutoEval methods, and the proposed meta-dataset covers a wider range of detection performance. The code is available at \url{https://github.com/YonseiML/autoeval-det}.
\end{abstract}

\section{Introduction}
\label{sec:intro}

\begin{figure}[t]
    \centering
    \includegraphics[width=\columnwidth]{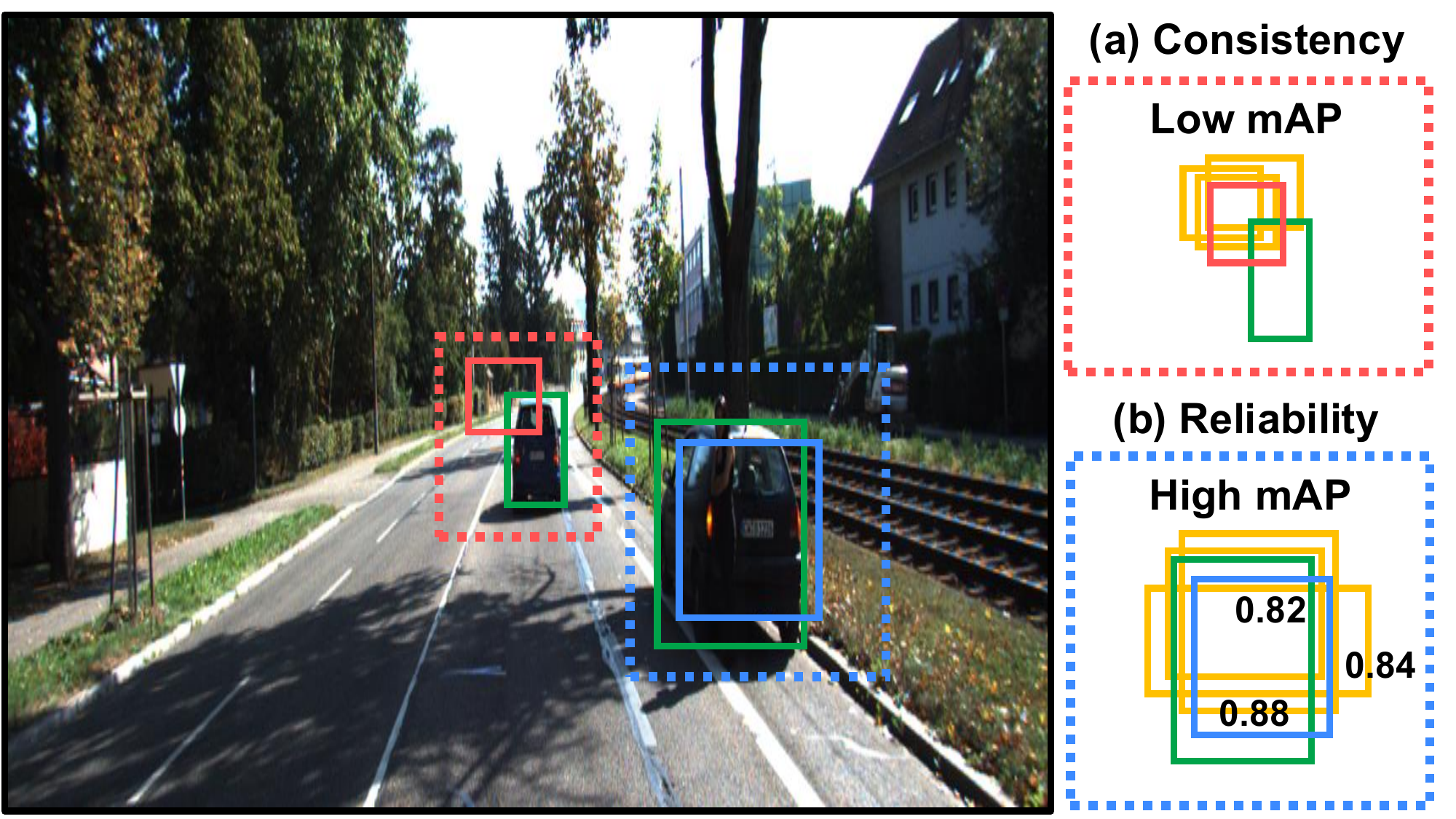}
    \caption{
    \textbf{Visual example of PCR.}
    \textcolor[HTML]{00A44A}{Green boxes} represent the ground-truth bounding boxes, and \textcolor[HTML]{FF5353}{red} and \textcolor[HTML]{3B8AFF}{blue} boxes denote an incorrect detection with low confidence and a correct detection with high confidence, respectively. \textcolor[HTML]{FFC000}{Orange boxes} show the pre-NMS candidate boxes, where the overlaid numbers indicate their confidence scores.
    \textbf{(a) Consistency.} The \textcolor[HTML]{FF5353}{red box} overlaps many \textcolor[HTML]{FFC000}{pre-NMS boxes}, yielding high consistency. Our consistency score measures spatial consistency with a merged pre-NMS box, motivated by the observation that \textbf{boxes with low confidence and high consistency} correlate with lower mAP.
    \textbf{(b) Reliability.} The \textcolor[HTML]{3B8AFF}{blue box} overlaps many \textcolor[HTML]{FFC000}{pre-NMS boxes} with high confidence scores, yielding high reliability. Our reliability score measures the proportion of overlapping pre-NMS boxes with high confidence scores, motivated by the observation that \textbf{boxes with high confidence and high reliability} correlate with higher mAP.
    }
    \vspace{2pt}
    \label{fig:Idea}
\end{figure}

Effective evaluation of machine learning models is essential before deployment, particularly when the target domain differs from the source domain because of environmental shifts or discrepancies in data distribution. However, assessing model performance in a new environment is challenging, as annotating test data is often costly and time-consuming. Automated model evaluation (AutoEval)~\cite{deng2021autoeval} addresses this issue by estimating performance on unlabeled test datasets.

Research on AutoEval has primarily focused on image classification, which serves as a foundation for the quick assessment of new methods~\cite{saito2019semi, guillory2021predicting, deng2021autoeval, arg2022leveragingunlabeleddatapredict}. In image classification, the limitations of evaluations on standard benchmarks often stem from domain discrepancies at the feature extraction stage, referred to as covariate shift, where features learned during training become less effective when deployed in new environments. Recent studies have attempted to address this by measuring distances between the feature distributions of the source and target datasets~\cite{deng2021autoeval, guillory2021predicting}. Although covariate shift is an important factor, we argue that it is insufficient for AutoEval in other computer vision tasks such as object detection, which introduce additional challenges requiring more advanced evaluation strategies.

Unlike image classification, object detection performance is sensitive to various factors such as variations in object scales, occlusion, background clutter, and object interactions; covariate shift cannot capture these intricate spatial relationships. Hence, we argue that estimating covariate shift alone cannot address the full spectrum of challenges in object detection. To address these challenges, we adopt a bottom-up strategy by 1)~collecting several measures from model outputs, 2)~analyzing their correlations with performance variations, and 3)~developing a method based on these insights. Specifically, we focus on the observation that conventional object detection models first generate many candidate bounding boxes, and then retain high-confidence boxes while discarding overlapping low-confidence boxes through non-maximum suppression (NMS)~\cite{ren2015fasterrcnnrealtimeobject, lin2017focallossdenseobject, tian2019fcos, he2017mask, redmon2016you}. We examine relationships between pre- and post-NMS boxes by measuring their geometric proximity and confidence scores, capturing both localization and classification aspects.

Building on these observations, we propose \textbf{Prediction Consistency and Reliability (PCR)} as an effective AutoEval method for object detection. \Cref{fig:Idea} illustrates the concepts of consistency and reliability in PCR. For low-confidence boxes, the high spatial consistency between pre- and post-NMS boxes indicates repeated mislocalization, suggesting that the model has misidentified objects. Because the associated pre-NMS boxes have even lower confidence than nearby post-NMS boxes, such consistent localization of low-confidence boxes is prone to incorrect detection results. In contrast, for high-confidence boxes, the reliability of post-NMS boxes is assessed based on the confidence scores of nearby pre-NMS boxes, capturing both localization and classification accuracy. Pre-NMS boxes with high confidence imply that the model is confident in both aspects, thereby promoting correct detection results.

Furthermore, we construct a meta-dataset to facilitate a more realistic and scalable evaluation of AutoEval methods for object detection. A recently proposed meta-dataset~\cite{yang2024bos} relies on data augmentation techniques, such as sharpness adjustment, equalization, color temperature shifts, solarization, autocontrast, brightness modification, and rotation; however, we argue that such strong augmentations may not adequately reflect the environmental shifts or discrepancies in data distribution observable in real-world applications, and that such a meta-dataset may fail to capture the full range of detection performance in practice. Instead, we apply image corruptions of varying severity, adopting the transformations used to generate ImageNet-C~\cite{hendrycks2019benchmarking}, which are specifically designed to simulate real-world corruptions. The proposed meta-dataset is \textit{realistic} in that it employs real-world corruptions rather than artificial augmentations, and \textit{scalable} in that varying severity levels enable evaluation across a broader spectrum of detection performance.

We summarize our contributions as follows:
\begin{itemize}
    \item \textbf{Prediction Consistency and Reliability (PCR).} We propose PCR, an effective AutoEval method for object detection that leverages the spatial alignment and confidence scores of the bounding boxes before and after NMS.
    \item \textbf{Corruption-based meta-dataset.} We construct a meta-dataset using image corruptions of varying severity levels, yielding a realistic and scalable benchmark that enables AutoEval for object detection.
    \item \textbf{Extensive empirical validation.} PCR consistently outperforms state-of-the-art AutoEval methods across various object detection models, including RetinaNet~\cite{lin2017focallossdenseobject} and Faster R-CNN~\cite{ren2015fasterrcnnrealtimeobject} paired with ResNet-50~\cite{he2016deep} and Swin Transformer~\cite{liu2021swin} backbones, across both augmentation- and corruption-based meta-datasets.
\end{itemize}

\section{Related Work}
\label{sec:related}

\noindent\textbf{Object Detection}
aims to localize and classify objects within an image by simultaneously predicting bounding boxes and class labels. Object detectors are typically categorized into 1) two-stage methods, such as R-CNN variants~\cite{girshick2014richfeaturehierarchiesaccurate, girshick2015fastrcnn, ren2015fasterrcnnrealtimeobject}, which generate and refine region proposals, and 2) one-stage methods, such as YOLO~\cite{Redmon_2016_CVPR} and RetinaNet~\cite{lin2017focallossdenseobject}, which perform detection in a single pass. While these models are commonly evaluated on standard benchmark datasets including COCO~\cite{lin2015microsoftcococommonobjects} and Pascal VOC~\cite{Everingham10}, distribution shifts in real-world scenarios, such as varying lighting conditions, often lead to discrepancies between benchmark performance and actual deployment performance. Moreover, acquiring high-quality annotations in real-world scenarios is often costly and time-consuming. To address these challenges, we explore a framework for estimating detector performance on unlabeled test datasets without relying on manual annotations.

\vspace{4pt}
\noindent\textbf{Automated Model Evaluation (AutoEval)}
aims to estimate model performance on unlabeled test datasets~\cite{deng2021autoeval}. In the absence of ground-truth labels, AutoEval measures scores that correlate with performance, which are then used to predict performance via regression.
Early studies utilized confidence-based measures, such as maximum softmax probabilities~\cite{hendrycks2018ps, guillory2021predicting, arg2022leveragingunlabeleddatapredict} and predictive entropy~\cite{saito2019semi}, demonstrating that simple scalar indicators derived from model outputs can effectively estimate classification performance.
Subsequent studies investigated prediction disagreement, either between independently trained models~\cite{jiang2022assessinggeneralizationsgddisagreement} or between model outputs with and without dropout~\cite{baek2022agree, lee2024aetta}, observing that higher disagreement often indicates poorer generalization.
Other studies introduced self-supervised surrogate tasks, such as rotation prediction~\cite{deng2021rotation} and contrastive learning~\cite{peng2023camecontrastiveautomatedmodel}, providing label-free proxies for representation quality.
Another line of research leveraged dataset-level feature statistics, such as Fr\'{e}chet distances~\cite{deng2021autoeval} and average feature energy~\cite{peng2024energy}, demonstrating correlations with model performance.
While these methods have proven effective in AutoEval for image classification, the extension to other computer vision tasks, such as object detection, remains underexplored.

\vspace{4pt}
\noindent\textbf{AutoEval for Object Detection}
extends the concept of AutoEval to object detection, where performance depends on both object localization and classification, requiring the method to capture intricate spatial relationships among objects. Recently, Yang et al.~\cite{yang2024bos} proposed a method applicable to AutoEval for object detection, which compares bounding boxes generated by a detector with and without Monte Carlo dropout~\cite{gal2016dropout} perturbations applied to feature maps. The bounding box stability is measured by the difference between the original and perturbed predictions, which serves as a score for estimating detection performance.
However, this method has several limitations.
First, it is stochastic at test time, leading to inconsistent estimates across trials.
Second, it requires an additional forward pass to compute perturbed predictions, effectively doubling the inference cost.
Third, it does not incorporate model confidence, which reflects prediction uncertainty~\cite{guo2017calibration}, leading to an overestimation of unreliable predictions.
Furthermore, the meta-dataset constructed in their work relies on strong data augmentations with a fixed set of hyperparameters, which often yield unrealistic images and may fail to capture the full spectrum of detection performance in real-world scenarios.
In response, we propose an efficient and effective AutoEval method for object detection that explicitly leverages confidence scores to estimate detection performance.
In contrast to the prior work incorporating an additional forward pass with dropout~\cite{yang2024bos}, our proposed method utilizes pre-NMS boxes that are generated in a single forward pass, while containing valuable information about localization~\cite{bodla2017soft}.
Moreover, we construct a meta-dataset using image corruptions of varying severity, enabling a more realistic and scalable evaluation.

\vspace{4pt}
\noindent\textbf{Model Confidence in Object Detection} is quantified as confidence scores per class associated with each bounding box predicted by object detection models. These scores are typically interpreted as classification probabilities, and prior work has primarily focused on calibrating them~\cite{Kuppers_2020conf_cal, pathiraja2023multiclassconf}, with limited exploration of alternative interpretations or uses. In contrast, Sun et al.~\cite{sun2023confidencedriven} proposed a confidence-based bounding box localization strategy for small objects. However, their approach considered confidence scores a measure of uncertainty, and the explicit relationship between confidence scores and localization quality remains unexplored. Motivated by our observation that confidence scores reflect localization quality, we propose a confidence-driven approach to estimate detector performance.

\section{Prediction Consistency and Reliability}
\label{sec:pcr}

\begin{figure}[t]
    \centering
    \includegraphics[width=\columnwidth]{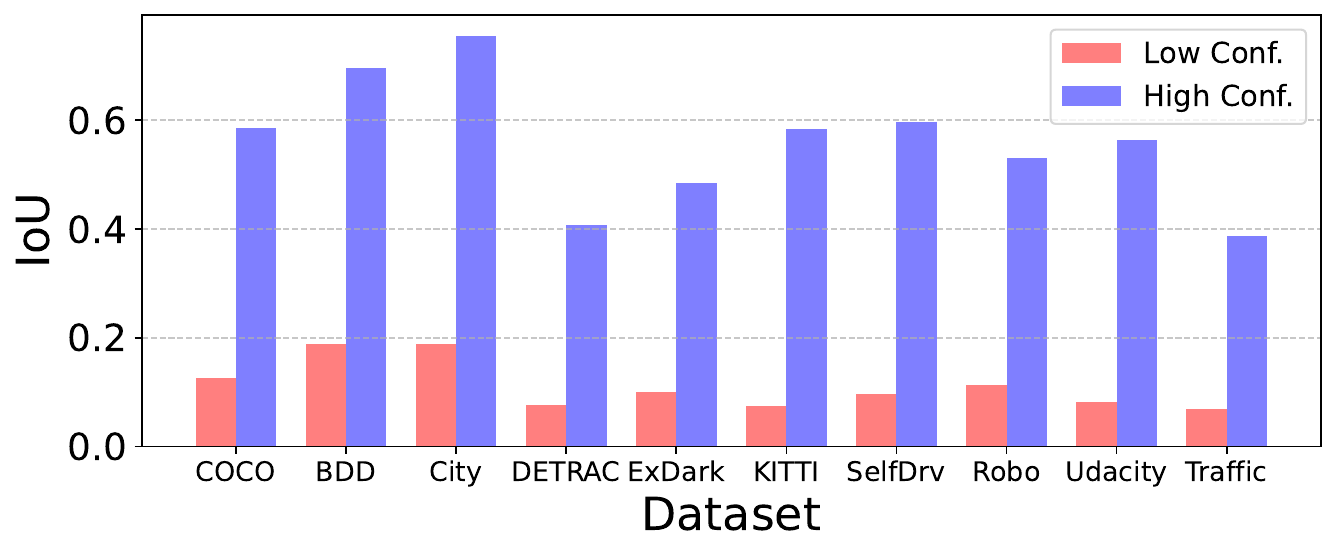}
    \caption{The average IoU between ground-truth boxes and final predictions, grouped by confidence level using a threshold of 0.5 across datasets. Predictions with low confidence generally exhibit lower IoU than those with high confidence, indicating a correlation between confidence and localization quality.}
    \label{fig:IoU}
\end{figure}

In this section, we introduce PCR, an effective AutoEval method for object detection. PCR consists of two components: for each final prediction---or equivalently, each post-NMS box---1) the \textit{consistency score} measures the spatial consistency with its corresponding merged pre-NMS box, and 2) the \textit{reliability score} measures the proportion of overlapping pre-NMS boxes with high confidence scores.

Conventional object detection models generate multiple candidate bounding boxes, followed by a filtering step such as Non-Maximum Suppression (NMS)~\cite{ren2015fasterrcnnrealtimeobject, lin2017focallossdenseobject, tian2019fcos, he2017mask, redmon2016you}.
Although NMS is essential for removing redundant boxes to produce final predictions, discarded pre-NMS boxes still contain valuable information about localization~\cite{bodla2017soft}.
This motivates us to investigate pre-NMS boxes associated with final predictions to facilitate AutoEval for object detection.

\subsection{Consistency}
\label{sec:consistency}

\begin{figure}[t]
    \centering
    \includegraphics[width=\columnwidth]{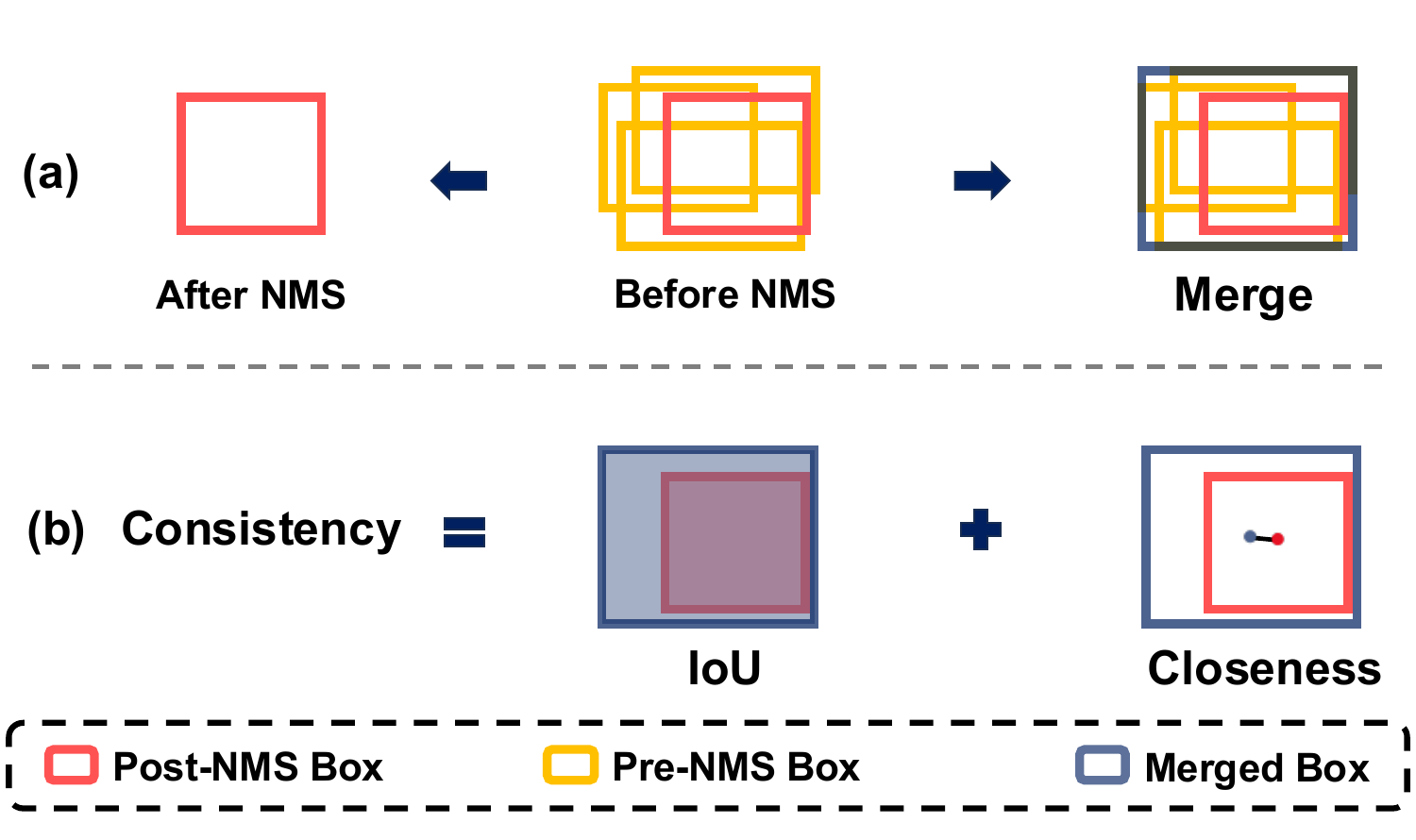}
    \caption{
        (a) A merged box tightly encloses all pre-NMS boxes associated with a post-NMS box.
        (b) Consistency is computed based on IoU and a closeness term measured by the normalized distance between the center points of a post-NMS box and its corresponding merged box.
    }
    \vspace{-20pt}
    \label{fig:merge_consistency}
\end{figure}

The consistency score is motivated by the observation in \Cref{fig:IoU} that the Intersection over Union (IoU) between a final prediction and its ground-truth box is correlated with the confidence score, \ie, predictions with low confidence tend to exhibit low IoU with ground-truth boxes.
This suggests that model confidence, which is often used as a proxy for classification accuracy~\cite{guo2017calibration}, also serves as an indicator of localization performance.
This becomes more evident when considering the pre-NMS boxes associated with a final prediction:
these boxes always have even lower confidence, implying that the detector consistently focuses on a region that does not contain any ground-truth object, which in turn correlates with low mAP.
However, measuring consistency for each individual pre-NMS box may introduce redundancy and lead to unstable assessments. Instead, we merge the pre-NMS boxes and compare the resulting merged box with the final prediction, as illustrated in \Cref{fig:merge_consistency}.

Let
\(\mathcal{B}^{(i)} = \left\{B^{(ij)}\right\}_{j=1}^{K_i}\) 
denote the set of pre-NMS boxes overlapping with the \(i\)-th final prediction \(B_{\mathrm{final}}^{(i)}\), where \(K_i\) is the number of pre-NMS boxes.
The merged box \(B_{\mathrm{merge}}^{(i)} = \left(x_{\mathrm{merge}}^{(i)}, y_{\mathrm{merge}}^{(i)}, w_{\mathrm{merge}}^{(i)}, h_{\mathrm{merge}}^{(i)}\right)\) is defined as follows:

\begin{align}
    x_{\mathrm{merge}}^{(i)} &= \left( \min_{j} x_{\mathrm{min}}^{(ij)} + \max_{j} x_{\mathrm{max}}^{(ij)} \right) / \, 2, \nonumber\\
    y_{\mathrm{merge}}^{(i)} &= \left( \min_{j} y_{\mathrm{min}}^{(ij)} + \max_{j} y_{\mathrm{max}}^{(ij)}\right) / \, 2, \nonumber\\
    w_{\mathrm{merge}}^{(i)} &= \max_{j} x_{\mathrm{max}}^{(ij)} - \min_{j} x_{\mathrm{min}}^{(ij)}, \nonumber\\
    h_{\mathrm{merge}}^{(i)} &= \max_{j} y_{\mathrm{max}}^{(ij)} - \min_{j} y_{\mathrm{min}}^{(ij)},
\label{eq:merge}
\end{align}
where \(x_{\mathrm{min}}^{(ij)}\), \(x_{\mathrm{max}}^{(ij)}\), \(y_{\mathrm{min}}^{(ij)}\), \(y_{\mathrm{max}}^{(ij)}\) denote the left, right, top, and bottom coordinates of the pre-NMS box \(B^{(ij)}\), respectively.
Then, the merged box tightly encloses all pre-NMS boxes, as shown in \Cref{fig:merge_consistency}(a).

Given a final prediction \(B_{\mathrm{final}}^{(i)}\) and its corresponding merged pre-NMS box \(B_{\mathrm{merge}}^{(i)}\), we measure their consistency using two metrics:
IoU and the closeness between their centers, as illustrated in \Cref{fig:merge_consistency}(b).
While IoU is a standard measure of spatial alignment, it can be misleading when the merged box is elongated, where small shifts along the shorter axis may cause significant drops in IoU, whereas large shifts along the longer axis may still yield high IoU despite localization errors.
Moreover, because the merged box is generally large and tends to enclose the final prediction, IoU alone may not accurately capture their geometric relationship~\cite{zheng2020distance}.
To address this, we introduce an additional measure that quantifies the \textbf{C}loseness between the \textbf{C}enters (CC) of the final prediction and the merged box:
\begin{align}
    &\operatorname{CC}\left(B_{\mathrm{final}}^{(i)}, B_{\mathrm{merge}}^{(i)}\right) = \nonumber\\
    &\quad 1 - \frac{\sqrt{\left(x_{\mathrm{final}}^{(i)}-x_{\mathrm{merge}}^{(i)}\right)^2 +\left(y_{\mathrm{final}}^{(i)}- y_{\mathrm{merge}}^{(i)}\right)^2}}{\sqrt{w_{\mathrm{final}}^{(i) \, 2} +h_{\mathrm{final}}^{(i) \, 2} } \, / \, 2}.
\label{eq:closeness}
\end{align}
Intuitively, CC is computed as one minus the normalized distance between the two centers, where the normalization factor is the half-diagonal of the final prediction box, ensuring scale invariance. A higher CC value indicates that two centers are closer, with a value of 1 implying perfect alignment.\footnote{CC resembles the normalized distance in DIoU~\cite{zheng2020distance}, differing in several aspects, \eg, the normalization factor.}
The consistency score of the \(i\)-th final prediction is defined as the average of IoU and CC:
{%
\thinmuskip=1mu 
\medmuskip=2mu plus 2mu minus 4mu 
\thickmuskip=3mu plus 5mu 
\begin{align}
    S^{\mathrm{C} (i)} = \frac{\mathrm{IoU}\left(B_{\mathrm{final}}^{(i)}, B_{\mathrm{merge}}^{(i)}\right) + \operatorname{CC}\left(B_{\mathrm{final}}^{(i)}, B_{\mathrm{merge}}^{(i)}\right)}{2}.
\label{eq:consistency_per}
\end{align}
The consistency score of an image is computed as the weighted average of per-prediction consistency scores, where each weight is a scaled version of the confidence score of the corresponding final prediction:
\begin{align}
    S^{\mathrm{C}} = \frac{1}{N}\sum_{i=1}^N S^{\mathrm{C} (i)} \cdot \sigma_C\left( h \left(B_{\mathrm{final}}^{(i)}\right) \right),  
\label{eq:consistency}
\end{align}
where
\(h(\cdot)\) denotes the confidence score of a prediction,
\(\sigma_C(x) = 1 / (1 + \exp(-k_C(x - c)))\) is a sigmoid function parameterized by a negative scale \(k_C\) and a confidence threshold \(c\), and
\(N\) is the number of final predictions.
The function \(\sigma_C\) approaches one for low-confidence boxes and zero for high-confidence ones, thus emphasizing the final predictions with low confidence.
A high consistency score indicates that the detector consistently focuses on a region without a ground-truth object, which is associated with low mAP.
\Cref{fig:IoU_corr}(a) confirms that the consistency score \(S^{\mathrm{C}}\) exhibits a strong negative correlation with mAP.

\begin{figure*}
    \subfloat[Consistency score vs. mAP.]
    {\includegraphics[width=0.58\columnwidth]{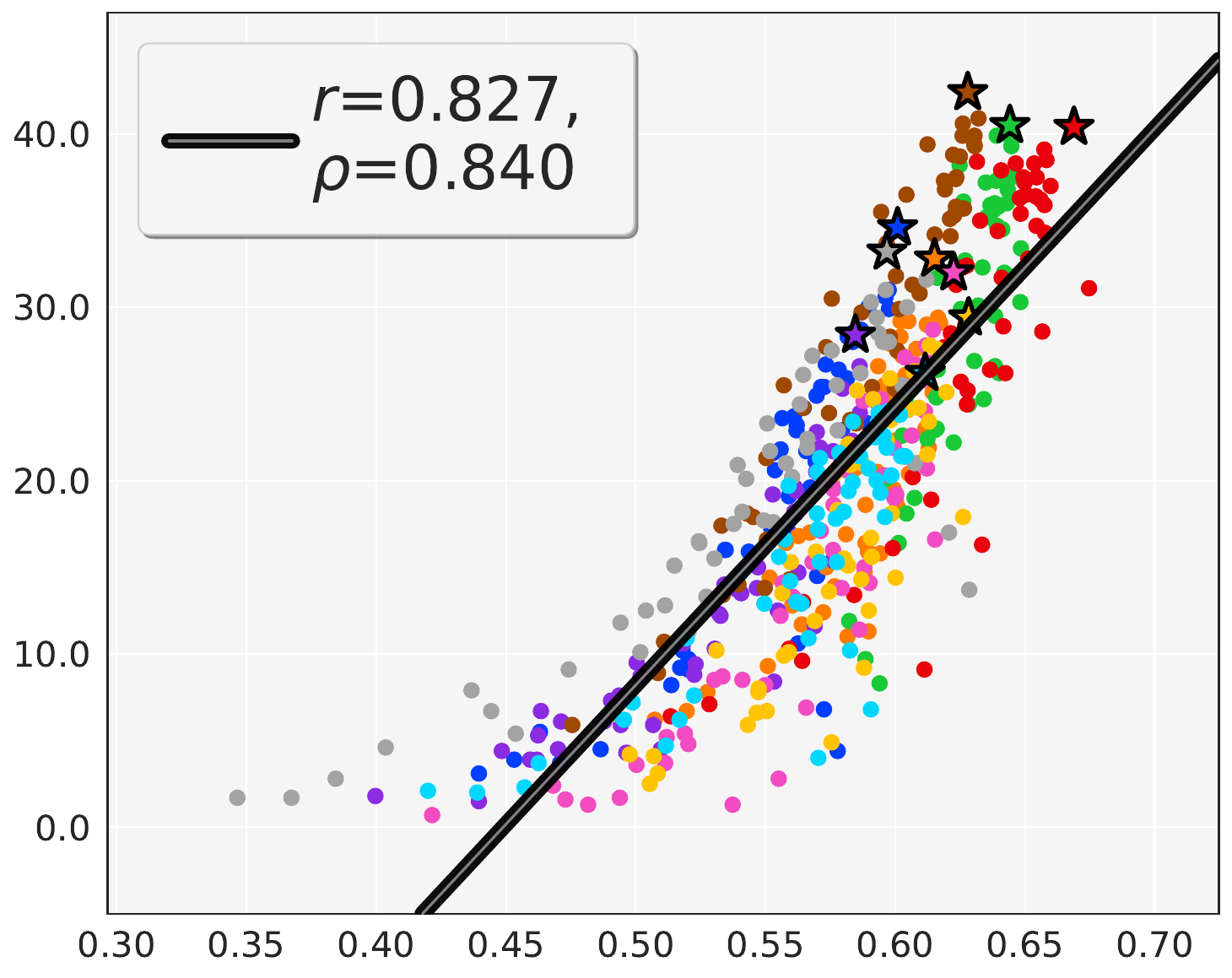}\label{fig:consistency}}
    \subfloat[Examples of \( S^{\mathrm{C}}\).]
    {\includegraphics[width=0.32\columnwidth]{figs/fig2_con2.pdf}\label{fig:consistency_ex}}
    \hspace{2pt}
    \includegraphics[width=0.29\columnwidth]{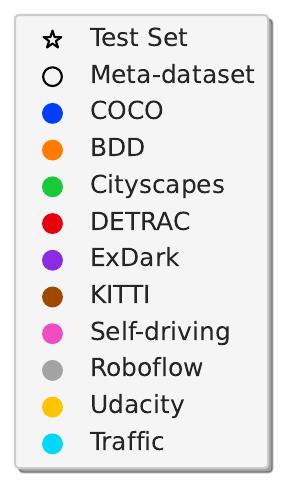}
    \hspace{-6pt}
    \subfloat[Reliability score vs. mAP.]
    {\includegraphics[width=0.58\columnwidth]{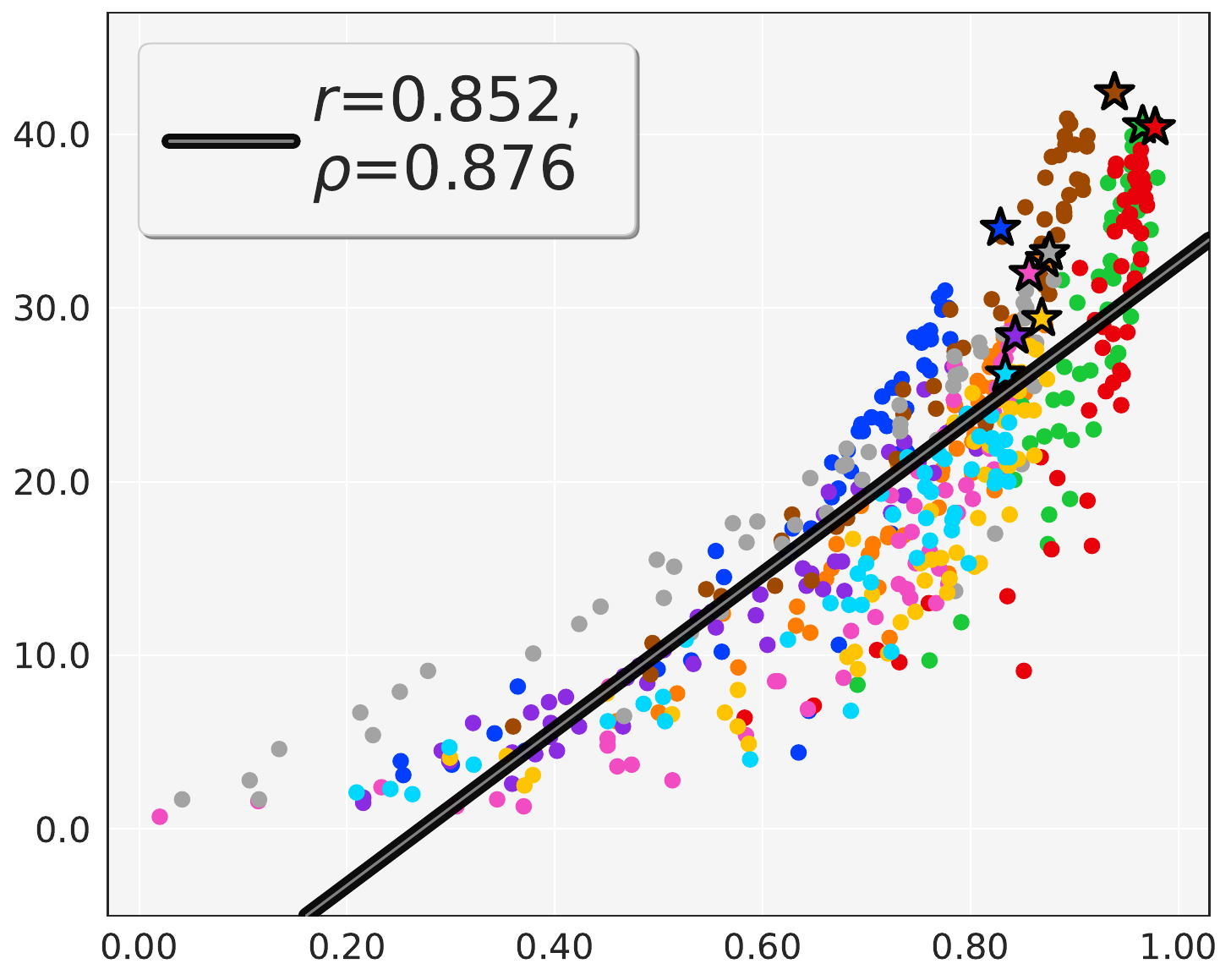}\label{fig:reliability}}
    \subfloat[Examples of \( S^{\mathrm{R}}\).]
    {\includegraphics[width=0.32\columnwidth]{figs/fig2_rel2.pdf}\label{fig:reliability_ex}}

    \vspace{-5pt}
    \caption{
        (a) The consistency score \(S^{\mathrm{C}}\) shows a strong \textit{negative} correlation with mAP. 
        (b) Predictions with low confidence and high consistency suggest that the detector consistently localizes the same region without any object, indicating a detection failure. In contrast, predictions with low confidence and low consistency provide insufficient information to make a decision.
        (c) The reliability score \(S^{\mathrm{R}}\) shows a strong \textit{positive} correlation with mAP. 
        (d) Predictions with high confidence and high reliability suggest that the detector repeatedly localizes and classifies the same object, indicating a successful detection. In contrast, predictions with high confidence and low reliability provide insufficient information to make a decision.
      }
    \label{fig:IoU_corr}
    \vspace{-10pt}
\end{figure*}

Note that our notion of consistency differs from prior works that assess the localization stability between model predictions and those obtained by additional forward passes with model transformations or image perturbations~\cite{jeong2019consistency, yang2024bos}.
Instead, our consistency score quantifies the spatial similarity among pre- and post-NMS boxes generated from a single forward pass.
Indeed, we observe no correlation between our consistency score and the BoS~\cite{yang2024bos} score, highlighting their conceptual distinction.

\subsection{Reliability}
\label{sec:reliability}

The reliability score is also motivated by \Cref{fig:IoU}, with a focus on final predictions with high confidence, which tend to exhibit high IoU with ground-truth boxes.
Intuitively, NMS is employed in object detection to select the most probable bounding box among overlapping pre-NMS candidate boxes, based on confidence scores.
A concentration of pre-NMS boxes with high confidence indicates that both the classification and regression heads repeatedly identify an object at the same location, thereby promoting the reliability of the prediction.
However, this reliability diminishes if some overlapping pre-NMS boxes have low confidence, indicating uncertainty in the prediction around the location.
Hence, we define the reliability score as the proportion of pre-NMS boxes with confidence scores above the confidence threshold \(c\).

Given the final predictions \(\{B_{\mathrm{final}}^{(i)}\}_{i=1}^{N}\) and 
the set of \textit{all} pre-NMS boxes for an image \(\mathcal{P} = \bigcup_{i} \mathcal{B}^{(i)}\)\footnote{%
Some pre-NMS boxes may overlap multiple final predictions;
to ensure that each pre-NMS box is counted only once, we take the union over all pre NMS-boxes.
}, the reliability score of an image is defined as:
\begin{align}
    S^{\mathrm{R}} = \frac{ \sum_{i=1}^{N}\sum_{j=1}^{K_i}
    \mathbb{I}{\left[h(B_{\mathrm{final}}^{(i)}) > c\right]}
    \sigma_R\left( h \left(B^{(ij)}\right) \right)}
    { \sum_{B^{(ij)} \in \mathcal{P}} \sigma_R\left( h \left(B^{(ij)}\right) \right) },
\label{eq:reliability}
\end{align}
where \(\mathbb{I}[\cdot]\) is the indicator function and \(\sigma_R(x) = \alpha + (1 - \alpha) / (1 + \exp(-k_R(x - c)))\) is a sigmoid function parameterized by a positive scale \(k_R\), the confidence threshold \(c\), and a floor value \(\alpha\).
The function \(\sigma_R\) approaches one for high-confidence boxes and \(\alpha\) for low-confidence ones, thus emphasizing the pre-NMS boxes with high confidence while retaining a minimal contribution from those with low confidence. A high reliability score indicates that the detector repeatedly assigns high confidence to a region, suggesting strong consensus in both classification and localization, which is associated with high mAP.
\Cref{fig:IoU_corr}(b) confirms that the reliability score \(S^{\mathrm{R}}\) exhibits a strong positive correlation with mAP.

\subsection{PCR for AutoEval}

We extend the notation of our scores to explicitly include an object detector \(f\) and an input image \(\mathcal{I}\).
Our proposed PCR consists of two scores:
the consistency score \(S^{\mathrm{C}}(f, \mathcal{I})\) and
the reliability score \(S^{\mathrm{R}}(f, \mathcal{I})\).
These scores are evaluated over a meta-dataset
\(\mathcal{D} = \{\mathcal{D}_m\}_{m=1}^{M}\), where each dataset \(\mathcal{D}_m\) is a transformed version of the source dataset, \eg, using augmentations or image corruptions.
Given an object detector \(f\) and the \(m\)-th dataset \(\mathcal{D}_m\), the average scores are computed as:
\begin{align}
    \bar{S}^{\mathrm{C}} (f, \mathcal{D}_m) &= \frac{1}{\lvert \mathcal{D}_m \rvert} \sum_{\mathcal{I} \in \mathcal{D}_m} S^{\mathrm{C}}(f, \mathcal{I}), \nonumber \\
    \bar{S}^{\mathrm{R}} (f, \mathcal{D}_m) &= \frac{1}{\lvert \mathcal{D}_m \rvert} \sum_{\mathcal{I} \in \mathcal{D}_m} S^{\mathrm{R}}(f, \mathcal{I}).
\end{align}
To perform AutoEval, we fit a linear regression model using least squares over the meta-dataset.
Given an object detector \(f\) and the \(m\)-th dataset $\mathcal{D}_m$, the mAP is estimated as:
\begin{align}
    \widehat{\mathrm{mAP}}(f, \mathcal{D}_m) = w_0 + \sum_{t=1}^{T} w_t \cdot \bar{S}_t(f,\mathcal{D}_m),
\end{align}
where $\{w_t\}_{t=0}^{T}$ are the regression coefficients and $\bar{S}_t$ is the $t$-th AutoEval score.
For PCR, the mAP estimate is:
\begin{align*}
    \widehat{\mathrm{mAP}}(f, \mathcal{D}_m) = w_0 + w_1 \cdot \bar{S}^{\mathrm{C}} (f, \mathcal{D}_m) + w_2 \cdot \bar{S}^{\mathrm{R}} (f, \mathcal{D}_m).
\end{align*}
For training and evaluation, we adopt leave-one-out cross-validation, following Yang et al.~\cite{yang2024bos}:
a meta-dataset is constructed using all source datasets except one for training, while the remaining source dataset is held out for evaluation and used without any transformations.

\section{Corruption-Based Meta-Dataset}
\label{sec:dataset}

\begin{figure}[t]
  \begin{minipage}[t]{0.32\linewidth}
    \centering
    \includegraphics[width=\linewidth]{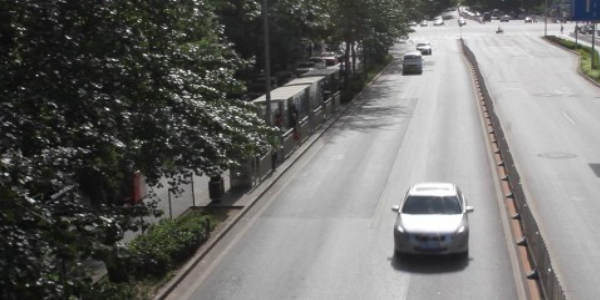} \\
    \includegraphics[width=\linewidth]{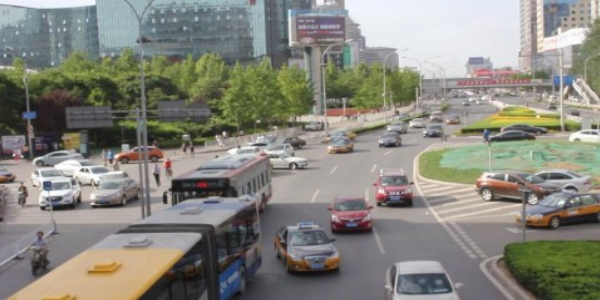} \\
    {\scriptsize (a) Source Dataset}
  \end{minipage}
  \begin{minipage}[t]{0.32\linewidth}
    \centering
    \includegraphics[width=\linewidth]{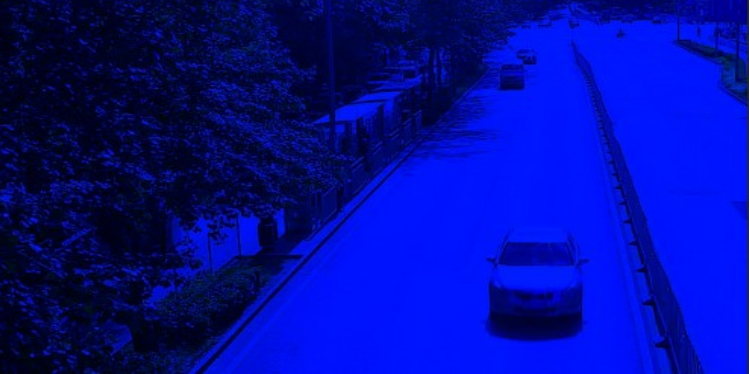} \\
    \includegraphics[width=\linewidth]{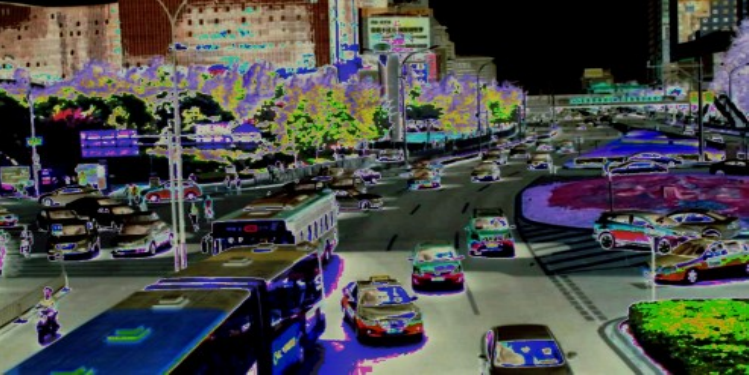} \\
    {\scriptsize (b) Augmented~\cite{yang2024bos}}
  \end{minipage}
  \begin{minipage}[t]{0.32\linewidth}
    \centering
    \includegraphics[width=\linewidth]{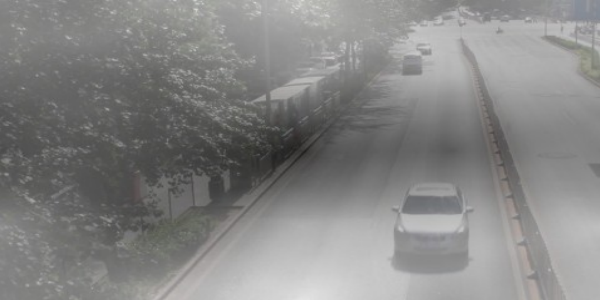} \\
    \includegraphics[width=\linewidth]{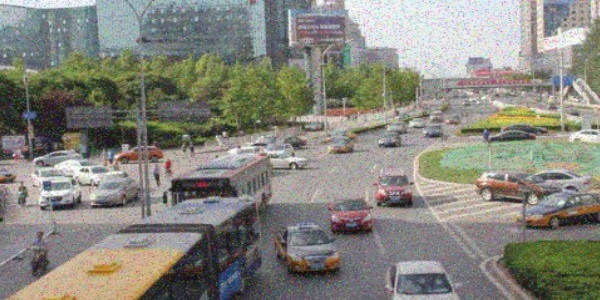} \\
    {\scriptsize (c) Corrupted (Ours)}
  \end{minipage}
  \centering
  \includegraphics[width=\columnwidth]{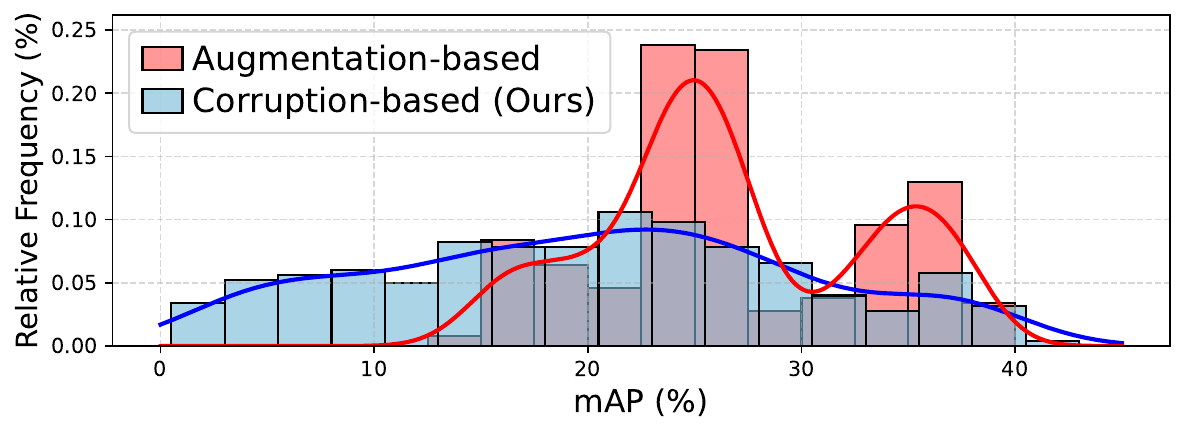} \\
  \vspace{-5pt}
  {\scriptsize (d) \textbf{Distribution of mAP:} Augmentation-based~\cite{yang2024bos} vs. Corruption-based (Ours)} \\
  \caption{
    (a--c) Sample images from:
    (a) the original DETRAC~\cite{wen2020ua} dataset,
    (b) the augmentation-based meta-dataset, and
    (c) the corruption-based meta-dataset.
    (d) Histogram of mAP distributions for the augmentation-based meta-dataset~\cite{yang2024bos} and our proposed corruption-based meta-dataset.
  }
  \label{fig:dataset_diff}
\end{figure}

To assess AutoEval methods for object detection, a meta-dataset is typically constructed by applying transformations to source datasets.
While prior work~\cite{yang2024bos} adopts strong data augmentation techniques for this purpose, we argue that such augmentations may not adequately capture the environmental shifts or discrepancies in data distribution observable in real-world scenarios.
Consequently, such a meta-dataset may fail to span the full spectrum of detection performance observable in practice.

To address this, we instead adopt the transformations applied to ImageNet-C~\cite{hendrycks2019benchmarking}, which simulate real-world corruptions.
To capture scenarios of varying difficulty, we apply each corruption at different severity levels.
Specifically, the meta-dataset consists of 50 datasets using ten corruptions \texttt{\{gaussian noise, shot noise, impulse noise, defocus blur, snow, frost, fog, contrast, pixelate, jpeg\}} with five severity levels \texttt{\{1, 2, 3, 4, 5\}}.
We exclude corruptions that alter the coordinates of bounding boxes, \eg, \texttt{\{zoom blur, elastic transformation\}}.
This strategy yields a meta-dataset that is both realistic---as it reflects real-world corruptions rather than artificial augmentations---and scalable---as the varying severity levels enable evaluation across a broader spectrum of detection performance.

\Cref{fig:dataset_diff} compares the augmentation-based meta-dataset from prior work~\cite{yang2024bos} and our proposed corruption-based meta-dataset.
As shown in \Cref{fig:dataset_diff}(b), images generated by strong augmentations often appear artificial, whereas those generated by corruptions are more realistic, as shown in \Cref{fig:dataset_diff}(c).
In terms of performance coverage, \Cref{fig:dataset_diff}(d) shows that the augmentation-based meta-dataset exhibits a relatively narrow and skewed distribution of mAP, with limited coverage below 15\% and a concentration around 25\% and 35\%.
In contrast, our corruption-based meta-dataset covers a broader range of mAP, spanning from near 0\% to 40\% with a smoother distribution.

\section{Experiments}
\label{sec:exp}

\subsection{Experimental Setup}

\noindent\textbf{Detectors.} We compare our proposed PCR with baseline AutoEval methods using four object detectors: RetinaNet~\cite{lin2017focallossdenseobject} and Faster R-CNN~\cite{ren2015fasterrcnnrealtimeobject} paired with ResNet-50~\cite{he2016deep} and Swin Transformer~\cite{liu2021swin} backbones.

\vspace{4pt}
\noindent\textbf{Meta-Dataset.} Following the dataset sampling strategy in prior work~\cite{yang2024bos}, each dataset in the meta-dataset contains 250 images sampled from a source dataset.
For \textbf{vehicle detection}, we use 10 source datasets:
COCO~\cite{lin2014microsoft},
BDD~\cite{yu2020bdd100k},
Cityscapes~\cite{cordts2015cityscapes},
DETRAC~\cite{wen2020ua},
ExDark~\cite{loh2019getting},
KITTI~\cite{geiger2013vision},
Self-driving~\cite{Kaggle_SelfDriving_2020},
Roboflow~\cite{Kaggle_Roboflow_2022},
Udacity~\cite{Kaggle_Udacity_2021}, and
Traffic~\cite{Kaggle_TrafficVehicles_2020}.
For \textbf{pedestrian detection}, we use 9 source datasets:
COCO~\cite{lin2014microsoft},
Caltech~\cite{griffin2007caltech},
CityPersons~\cite{zhang2017citypersons},
Cityscapes~\cite{cordts2015cityscapes},
CrowdHuman~\cite{shao2018crowdhuman},
ECP~\cite{braun2019eurocity},
ExDark~\cite{loh2019getting},
KITTI~\cite{geiger2013vision}, and
Self-driving~\cite{Kaggle_SelfDriving_2020}.

\vspace{4pt}
\noindent\textbf{Baselines.}
We compare PCR with five AutoEval methods:
1) Prediction Score (PS)~\cite{hendrycks2018ps},
2) Entropy Score (ES)~\cite{saito2019semi},
3) Average Confidence (AC)~\cite{guillory2021predicting},
4) Average Thresholded Confidence (ATC)~\cite{arg2022leveragingunlabeleddatapredict}, and
5) Box Stability (BoS)~\cite{yang2024bos}. 
The first four methods were originally proposed for image classification;
we adapt them to object detection by computing their metrics using the confidence scores of the predicted bounding boxes.
For BoS, we report the average over three runs to account for the stochasticity introduced by MC dropout.
We use the same hyperparameters as reported in BoS~\cite{yang2024bos} for all baselines and present results based on our own replications.

\vspace{4pt}
\noindent\textbf{Implementation Details.}
For hyperparameters of PCR, we set 
the confidence threshold to 0.5,
the scale in \(\sigma^{\mathrm{C}}\) to \(k_C=-60\),
the scale in \(\sigma^{\mathrm{R}}\) to \(k_R=10\), and
the floor value in \(\sigma^{\mathrm{R}}\) to \(\alpha=0.2\).
Other implementation details and hyperparameter tuning results are provided in the supplementary material.

\vspace{4pt}
\noindent\textbf{Evaluation Metric.}
We evaluate performance using the root mean squared error (RMSE) between the estimated mAP and the true mAP, where a lower RMSE indicates a more accurate estimation.
Results in the correlation metric are reported in the supplementary material.

\subsection{Vehicle Detection}

\Cref{tab:car_detection} summarizes the mAP estimation results for vehicle detection.
We report results on both the augmentation-based meta-dataset~\cite{yang2024bos} and our proposed corruption-based meta-dataset.
Each cell presents the average RMSE between the estimated mAP and the true mAP for each method with a pair of detector and meta-dataset, where each detector is trained on the ``car'' class from COCO~\cite{lin2014microsoft}.
We also report the average RMSE across all four detectors and both meta-datasets, along with the average performance rank.
PCR achieves the lowest average RMSE of 5.03 and the best average rank of 1.13, consistently outperforming all baselines.

\subsection{Pedestrian Detection}

\Cref{tab:person_detection} summarizes the mAP estimation results for pedestrian detection.
For this experiment, detectors are trained on CrowdHuman~\cite{shao2018crowdhuman}.
Again, we report results on both the augmentation-based meta-dataset~\cite{yang2024bos} and our proposed corruption-based meta-dataset.
PCR achieves the lowest average RMSE of 3.60 and the best average performance rank of 1.00.
These results suggest that PCR offers accurate mAP estimates, consistently outperforming all baselines under fair evaluation settings.

\begin{table*}[t]
    \centering
    \renewcommand{\arraystretch}{1.1}
    \resizebox{\textwidth}{!}{
    \begin{tabular}{l|cc|cc|cc|cc|c|c}
        \toprule
         Meta-dataset& \multicolumn{4}{c|}{\text{Augmentation-based~\cite{yang2024bos}}} & \multicolumn{4}{c|}{\text{Corruption-based (Ours)}} & \multirow{3}{*}{\makecell{Avg. \\ RMSE}} & \multirow{3}{*}{\makecell{Avg. \\ Rank}}\\
        \cmidrule{1-9}
        \multirow{2}{*}{\text{Detector}} & \multicolumn{2}{c|}{\textbf{RetinaNet}} & \multicolumn{2}{c|}{\textbf{Faster R-CNN}} &  \multicolumn{2}{c|}{\textbf{RetinaNet}} & \multicolumn{2}{c|}{\textbf{Faster R-CNN}} & \\
        & ResNet-50 & Swin-T & ResNet-50 & Swin-T &  ResNet-50 & Swin-T & ResNet-50 & Swin-T &\\
        \midrule
        PS~\cite{hendrycks2018ps}   & 4.85 & 7.39 & 4.95 & 6.28 & \underline{11.30} & 10.44 & 10.41 & 6.30 & 7.74 & 3.13 \\
        ES~\cite{saito2019semi}   & 6.51 & \underline{5.58} & 5.74 & 6.46 & 14.87 & 7.20 & 12.53 & 10.02 & 8.62 & 4.75 \\
        AC~\cite{guillory2021predicting}   & 9.50 & 8.17 & 5.68 & 6.97 & 14.23 & 10.65 & 11.77 & 7.23 & 9.27 & 5.13 \\
        ATC~\cite{arg2022leveragingunlabeleddatapredict}  & 5.52 & 11.20 & 4.89 & 6.86 & 14.10 & 12.52 & 11.98 & \underline{6.29} & 9.17 & 4.38 \\
        BoS~\cite{yang2024bos}  & \underline{3.11} & 7.69 & \textbf{3.33} & \underline{5.57} & 13.50 & \underline{5.18} & \underline{10.32} & 6.85 & \underline{6.94} & \underline{2.50} \\
        \midrule
        PCR (Ours)  & \textbf{2.99} & \textbf{3.82} & \underline{3.98} & \textbf{3.36} & \textbf{6.57} & \textbf{4.26} & \textbf{7.23} & \textbf{4.70} & \textbf{4.61} & \textbf{1.13} \\
        \bottomrule 
    \end{tabular}
    }
    \caption{Comparison of AutoEval methods for vehicle detection using four detectors on two meta-datasets.
    The best result for each combination of detector and meta-dataset is highlighted in \textbf{bold} and the second-best is \underline{underlined}.}
    \label{tab:car_detection}
\end{table*}

\begin{table*}[t]
    \centering
    \renewcommand{\arraystretch}{1.1}
    \resizebox{\textwidth}{!}{
    \begin{tabular}{l|cc|cc|cc|cc|c|c}
        \toprule
         Meta-dataset & \multicolumn{4}{c|}{\text{Augmentation-based~\cite{yang2024bos}}} & \multicolumn{4}{c|}{\text{Corruption-based (Ours)}} & \multirow{3}{*}{\makecell{Avg. \\ RMSE}} & \multirow{3}{*}{\makecell{Avg. \\ Rank}}\\
        \cmidrule{1-9}
        \multirow{2}{*}{\text{Detector}} & \multicolumn{2}{c|}{\textbf{RetinaNet}} & \multicolumn{2}{c|}{\textbf{Faster R-CNN}} &  \multicolumn{2}{c|}{\textbf{RetinaNet}} & \multicolumn{2}{c|}{\textbf{Faster R-CNN}} & \\
        & ResNet-50 & Swin-T & ResNet-50 & Swin-T &  ResNet-50 & Swin-T & ResNet-50 & Swin-T &\\
        \midrule
        PS~\cite{hendrycks2018ps} & 7.47 & 7.35 & 6.20 & 4.64 & 10.29 & 10.71 & \underline{6.84} & 5.93 & 7.43 & 4.50 \\
        ES~\cite{saito2019semi} & 5.13 & 3.55 & 8.16 & 6.89 & 8.67 & 7.40 & 10.66 & 10.00 & 7.56 & 5.00 \\
        AC~\cite{guillory2021predicting} & 5.03 & 4.39 & 6.80 & 4.07 & \underline{7.54} & \underline{5.90} & 9.15 & \underline{5.37} & \underline{6.03} & \underline{3.13} \\
        ATC~\cite{arg2022leveragingunlabeleddatapredict} & 6.21 & 7.46 & 6.98 & 3.88 & 7.66 & 6.51 & 8.33 & 5.95 & 6.62 & 4.13 \\
        BoS~\cite{yang2024bos} & \underline{3.77} & \underline{3.17} & \underline{5.88} & \underline{3.57} & 10.23 & 6.08 & 10.09 & 6.98 & 6.22 & 3.25 \\
        \midrule
        PCR (Ours) & \textbf{2.88} & \textbf{2.71} & \textbf{3.39} & \textbf{2.57} & \textbf{2.96} & \textbf{3.65} & \textbf{5.54} & \textbf{4.88} & \textbf{3.57} & \textbf{1.00} \\
        \bottomrule
    \end{tabular}
    }
    \caption{Comparison of AutoEval methods for pedestrian detection using four detectors on two meta-datasets.
    The best result for each combination of detector and meta-dataset is highlighted in \textbf{bold} and the second-best is \underline{underlined}.}
    \label{tab:person_detection}
\end{table*}

\subsection{Analysis}

We conduct experiments to analyze both the proposed method and the meta-dataset.
Unless otherwise specified, all experiments use RetinaNet~\cite{lin2017focallossdenseobject} with a ResNet-50~\cite{he2016deep} backbone evaluated on the corruption-based meta-dataset.

\vspace{4pt}
\noindent\textbf{Robustness of PCR Across Varying Difficulty.}
Recall that the meta-dataset includes five severity levels of corruption to reflect varying difficulties observable in real-world applications.
To assess the robustness of our proposed PCR across a broader spectrum of detection performance, we perform linear regression separately on each 10 datasets corresponding to a single severity level.

\Cref{fig:severity_PCR_BoS} compares the average RMSE of BoS~\cite{yang2024bos} and PCR.
While the average RMSE increases with the severity level for both methods, PCR consistently achieves a lower RMSE and demonstrates robustness across different severity levels.
This suggests that PCR not only performs well within a specific range of mAP, but also generalizes across diverse conditions, enabling reliable AutoEval across a wide range of performance distributions.

\begin{figure}[t]
  \begin{minipage}[t]{0.46\columnwidth}
    \vspace{0pt}
    \centering
    \includegraphics[width=\linewidth]{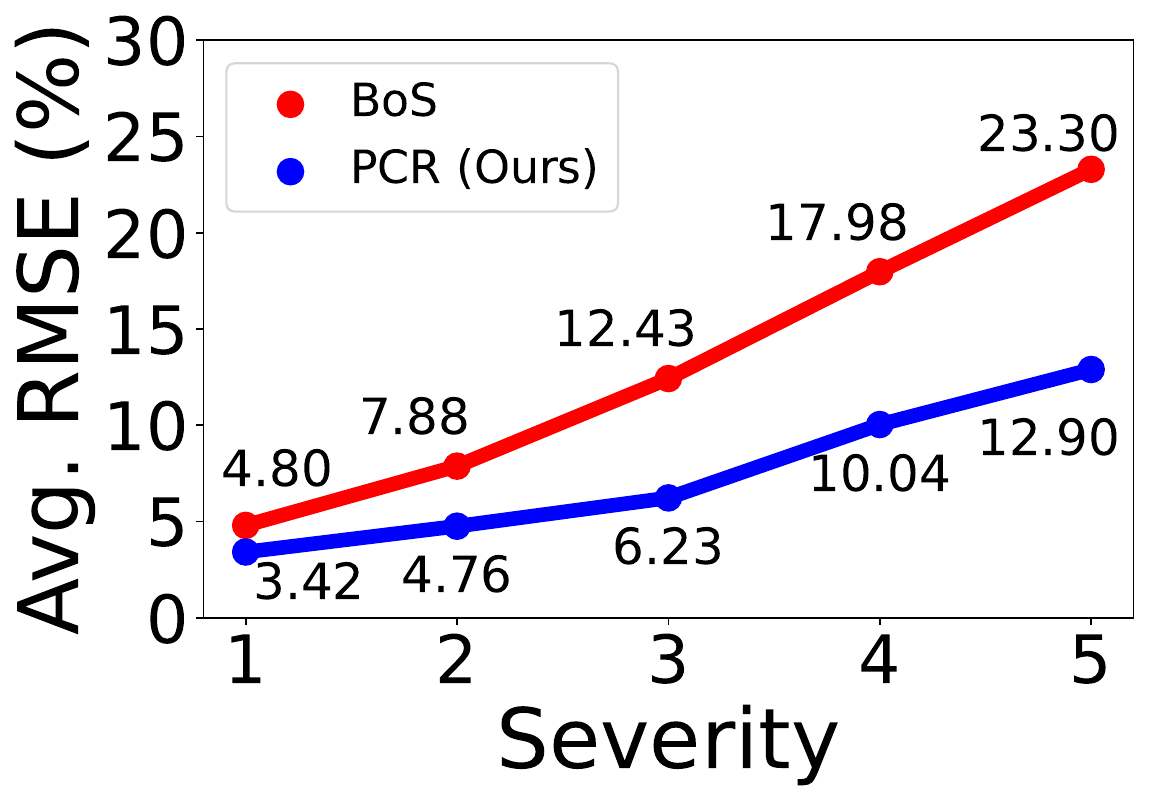}
    \vspace{-20pt}
    \captionof{figure}{Severity vs. RMSE.}
    \label{fig:severity_PCR_BoS}
  \end{minipage}
  \hfill
  \begin{minipage}[t]{0.51\columnwidth}
    \vspace{0pt}
    \centering
    \resizebox{\linewidth}{!}{%
    \begin{tabular}{cc|c}
      \toprule
      \(S^{\mathrm{C}}\) & \(S^{\mathrm{R}}\) & Avg.\ RMSE \\
      \midrule
      \checkmark &             & 6.75 \\
                 & \checkmark  & 6.64 \\
      \midrule
      \checkmark & \checkmark  & \textbf{6.57} \\
      \bottomrule
    \end{tabular}
    }
    \captionof{table}{Ablation on \(S^{\mathrm{C}}\) and \(S^{\mathrm{R}}\).}
    \label{tab:PCR_key_components}
  \end{minipage}
  \label{fig:conf_thresh_n_component}
\end{figure}

\vspace{4pt}
\noindent\textbf{Ablation on Components of PCR.}
To evaluate the contribution of each component in PCR, we conduct an ablation study by selectively removing either the consistency or the reliability score.
\Cref{tab:PCR_key_components} shows that using either score alone already outperforms the baseline methods, while combining both scores achieves the best performance.
This result suggests that these two scores capture complementary aspects of detection performance:
while the consistency score focuses on low-confidence predictions, the reliability score focuses on high-confidence predictions.
Their combination allows PCR to effectively capture the relationships among pre- and post-NMS boxes based on the confidence scores.

\begin{figure}[t]
  \centering
  \begin{minipage}[t]{0.48\columnwidth}
    \vspace{0pt}
    \centering
    \renewcommand{\arraystretch}{1.1}
    \setlength{\tabcolsep}{4pt}
    \resizebox{\linewidth}{!}{%
      \begin{tabular}{c|c}
        \toprule
        Consistency Score & Avg.\ RMSE \\
        \midrule
        IoU only      & 6.75 \\
        + Closeness   & \textbf{6.64} \\
        \bottomrule
      \end{tabular}%
    }
    \captionof{table}{Ablation on components of consistency.}
    \label{tab:consistency_component}
  \end{minipage}%
  \hfill
  \begin{minipage}[t]{0.5\columnwidth}
    \vspace{0pt}
    \centering
    \renewcommand{\arraystretch}{1.1}
    \setlength{\tabcolsep}{4pt}
    \resizebox{\linewidth}{!}{%
      \begin{tabular}{c|c}
        \toprule
        Consistency Scaling & Avg.\ RMSE \\
        \midrule
        All (\(S^{\mathrm{C}}_{\text{all}}\))   & 8.24 \\
        Low (\(S^{\mathrm{C}}\)) & \textbf{6.64} \\
        \bottomrule
      \end{tabular}%
    }
    \captionof{table}{Ablation on confidence-based scaling.}
    \label{tab:consistency_range}
  \end{minipage}
\end{figure}

\vspace{4pt}
\noindent\textbf{Ablation on Components of Consistency.}
While IoU is a widely used metric to quantify the similarity between two boxes, our consistency score additionally incorporates the closeness between the centers of the final prediction and the corresponding merged pre-NMS box.
To confirm the effectiveness of this additional component, we conduct an ablation study with and without it.
As shown in \Cref{tab:consistency_component}, incorporating the closeness of centers improves the consistency measure, leading to improved performance.

\vspace{4pt}
\noindent\textbf{Effect of Confidence-Based Scaling in Consistency.}
In PCR, the consistency score emphasizes low-confidence predictions by scaling with confidence scores, following the intuition discussed in \Cref{sec:consistency}.
To validate this design choice, we conduct an ablation study comparing the proposed consistency score in \cref{eq:consistency} and a variant that does not apply confidence-based scaling, which is formulated as:
\begin{align}
    S^{\mathrm{C}}_{\mathrm{all}} = \frac{1}{N}\sum_{i=1}^N S^{\mathrm{C} (i)},  
\label{eq:consistency_all}
\end{align}
\Cref{tab:consistency_range} presents the RMSE with and without confidence-based scaling.
We observe that scaling the consistency scores to weight low-confidence predictions more results in better performance.
This supports our design choice to focus on low-confidence predictions. 

\begin{table}[t]
    \centering
    \begin{tabular}{l|ccc}
    \toprule
    \multirow{2}{*}{\text{Method}} & \multicolumn{3}{c}{\text{RetinaNet + ResNet-50}} \\
    & \text{mAP} & \text{mAP\textsubscript{50}} & \text{mAP\textsubscript{75}} \\
    \midrule
    PS~\cite{hendrycks2018ps}  &  11.30 & 18.64 & 12.16 \\
    ES~\cite{saito2019semi} &  14.87 & 25.12 & 15.63 \\
    AC~\cite{guillory2021predicting}  & 14.23 & 23.58 & 15.24 \\
    ATC~\cite{arg2022leveragingunlabeleddatapredict} & 14.10 & 23.33 & 15.10\\
    BoS~\cite{yang2024bos} &  13.50 & 22.75 & 14.43\\
    \midrule
    PCR (Ours) & \textbf{6.57} & \textbf{10.18} & \textbf{7.94} \\
    \bottomrule
    \end{tabular}
    \caption{RMSE Comparison of AutoEval methods in estimating mAP, mAP\textsubscript{50} and mAP\textsubscript{75}.}
    \label{tab:mAP50andmAP75}
\end{table}

\vspace{4pt}
\noindent\textbf{Performance on mAP\textsubscript{50} and mAP\textsubscript{75}.}
So far, our experiments have focused on estimating mAP, which averages performance across 10 IoU thresholds ranging from 0.50 to 0.95.
To assess whether our proposed PCR is also eligible to estimate other metrics, we conduct experiments on estimating mAP\textsubscript{50} and mAP\textsubscript{75}, which correspond to fixed IoU thresholds of 0.50 and 0.75, respectively, reflecting specific preferences for localization precision.
As shown in \Cref{tab:mAP50andmAP75}, PCR achieves the best estimate for both mAP\textsubscript{50} and mAP\textsubscript{75}, consistently outperforming all baseline methods.

\newlength{\defaultcolumnsep}
\newlength{\defaultintextsep}
\setlength{\defaultcolumnsep}{\columnsep}
\setlength{\defaultintextsep}{\intextsep}
\setlength{\columnsep}{5pt} 
\setlength{\intextsep}{2pt plus 2pt minus 2pt} 
\vspace{4pt}
\paragraph{Combination of PCR and BoS.}
As discussed in \Cref{sec:consistency}, BoS~\cite{yang2024bos} and our proposed PCR capture different aspects of consistency.
As combining complementary components often leads to improved performance, evidenced by PCR incorporating consistency and reliability, we evaluate whether combining BoS and PCR yields further improvements.
\Cref{tab:BoS+PCR} compares BoS, PCR, and their combination, where we perform linear regression using scores from both methods for the combination.
We observe that the combi-
\begin{wrapfigure}{r}{0.44\columnwidth}
    \centering
    \includegraphics[width=0.44\columnwidth]{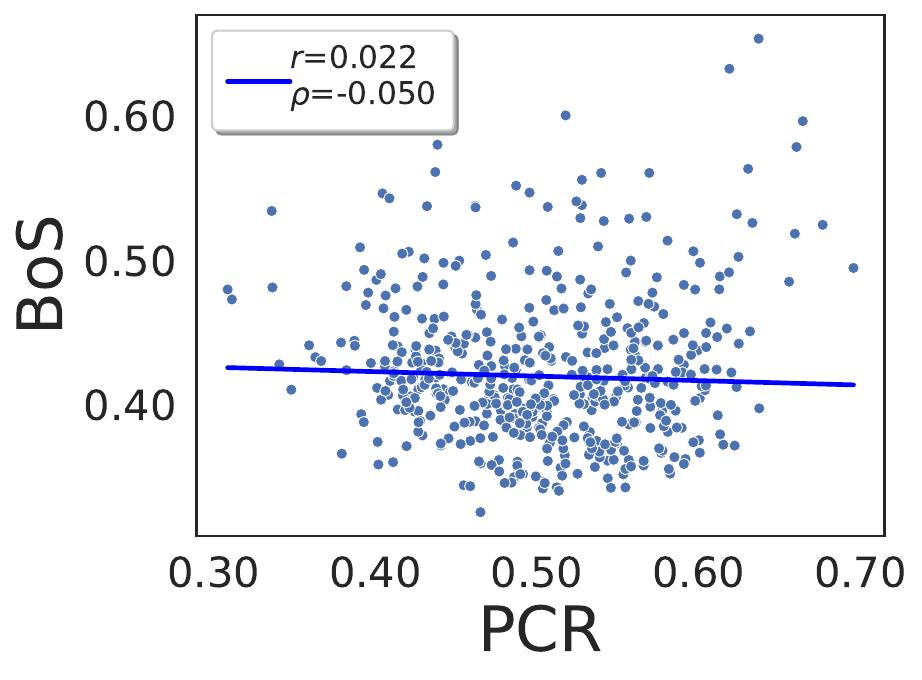}
    \vspace{-10pt}
    \caption{Consistency scores of PCR and BoS~\cite{yang2024bos}.}
    \label{fig:consistency_diff}
\end{wrapfigure}
nation outperforms individual methods, suggesting that BoS and PCR capture distinct yet complementary aspects of consistency.
This is further supported \Cref{fig:consistency_diff}, which shows that the consistency scores defined in BoS and those in PCR exhibit low correlation.
\setlength{\columnsep}{\defaultcolumnsep} 
\setlength{\intextsep}{\defaultintextsep} 

\begin{table}[t]
    \centering
    \renewcommand{\arraystretch}{1.1}
    \scalebox{0.8}{
    \begin{tabular}{l|cc|cc|c}
            \toprule
             Meta-dataset& \multicolumn{4}{c|}{\text{Corruption-based (Ours)}} & \multirow{3}{*}{\makecell{Avg. \\ RMSE}}\\
            \cmidrule{1-5}
            \multirow{2}{*}{\text{Detector}} & \multicolumn{2}{c|}{\textbf{RetinaNet}} & \multicolumn{2}{c|}{\textbf{Faster R-CNN}} &  \\
            & ResNet-50 & Swin-T & ResNet-50 & Swin-T &\\
            \midrule
            BoS~\cite{yang2024bos} & 13.50 & 5.18 & 10.32 & 6.85 & 8.96 \\
            PCR (Ours) & 6.57 & 4.26 & 7.23 & 4.70 & 5.69 \\
            \midrule
            PCR + BoS & \textbf{6.29} & \textbf{3.80} & \textbf{6.31} & \textbf{4.18}  & \textbf{5.15} \\
            \bottomrule
        \end{tabular}
        }
    \caption{Performance of the combined BoS~\cite{yang2024bos} and PCR in vehicle detection on the corruption-based meta-dataset. A detector trained on the COCO training set is used, and the average RMSE $(\%)$ across 10 datasets is reported.}
    \label{tab:BoS+PCR}
\end{table}

\section{Conclusion}
\label{sec:conclusion}

In this paper, we develop an AutoEval framework for object detection.
We propose Prediction Consistency and Reliability (PCR) as an AutoEval method for object detection, which leverages the relationships between pre- and post-NMS boxes to estimate detection performance without ground-truth labels.
PCR captures both localization and classification aspects through consistency and reliability measures conditioned on confidence scores.
To enable realistic and scalable evaluation, we construct a meta-dataset using image corruptions of varying severity. We hope that our proposed method and meta-dataset provide a solid foundation for future research on AutoEval for object detection.

\section*{Acknowledgements}
This work was partially supported by the National Research Foundation of Korea (NRF) grant funded by the Ministry of Science and ICT (MSIT) of the Korean government (RS2024-00341749), and Institute of Information \& Communications Technology Planning \& Evaluation (IITP) grant funded by MSIT (RS-2022-II220124, RS-2023-00259934, RS-2025-02283048).

{
    \small
    \bibliographystyle{ieeenat_fullname}
    \bibliography{main}
}

\onecolumn
   {
   \newpage
        \centering
        \Large
        \textbf{\thetitle}\\
        \vspace{0.5em}Supplementary Material \\
        \vspace{1.0em}
   }

\appendix
\numberwithin{table}{section}
\numberwithin{figure}{section}
\numberwithin{equation}{section}

\section{Experimental Details}

\noindent\textbf{Detectors.}
We compare our proposed PCR with baseline AutoEval methods across both one-stage and two-stage detectors with four backbone--head combinations: RetinaNet~\cite{lin2017focallossdenseobject} and Faster R-CNN~\cite{ren2015fasterrcnnrealtimeobject} paired with ResNet-50 (R50)~\cite{he2016deep} and Swin Transformer (Swin)~\cite{liu2021swin} backbones.

\vspace{4pt}
\noindent\textbf{Training.}
Backbones are pretrained on ImageNet-1k, and detectors are trained for a \(3\times\) schedule (36 epochs) using the ``car''-class subset of COCO for vehicle detection and CrowdHuman for pedestrian detection.
We use stochastic gradient descent (SGD) with an initial learning rate of \(10^{-3}\), weight decay of \(10^{-4}\), and momentum of \(0.9\), employing synchronized SGD over four GPUs with a total batch size of 8 images (\ie, 2 images per GPU).
Images are resized so that the shorter side is at most 800 pixels and the longer side does not exceed 1333 pixels, with horizontal flipping as the only data augmentation.

\vspace{4pt}
\noindent\textbf{AutoEval.}
For AutoEval, test images are resized to a fixed resolution of \(800 \times 1333\).

\vspace{4pt}
\noindent\textbf{Computational Resources.}
All experiments are conducted on a server with two Intel(R) Xeon(R) Gold 6226R \@ 2.90~GHz CPUs and eight NVIDIA RTX A5000 GPUs. Training is performed on four GPUs using MMDetection~\cite{chen2019mmdetection}, and AutoEval is run on a single GPU using PyTorch~1.13.1 with CUDA~11.6.

\vspace{4pt}
\noindent\textbf{Replication of BoS Results.}
All BoS~\cite{yang2024bos} results presented in our paper are based on our replication, with modifications made to the publicly released code.
We found that the experimental settings described in the paper slightly differ from those implemented in the code, and certain parts of the experiments are not directly reproducible.
Therefore, we modified the code to match the reported performance as closely as possible.

\section{Additional Experimental Results}

\subsection{Robustness to Varying Set Sizes}

To assess the robustness of our proposed PCR across different set sizes, we vary:
(a) the \textbf{meta-dataset size}, the number of datasets in the meta-dataset determined by the product of the number of corruptions and the number of severity levels;
(b) the \textbf{sample set size}, the number of images sampled from a source dataset to form each dataset in the meta-dataset; and
(c) the \textbf{test set size}, the number of test images sampled from the held-out source dataset to compute the AutoEval score.
As shown in \Cref{fig:set_size}, increasing the meta-dataset and sample set sizes provides richer information, leading to lower RMSE in general.
For the test set size, performance remains stable when the number of images exceeds 150.
\begin{figure}[ht]
    \centering
    \hspace*{\fill}
    \begin{minipage}[b]{0.255\textwidth}
    \centering
    \includegraphics[width=\columnwidth]{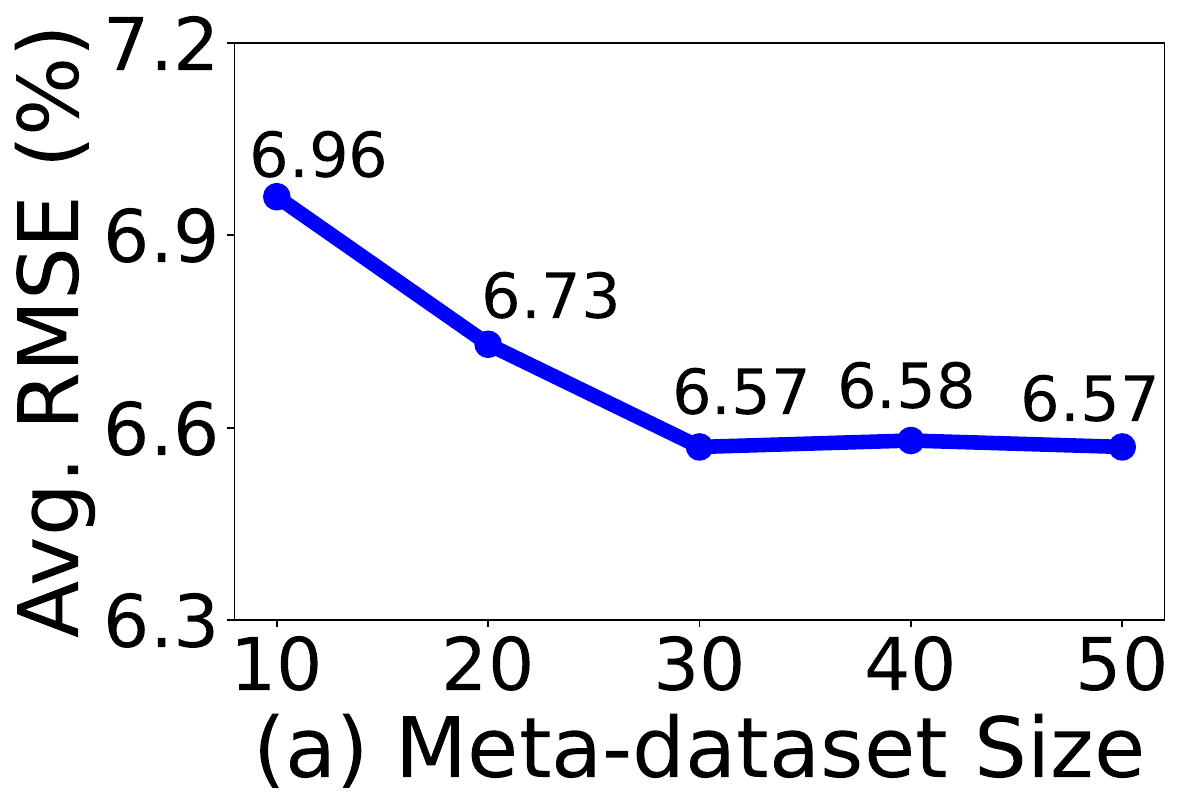}
    \label{fig:metaset_size}
    \end{minipage}
    \hspace*{\fill}
    \begin{minipage}[b]{0.245\textwidth}
    \centering
    \includegraphics[width=\columnwidth]{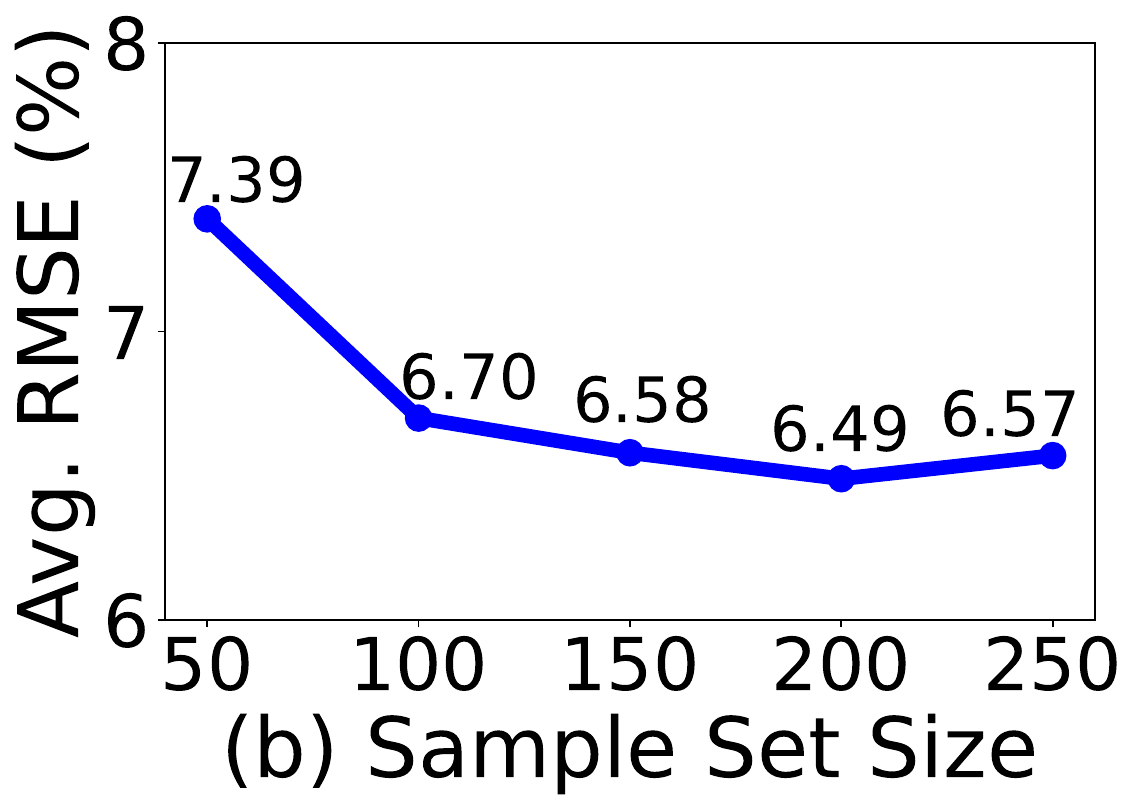}
    \label{fig:sample_size}
    \end{minipage}
    \hspace*{\fill}
    \begin{minipage}[b]{0.260\textwidth}
        \includegraphics[width=\columnwidth]{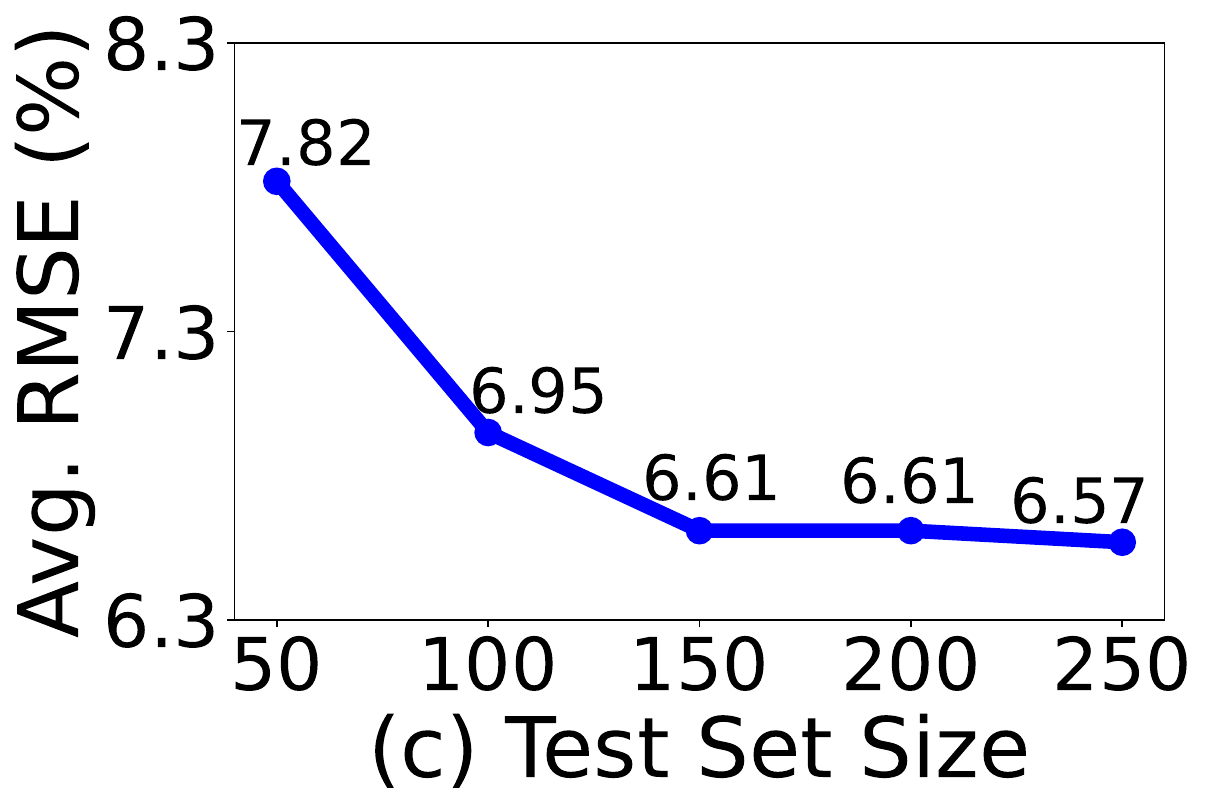}
        \label{fig:testset_size}
    \end{minipage}
    \hspace*{\fill}
  \vspace{-10pt}
    \caption{Effect of varying set sizes on the performance of PCR for vehicle detection using RetinaNet+R50 on the corruption-based meta-dataset: (a) meta-dataset, (b) sample set, and (c) test set.}
  \vspace{-5pt}
    \label{fig:set_size}
\end{figure}

\newpage

\subsection{Hyperparameter Analysis}

We conduct experiments with varying hyperparameters for both tuning and sensitivity analysis.
Unless otherwise specified, all experiments use RetinaNet~\cite{lin2017focallossdenseobject} with a ResNet-50~\cite{he2016deep} backbone evaluated on the corruption-based meta-dataset, with each analysis performed using the scores to which the hyperparameter applies.
Performance is averaged over vehicle and pedestrian detection;
for example,
from \Cref{tab:car_detection,tab:person_detection},
\(11.86 = (13.50 + 10.23) / 2\) and \(4.77 = (6.57 + 2.96) / 2\) correspond to the performance of BoS~\cite{yang2024bos} and our proposed PCR, respectively.

\vspace{4pt}
\noindent\textbf{Confidence Threshold} is a reference value used to assess whether a given prediction is considered confident, employed both in the consistency and reliability scores.
\Cref{fig:hyperparameter}(a) illustrates the effect of varying the confidence threshold on mAP estimation.
Based on these results, we set \(c=0.4\) as the confidence threshold throughout the experiments.

\vspace{4pt}
\noindent\textbf{Floor Value} is used in the reliability score, which provides a lower bound to low-confidence predictions, ensuring a minimal contribution from all predictions.
\Cref{fig:hyperparameter}(b) illustrates the effect of varying the floor value on mAP estimation, averaged over vehicle and pedestrian detection. 
Based on these results, we set \(\alpha=0.1\) as the floor value throughout the experiments.

\vspace{4pt}
\noindent\textbf{Scale Parameters} control the sensitivity of the sigmoid function, which modulates the contribution of confidence to the consistency and reliability scores.
For consistency, the scale parameter \(k_C\) is negative because the consistency score is negatively correlated with mAP, while for reliability, the scale parameter \(k_R\) is positive because the reliability score is positively correlated with mAP.
\Cref{fig:hyperparameter}(c--d) illustrates the effect of varying the scale parameters on mAP estimation, averaged over vehicle and pedestrian detection.
Based on these results, we set \(k_C=-10\) and \(k_R=20\) as the scale parameters throughout the experiments.

\vspace{4pt}
\noindent\textbf{Hyperparameter Sensitivity Analysis.}
While hyperparameter tuning provides additional performance gains for PCR, it consistently outperforms BoS across all hyperparameter settings; the performance in \Cref{fig:hyperparameter} remains below 11.86 in all cases.
This indicates that the superior performance of PCR is robust to hyperparameter choices.

\begin{figure*}[ht]
  \begin{minipage}[b]{0.25\textwidth}
    \centering
    \includegraphics[width=\columnwidth]{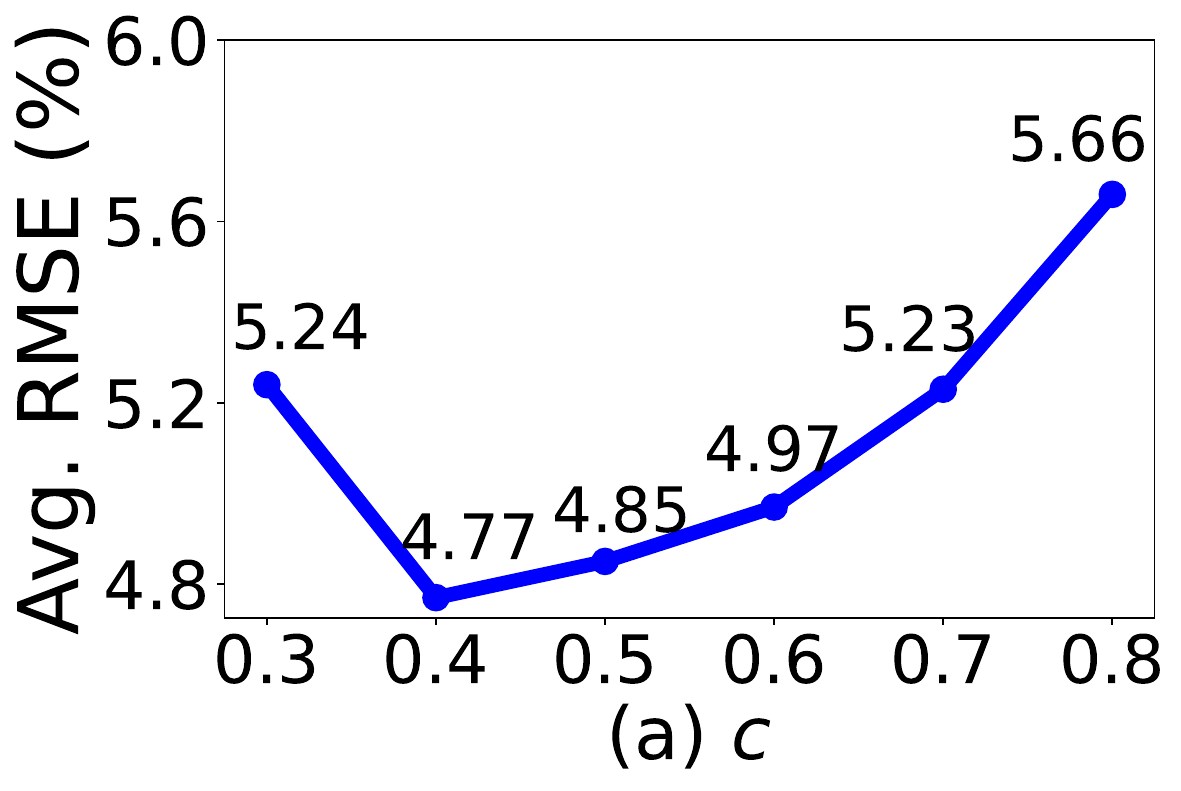}
    \label{fig:confidence_threshold}
  \end{minipage}%
  \begin{minipage}[b]{0.235\textwidth}
    \centering
    \includegraphics[width=\columnwidth]{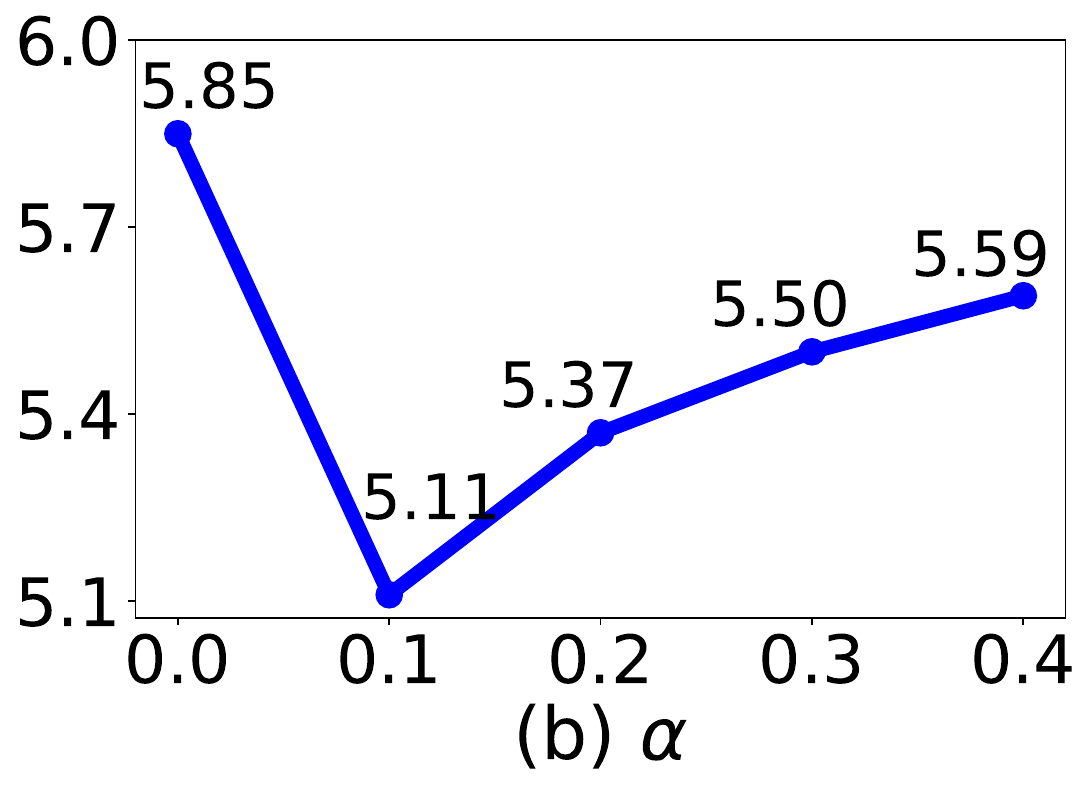}
    \label{fig:confidence_threshold2}
  \end{minipage}
  \begin{minipage}[b]{0.245\textwidth}
    \centering
    \includegraphics[width=\columnwidth]{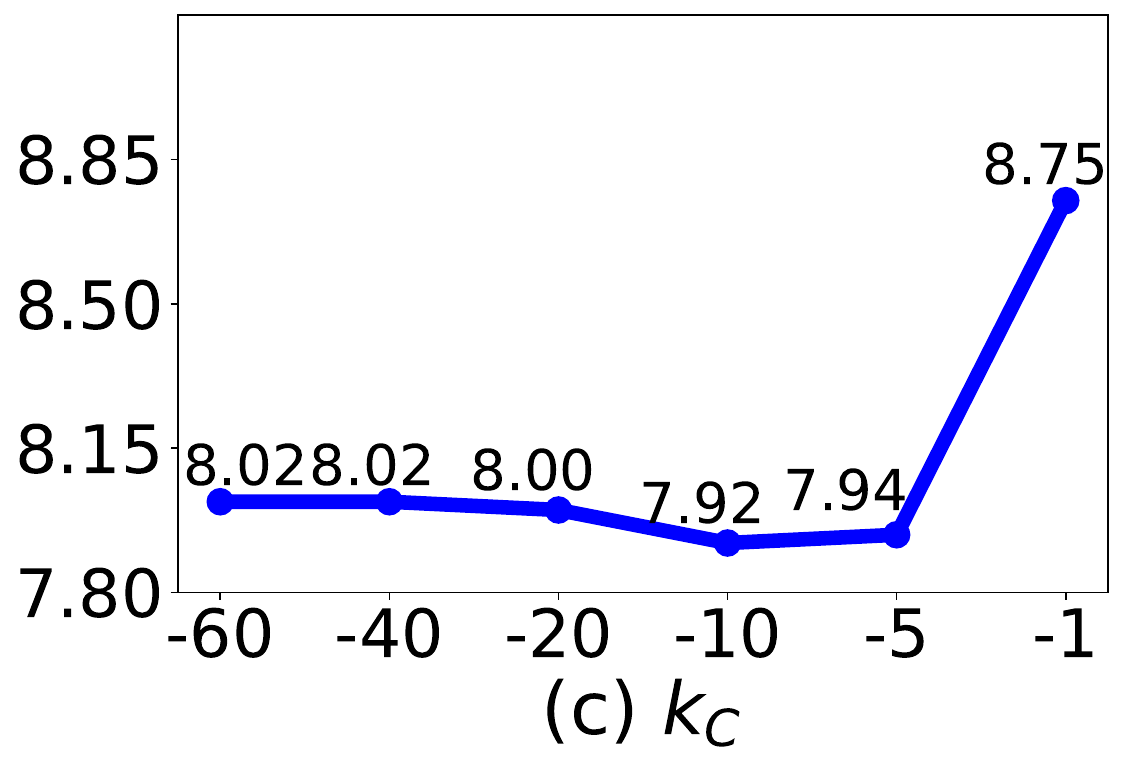}
    \label{fig:slope_consistency}
  \end{minipage}  
    \begin{minipage}[b]{0.235\textwidth}
    \centering
    \includegraphics[width=\columnwidth]{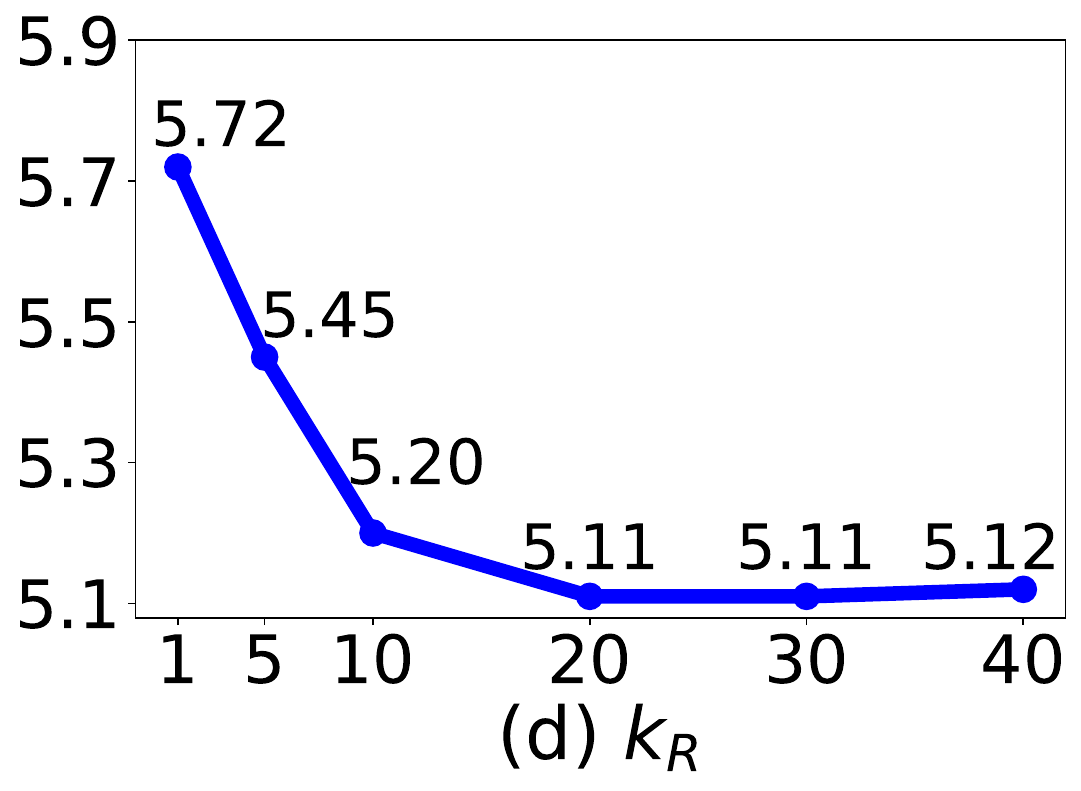}
    \label{fig:slope_reliability}
  \end{minipage}%
  \vspace{-10pt}
  \caption{
  Hyperparameter tuning results of PCR for:
  (a) confidence threshold \(c\),
  (b) floor value \(\alpha\) in \(\sigma_R\),
  (c) scale parameter \(k_C\) in \(\sigma_C\), and
  (d) scale parameter \(k_R\) in \(\sigma_R\).
  Performance is measured for vehicle detection using RetinaNet+R50 on the corruption-based meta-dataset, except that (a) averages the performance across vehicle and pedestrian detection.
  }
  \label{fig:hyperparameter}
\end{figure*}

\subsection{Multi-Class Detection}

We further extend our experiments to the multi-class detection setting to assess whether AutoEval for object detection generalizes to broader scenarios.
In this setting, detectors are trained on the COCO training set~\cite{lin2014microsoft}.
For the meta-dataset, we use five source datasets: Cityscapes~\cite{cordts2015cityscapes}, COCO~\cite{lin2014microsoft}, ExDark~\cite{loh2019getting}, KITTI~\cite{geiger2013vision}, and Self-driving~\cite{Kaggle_SelfDriving_2020}.
As shown in \Cref{tab:multi_class}, PCR outperforms BoS~\cite{yang2024bos} on average across four detectors, demonstrating superior performance in multi-class detection scenarios.
\begin{table}[ht]
    \centering
    \renewcommand{\arraystretch}{1.1}
    \scalebox{0.8}{
    \begin{tabular}{l|cc|cc|c}
            \toprule
             Meta-dataset& \multicolumn{4}{c|}{\text{Corruption-based (Ours)}} & \multirow{3}{*}{\makecell{Avg. \\ RMSE}}\\
            \cmidrule{1-5}
            \multirow{2}{*}{\text{Detector}} & \multicolumn{2}{c|}{\textbf{RetinaNet}} & \multicolumn{2}{c|}{\textbf{Faster R-CNN}} &  \\
            & ResNet-50 & Swin-T & ResNet-50 & Swin-T &\\
            \midrule
            BoS~\cite{yang2024bos} & 10.04 & \textbf{5.07} & 9.38 & \textbf{7.34} & 7.96 \\
            PCR (Ours) & \textbf{7.07} & 6.95 & \textbf{8.92} & 8.07 & \textbf{7.75}\\
            
            \bottomrule
        \end{tabular}
        }
    \caption{Comparison of AutoEval methods for multi-class detection across four detectors on the corruption-based meta-dataset. The best result is highlighted in \textbf{bold}.}
    \label{tab:multi_class}
\end{table}

\subsection{Piecewise Regression}

Object detectors may encounter challenging scenarios with extremely low mAP.
As illustrated in \Cref{fig:IoU_corr}---and comprehensively shown in \Cref{fig:vehicle_corr,fig:Pedestrian_corr}---our proposed consistency and reliability scores often exhibit different trends in the low-mAP regime compared to the high-mAP regime, suggesting that applying a single linear regression may not adequately capture the relationship across a wide range of detection performance.
Instead, a piecewise regression approach may yield more accurate AutoEval results in such challenging scenarios, where AutoEval for low mAP requires a separate estimation from high mAP.
To simulate this, we split the meta-dataset into low- and high-mAP regimes:
datasets with mAP below 5\% are used for AutoEval in the low-mAP regime, while those with mAP above 5\% are used for AutoEval in the high-mAP regime.
To assess AutoEval in the low-mAP regime, we extract the bottom 10\% of images in the original test sets, ranked by mAP, as low-mAP test sets.
From \Cref{fig:IoU_corr}, we hypothesize that only a few datasets with low mAP deviate from the general trend, implying that AutoEval in the high-mAP regime provides more accurate performance estimation for the original test sets, which generally exhibit mAP higher than 5\%.

\Cref{tab:split} reports AutoEval performance for vehicle detection across four detectors.
While PCR consistently outperforms BoS~\cite{yang2024bos}, piecewise regression further improves AutoEval performance, suggesting that AutoEval scores might exhibit nonlinear correlations with detection performance.
\Cref{fig:IoU_corr_split} visualizes the estimated correlation between each AutoEval score and mAP when using RetinaNet~\cite{lin2017focallossdenseobject} with a ResNet-50~\cite{he2016deep} backbone, under both single linear regression and piecewise regression, demonstrating that separate low- and high-mAP models provide a better fit to the meta-dataset.
\begin{table}[ht]
    \centering
    \renewcommand{\arraystretch}{1.1}
    \scalebox{0.8}{
    \begin{tabular}{l|l|c|cc|cc|c}
            \toprule
            \multirow{2}{*}{\text{Test Set}} & \multirow{2}{*}{\text{Meta-dataset}} & \multirow{2}{*}{\text{Method}} & \multicolumn{2}{c|}{\textbf{RetinaNet}} & \multicolumn{2}{c|}{\textbf{Faster R-CNN}} & \multirow{2}{*}{\makecell{Avg. \\ RMSE}} \\
            & & & ResNet-50 & Swin-T & ResNet-50 & Swin-T & \\
            \midrule
            \multirow{4}{*}{\text{All}} & \multirow{2}{*}{\text{All}} & BoS~\cite{yang2024bos} & 13.50 & 5.18 & 10.32 & 6.85 & 8.96 \\
            & & PCR (Ours) & \textbf{6.57} & \textbf{4.26} & \textbf{7.23} & \textbf{4.70} & \textbf{5.69} \\
            \cmidrule{2-8}
            & \multirow{2}{*}{\text{High mAP}} & BoS~\cite{yang2024bos} & 11.65 & 5.09 & 8.22 & 6.55 & 7.88 \\
            & & PCR (Ours) & \textbf{5.87} & \textbf{4.31} & \textbf{5.95} & \textbf{4.51} & \textbf{5.16} \\
            \midrule
            \multirow{4}{*}{\text{Low mAP}} & \multirow{2}{*}{\text{All}} & BoS~\cite{yang2024bos} & 13.52 & 17.00 & \textbf{10.21} & 14.74 & 13.86 \\
            & & PCR (Ours) & \textbf{12.91} & \textbf{13.51} & 13.19 & \textbf{12.64} & \textbf{13.06} \\
            \cmidrule{2-8}
            & \multirow{2}{*}{\text{Low mAP}} & BoS~\cite{yang2024bos} & 5.77 &7.92 & 5.36 & 7.76 & 6.70 \\
            & & PCR (Ours) & \textbf{5.40} & \textbf{5.72} & \textbf{5.00} & \textbf{6.41} & \textbf{5.63} \\
            \bottomrule
        \end{tabular}
        }
    \caption{Comparison of AutoEval methods for standard and difficult vehicle detection across four detectors on corruption-based meta-dataset variants: the full meta-dataset (``All''), the subset with \(\text{mAP} \geq 5\) (``High mAP''), and the subset with \(\text{mAP} < 5\) (``Low mAP''). The best result is highlighted in \textbf{bold}.}
    \label{tab:split}
\end{table}

\begin{figure}[ht]
    \centering
    \makebox[\textwidth]{%
        \hspace*{\fill}
        \begin{minipage}[t]{0.32\textwidth}\vspace{0pt}
            \includegraphics[width=\textwidth]{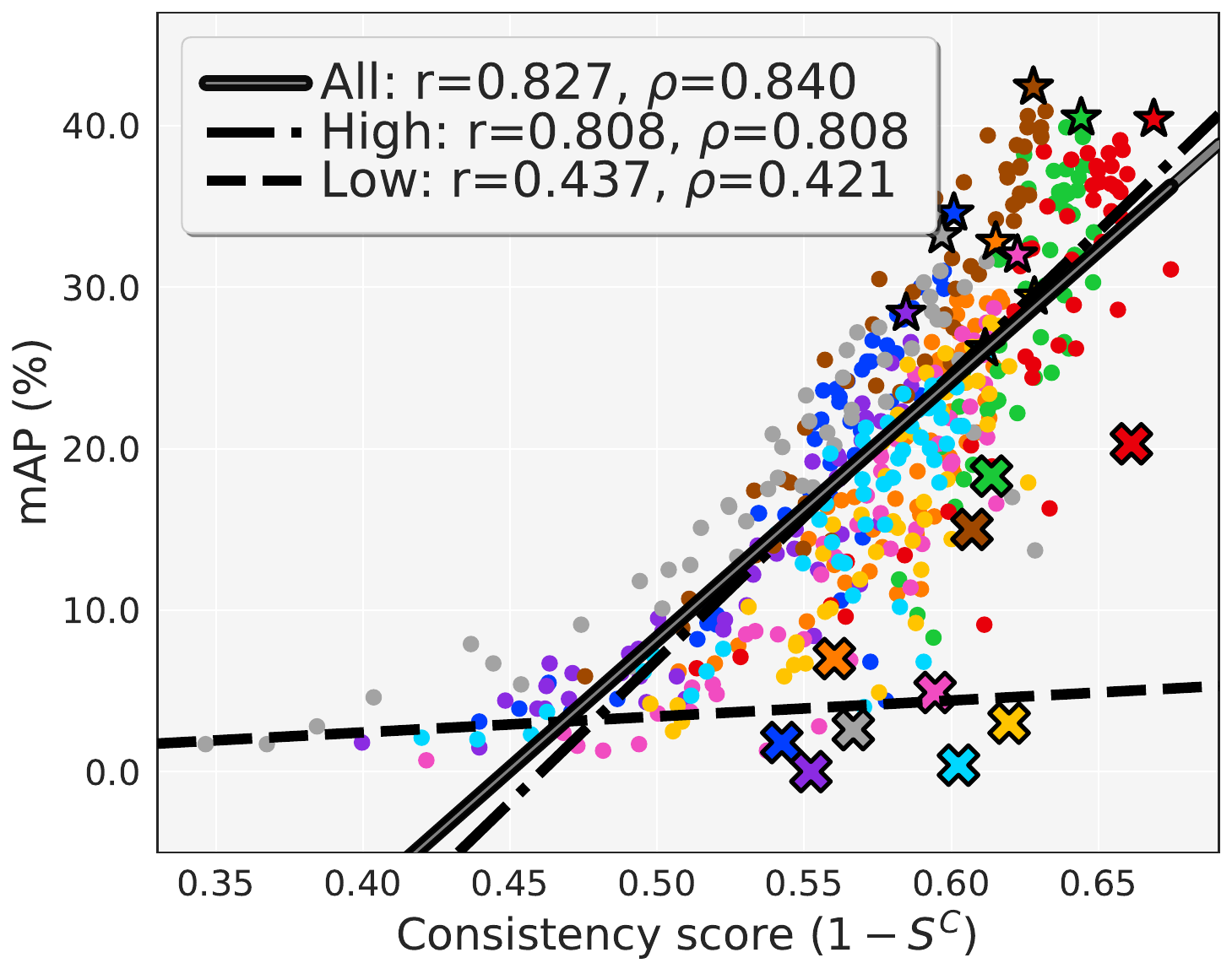}
        \end{minipage}%
        \hspace*{\fill}
        \begin{minipage}[t]{0.32\textwidth}\vspace{0pt}
            \includegraphics[width=\textwidth]{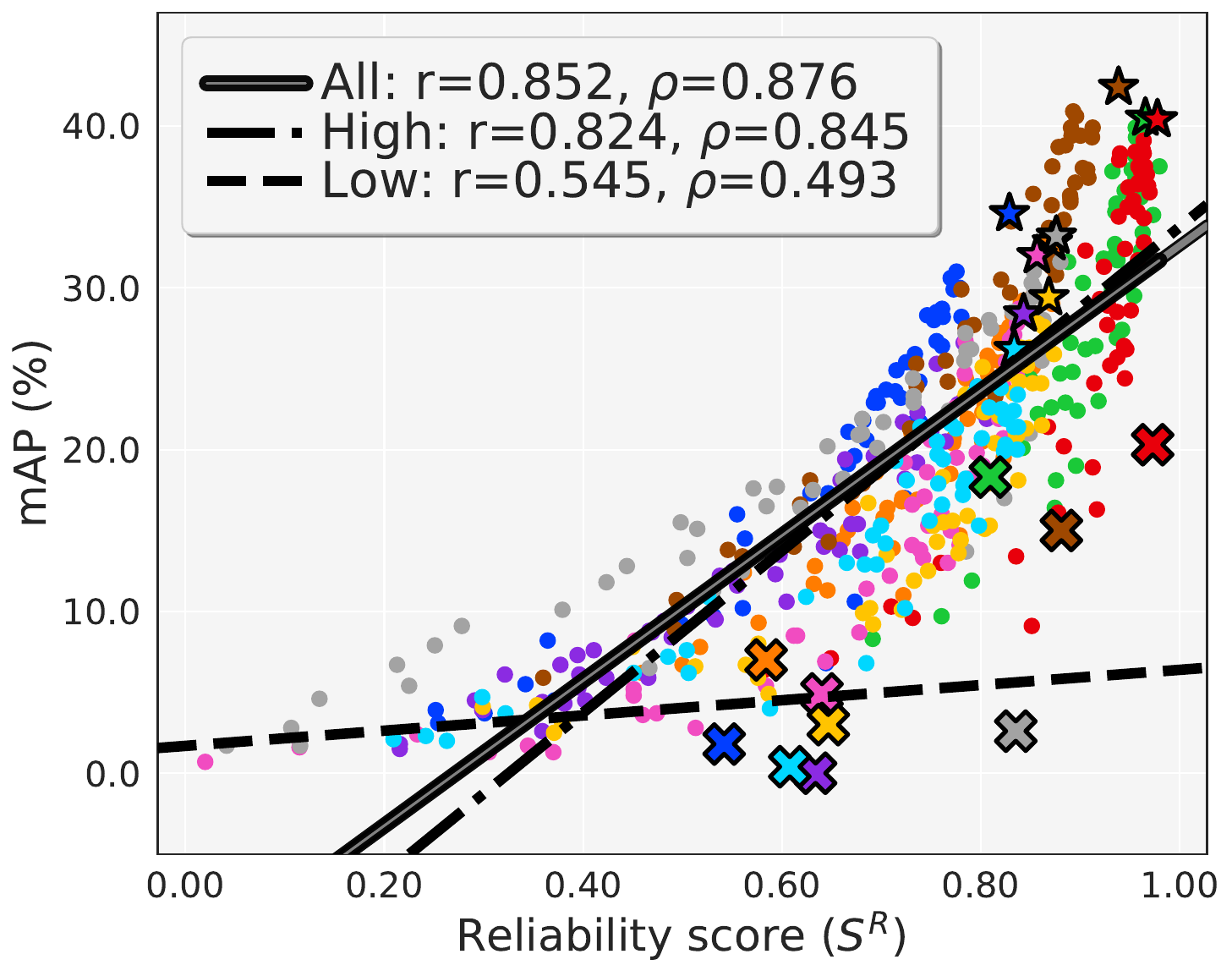}
        \end{minipage}%
        \hspace*{\fill}
        \begin{minipage}[t]{0.16\textwidth}\vspace{0pt}
            \includegraphics[width=\textwidth]{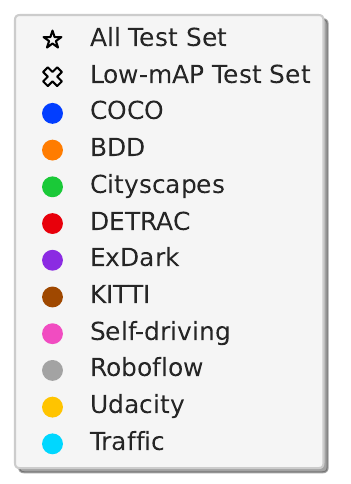}
        \end{minipage}%
        \hspace*{\fill}
    }
    \caption{Correlation analysis of the consistency and reliability scores used in our proposed PCR for \textbf{vehicle detection} using \textbf{RetinaNet with ResNet-50} as the detector.
    Linear regression of mAP on each AutoEval score is performed on the full meta-dataset (``All''; solid line), the subset with \(\text{mAP} \geq 5\) (``High''; dash-dotted line), and the subset with \(\text{mAP} < 5\) (``Low''; dashed line). 
    The full test sets (\(\star\)), their bottom-10\% subsets with low mAP (\(\times\)), and datasets in the meta-dataset (\(\circ\)) are indicated by different markers, where each color corresponds to a distinct source dataset.
    Correlation is measured by Pearson's correlation coefficient (\(r\)) and Spearman's rank correlation coefficient (\(\rho\)).}
    \label{fig:IoU_corr_split}
\end{figure}

\newpage

\subsection{Per-Test Set AutoEval Results}

Each cell in \Cref{tab:car_detection,tab:person_detection} reports the average RMSE between the estimated mAP and the true mAP across source datasets.
We provide AutoEval results for each test set in
\Cref{tab:car_r50_retina_bos,tab:car_swin_retina_bos,tab:car_r50_faster_bos,tab:car_swin_faster_bos,tab:car_r50_retina_ccc,tab:car_swin_retina_ccc,tab:car_r50_faster_ccc,tab:car_swin_faster_ccc,tab:person_r50_retina_bos,tab:person_swin_retina_bos,tab:person_r50_faster_bos,tab:person_swin_faster_bos,tab:person_r50_retina_ccc,tab:person_swin_retina_ccc,tab:person_r50_faster_ccc,tab:person_swin_faster_ccc},
where 
\Cref{tab:car_r50_retina_bos,tab:car_swin_retina_bos,tab:car_r50_faster_bos,tab:car_swin_faster_bos}
present results for vehicle detection on the augmentation-based meta-dataset,
\Cref{tab:car_r50_retina_ccc,tab:car_swin_retina_ccc,tab:car_r50_faster_ccc,tab:car_swin_faster_ccc}
are results for vehicle detection on the corruption-based meta-dataset,
\Cref{tab:person_r50_retina_bos,tab:person_swin_retina_bos,tab:person_r50_faster_bos,tab:person_swin_faster_bos}
are results for pedestrian detection on the augmentation-based meta-dataset, and 
\Cref{tab:person_r50_retina_ccc,tab:person_swin_retina_ccc,tab:person_r50_faster_ccc,tab:person_swin_faster_ccc}
are results for pedestrian detection on the corruption-based meta-dataset.
Overall, our proposed PCR outperforms all baselines across most experimental settings, with particularly large gains observed on the corruption-based meta-dataset.
In contrast, while BoS performs competitively on the augmentation-based meta-dataset, its performance often degrades substantially on the corruption-based meta-dataset.

\subsection{Correlation Analysis}

Correlation can be used to assess the quality of AutoEval scores, as a strong positive or negative correlation indicates that the score accurately captures model performance.
We visualize the correlation between each AutoEval score and mAP in \Cref{fig:vehicle_corr,fig:Pedestrian_corr}, comparing the BoS score~\cite{yang2024bos} and the consistency and reliability scores used in our proposed PCR.
Our consistency and reliability scores generally exhibit stronger correlations with mAP than the BoS score, demonstrating their effectiveness.

\newpage

\begin{table*}[ht]
\centering
\renewcommand{\arraystretch}{1.1}
\resizebox{\textwidth}{!}{%
\begin{tabular}{l|cccccccccc|c}
\toprule
\multirow{2}{*}{Method} & COCO & BDD & Cityscapes & DETRAC & ExDark & KITTI & Self-driving & Roboflow & Udacity & Traffic & \multirow{2}{*}{\makecell{Avg. \\ RMSE}} \\
 & 34.6 & 32.8 & 40.5 & 40.4 & 28.4 & 42.4 & 32.0 & 33.2 & 29.4 & 26.2 & \\
\midrule
PS~\cite{hendrycks2018ps}  & \underline{1.70} & 7.67 & 3.84 & 12.23 & 6.47 & 11.60 & \underline{2.58} & 2.18 & \textbf{0.08} & \underline{0.19} & 4.85 \\
ES~\cite{saito2019semi} & 6.39 & 3.83 & 15.28 & 4.25 & \textbf{0.77} & 18.85 & 4.85 & 5.30 & \underline{2.02} & 3.59 & 6.51 \\
AC~\cite{guillory2021predicting}  & 6.88 & 5.04 & 10.55 & 44.09 & 2.89 & 15.37 & 4.23 & 2.31 & \underline{2.02} & 1.57 & 9.50 \\
ATC~\cite{arg2022leveragingunlabeleddatapredict} & 4.53 & 7.19 & 11.48 & \textbf{0.84} & 4.67 & 15.79 & 6.06 & \textbf{0.86} & 3.71 & \textbf{0.05} & 5.52 \\
BoS~\cite{yang2024bos} & \textbf{1.20} & \textbf{0.47} & \textbf{2.06} & 4.53 & 4.57 & \underline{7.71} & 2.80 & 2.85 & 4.06 & 0.82 & \underline{3.11} \\
\midrule
PCR & 5.57 & \underline{0.82} & \underline{3.83} & \underline{1.59} & \underline{2.70} & \textbf{6.94} & \textbf{1.26} & \underline{1.43} & 2.26 & 3.51 & \textbf{2.99} \\

\bottomrule
\end{tabular}%
}
\caption{Comparison of AutoEval methods for \textbf{vehicle detection} on the \textbf{augmentation-based meta-dataset}~\cite{yang2024bos}.
\textbf{RetinaNet with ResNet-50} trained on the ``car''-class subset of COCO is used as the detector, and its true mAP (\%) is reported below the name of each test set.
The best result for each test set is highlighted in \textbf{bold} and the second-best is \underline{underlined}.}\vspace{-15pt}
\label{tab:car_r50_retina_bos}
\end{table*}

\begin{table*}[ht]
\centering
\renewcommand{\arraystretch}{1.1}
\resizebox{\textwidth}{!}{%
\begin{tabular}{l|cccccccccc|c}
\toprule
\multirow{2}{*}{Method} & COCO & BDD & Cityscapes & DETRAC & ExDark & KITTI & Self-driving & Roboflow & Udacity & Traffic & \multirow{2}{*}{\makecell{Avg. \\ RMSE}} \\
 & 35.8 & 33.5 & 40.2 & 39.9 & 25.8 & 42.1 & 30.6 & 27.3 & 28.7 & 25.5 & \\
\midrule
PS~\cite{hendrycks2018ps} & 9.29 & 4.84 & 5.93 & 22.51 & 2.61 & 17.66 & 5.05 & \underline{0.42} & 2.16 & 3.38 & 7.38 \\
ES~\cite{saito2019semi} & 10.85 & \textbf{1.21} & 14.04 & \textbf{4.25} & \textbf{0.91} & 16.91 & \underline{1.59} & 1.19 & 1.60 & \underline{3.27} & \underline{5.58} \\
AC~\cite{guillory2021predicting} & 9.55 & 3.49 & 11.59 & 31.93 & 1.84 & 16.28 & 2.27 & \underline{0.42} & \textbf{0.87} & 3.49 & 8.17 \\
ATC~\cite{arg2022leveragingunlabeleddatapredict} & 8.85

 n  & 4.70 & 11.20 & 62.18 & 2.22 & 15.54 & 3.06 & \textbf{0.06} & \underline{1.04} & \textbf{3.13} & 11.20 \\
BoS~\cite{yang2024bos} & \textbf{2.78} & 5.57 & \textbf{2.45} & 8.88 & 10.29 & \textbf{4.23} & 10.30 & 13.87 & 10.59 & 7.90 & 7.69 \\
\midrule
PCR & \underline{8.68} & \underline{1.57} & \underline{4.86} & \underline{5.33} & \underline{1.03} & \underline{7.34} & \textbf{0.37} & 2.98 & 2.22 & 3.85 & \textbf{3.82} \\

\bottomrule
\end{tabular}%
}
\caption{Comparison of AutoEval methods for \textbf{vehicle detection} on the \textbf{augmentation-based meta-dataset}~\cite{yang2024bos}.
\textbf{RetinaNet with Swin Transformer} trained on the ``car''-class subset of COCO is used as the detector, and its true mAP (\%) is reported below the name of each test set.
The best result for each test set is highlighted in \textbf{bold} and the second-best is \underline{underlined}.}\vspace{-15pt}
\label{tab:car_swin_retina_bos}
\end{table*}

\begin{table*}[ht]
\centering
\renewcommand{\arraystretch}{1.1}
\resizebox{\textwidth}{!}{%
\begin{tabular}{l|cccccccccc|c}
\toprule
\multirow{2}{*}{Method} & COCO & BDD & Cityscapes & DETRAC & ExDark & KITTI & Self-driving & Roboflow & Udacity & Traffic & \multirow{2}{*}{\makecell{Avg. \\ RMSE}} \\
 & 36.0 & 32.2 & 40.5 & 36.2 & 25.9 & 38.9 & 29.4 & 29.4 & 27.8 & 27.7 & \\
\midrule
PS~\cite{hendrycks2018ps} & 8.85 & 5.96 & 8.48 & 2.36 & 2.99 & \underline{4.14} & 2.63 & 8.02 & 5.01 & 1.10 & 4.95 \\
ES~\cite{saito2019semi} & 13.57 & \underline{4.28} & 12.64 & \textbf{0.52} & \textbf{0.26} & 8.67 & \textbf{0.16} & 12.44 & \underline{3.32} & 1.58 & 5.74 \\
AC~\cite{guillory2021predicting} & 11.79 & 5.63 & 11.15 & 7.83 & \underline{0.50} & 8.53 & 1.40 & 6.47 & \textbf{2.06} & 1.42 & 5.68 \\
ATC~\cite{arg2022leveragingunlabeleddatapredict} & 8.80 & 7.66 & 10.60 & 4.15 & 2.66 & 5.63 & 2.16 & \underline{3.30} & 3.64 & \textbf{0.28} & 4.89 \\
BoS~\cite{yang2024bos} & \underline{6.93} & 7.23 & \textbf{5.01} & 2.32 & 4.11 & \textbf{0.66} & \underline{0.95} & \textbf{2.31} & 3.50 & \underline{0.30} & \textbf{3.33} \\
\midrule
PCR & \textbf{4.94} & \textbf{4.12} & \underline{7.90} & \underline{1.02} & 3.57 & 6.38 & 3.29 & 3.57 & 4.28 & 0.73 & \underline{3.98} \\
\bottomrule
\end{tabular}%
}
\caption{Comparison of AutoEval methods for \textbf{vehicle detection} on the \textbf{augmentation-based meta-dataset}~\cite{yang2024bos}.
\textbf{Faster R-CNN with ResNet-50} trained on the ``car''-class subset of COCO is used as the detector, and its true mAP (\%) is reported below the name of each test set.
The best result for each test set is highlighted in \textbf{bold} and the second-best is \underline{underlined}.}\vspace{-15pt}
\label{tab:car_r50_faster_bos}
\end{table*}

\begin{table*}[ht]
\centering
\renewcommand{\arraystretch}{1.1}
\resizebox{\textwidth}{!}{%
\begin{tabular}{l|cccccccccc|c}
\toprule
\multirow{2}{*}{Method} & COCO & BDD & Cityscapes & DETRAC & ExDark & KITTI & Self-driving & Roboflow & Udacity & Traffic & \multirow{2}{*}{\makecell{Avg. \\ RMSE}} \\
 & 38.7 & 34.4 & 42.6 & 36.4 & 31.6 & 44.4 & 31.4 & 31.9 & 29.2 & 27.3 & \\
\midrule
PS~\cite{hendrycks2018ps} & 7.16 & 4.39 & 6.06 & 18.93 & 2.15 & 7.42 & 2.76 & 4.64 & 6.95 & 2.33 & 6.28 \\
ES~\cite{saito2019semi} & 12.75 & 7.17 & 9.34 & \underline{7.27} & \underline{1.32} & 8.97 & \textbf{0.39} & 7.58 & 7.90 & \underline{1.88} & 6.46 \\
AC~\cite{guillory2021predicting} & 8.34 & 3.79 & 8.56 & 19.68 & 2.68 & 11.26 & 2.45 & 4.40 & \underline{4.49} & 4.05 & 6.97 \\
ATC~\cite{arg2022leveragingunlabeleddatapredict} & 7.03 & 3.63 & 7.62 & 23.44 & 1.61 & 9.86 & 3.30 & \underline{3.18} & 5.10 & 3.78 & 6.86 \\
BoS~\cite{yang2024bos} & \textbf{0.42} & \underline{1.04} & \underline{5.55} & \textbf{3.65} & 2.32 & \textbf{7.40} & 6.66 & 13.30 & 9.19 & 6.19 & \underline{5.57} \\
\midrule
PCR & \underline{6.09} & \textbf{0.08} & \textbf{1.57} & 7.95 & \textbf{1.24} & \underline{8.08} & \underline{0.56} & \textbf{3.09} & \textbf{3.24} & \textbf{1.66} & \textbf{3.36} \\

\bottomrule
\end{tabular}%
}
\caption{Comparison of AutoEval methods for \textbf{vehicle detection} on the \textbf{augmentation-based meta-dataset}~\cite{yang2024bos}.
\textbf{Faster R-CNN with Swin Transformer} trained on the ``car''-class subset of COCO is used as the detector, and its true mAP (\%) is reported below the name of each test set.
The best result for each test set is highlighted in \textbf{bold} and the second-best is \underline{underlined}.}\vspace{-15pt}
\label{tab:car_swin_faster_bos}
\end{table*}

\begin{table*}[ht]
\centering
\renewcommand{\arraystretch}{1.1}
\resizebox{\textwidth}{!}{%
\begin{tabular}{l|cccccccccc|c}
\toprule
\multirow{2}{*}{Method} & COCO & BDD & Cityscapes & DETRAC & ExDark & KITTI & Self-driving & Roboflow & Udacity & Traffic & \multirow{2}{*}{\makecell{Avg. \\ RMSE}} \\
 & 34.6 & 32.8 & 40.5 & 40.4 & 28.4 & 42.4 & 32.0 & 33.2 & 29.4 & 26.2 & \\
\midrule
PS~\cite{hendrycks2018ps} & \underline{11.71} & 14.90 & \underline{14.48} & \underline{9.31} & \underline{3.97} & \underline{21.11} & 10.68 & 11.59 & \underline{8.03} & 7.18 & \underline{11.30} \\
ES~\cite{saito2019semi} & 14.23 & 12.59 & 21.40 & 31.70 & 7.24 & 23.48 & 11.13 & 12.83 & 8.74 & 5.39 & 14.87 \\
AC~\cite{guillory2021predicting} & 15.11 & 13.45 & 20.87 & 20.82 & 7.22 & 24.02 & 11.88 & 12.72 & 9.78 & 6.44 & 14.23 \\
ATC~\cite{arg2022leveragingunlabeleddatapredict} & 14.05 & 14.59 & 20.65 & 18.41 & 6.51 & 24.19 & 12.75 & \underline{11.56} & 10.78 & 7.46 & 14.10 \\
BoS~\cite{yang2024bos} & 13.77 & \underline{12.16} & 21.94 & 20.59 & 7.01 & 23.06 & \underline{10.49} & 12.46 & 8.16 & \underline{5.32} & 13.50 \\
\midrule
PCR & \textbf{9.96} & \textbf{5.72} & \textbf{9.54} & \textbf{8.69} & \textbf{3.57} & \textbf{13.77} & \textbf{5.11} & \textbf{7.22} & \textbf{1.67} & \textbf{0.44} & \textbf{6.57} \\

\bottomrule
\end{tabular}%
}
\caption{Comparison of AutoEval methods for \textbf{vehicle detection} on the \textbf{corruption-based meta-dataset}.
\textbf{RetinaNet with ResNet-50} trained on the ``car''-class subset of COCO is used as the detector, and its true mAP (\%) is reported below the name of each test set.
The best result for each test set is highlighted in \textbf{bold} and the second-best is \underline{underlined}.}\vspace{-15pt}
\label{tab:car_r50_retina_ccc}
\end{table*}

\begin{table*}[ht]
\centering
\renewcommand{\arraystretch}{1.1}
\resizebox{\textwidth}{!}{%
\begin{tabular}{l|cccccccccc|c}
\toprule
\multirow{2}{*}{Method} & COCO & BDD & Cityscapes & DETRAC & ExDark & KITTI & Self-driving & Roboflow & Udacity & Traffic & \multirow{2}{*}{\makecell{Avg. \\ RMSE}} \\
 & 35.8 & 33.5 & 40.2 & 39.9 & 25.8 & 42.1 & 30.6 & 27.3 & 28.7 & 25.5 & \\
\midrule
PS~\cite{hendrycks2018ps} & 12.79 & 7.87 & \textbf{4.23} & 38.03 & \textbf{1.11} & 21.21 & 8.71 & \textbf{4.32} & 5.96 & \underline{0.12} & 10.44 \\
ES~\cite{saito2019semi} & 13.94 & \underline{2.91} & 16.53 & \textbf{1.85} & 3.54 & 19.70 & 3.52 & \underline{4.51} & 4.57 & 0.93 & 7.20 \\
AC~\cite{guillory2021predicting} & 13.68 & 6.21 & 13.87 & 34.55 & 2.92 & 20.12 & 5.49 & 4.92 & 4.68 & \textbf{0.08} & 10.65 \\
ATC~\cite{arg2022leveragingunlabeleddatapredict} & 12.76 & 7.92 & 13.00 & 53.19 & \underline{2.42} & 19.06 & 6.70 & 4.58 & 4.97 & 0.55 & 12.52 \\
BoS~\cite{yang2024bos} & \textbf{6.14} & 4.02 & 9.78 & 8.04 & 3.17 & \underline{13.51} & \textbf{0.12} & 4.92 & \textbf{1.11} & 0.97 & \underline{5.18} \\
\midrule
PCR & \underline{9.66} & \textbf{2.72} & \underline{4.63} & \underline{3.98} & 3.35 & \textbf{6.96} & \underline{0.93} & 5.29 & \underline{1.62} & 3.41 & \textbf{4.26} \\

\bottomrule
\end{tabular}%
}
\caption{Comparison of AutoEval methods for \textbf{vehicle detection} on the \textbf{corruption-based meta-dataset}.
\textbf{RetinaNet with Swin Transformer} trained on the ``car''-class subset of COCO is used as the detector, and its true mAP (\%) is reported below the name of each test set.
The best result for each test set is highlighted in \textbf{bold} and the second-best is \underline{underlined}.}\vspace{-15pt}
\label{tab:car_swin_retina_ccc}
\end{table*}

\begin{table*}[ht]
\centering
\renewcommand{\arraystretch}{1.1}
\resizebox{\textwidth}{!}{%
\begin{tabular}{l|cccccccccc|c}
\toprule
\multirow{2}{*}{Method} & COCO & BDD & Cityscapes & DETRAC & ExDark & KITTI & Self-driving & Roboflow & Udacity & Traffic & \multirow{2}{*}{\makecell{Avg. \\ RMSE}} \\
 & 36.0 & 32.2 & 40.5 & 36.2 & 25.9 & 38.9 & 29.4 & 29.4 & 27.8 & 27.7 & \\
\midrule
PS~\cite{hendrycks2018ps} & 16.36 & 13.26 & \underline{16.39} & \textbf{9.57} & 4.89 & \underline{13.86} & \underline{5.17} & 13.81 & \textbf{2.62} & 8.15 & 10.41 \\
ES~\cite{saito2019semi} & 19.41 & \underline{12.04} & 20.07 & 12.84 & 7.02 & 17.44 & 7.49 & 15.59 & 4.79 & 8.64 & 12.53 \\
AC~\cite{guillory2021predicting} & 18.37 & 13.16 & 19.15 & \underline{9.61} & 6.86 & 17.37 & 6.57 & 12.44 & 5.64 & 8.49 & 11.77 \\
ATC~\cite{arg2022leveragingunlabeleddatapredict} & 17.14 & 15.01 & 20.02 & 12.20 & 6.01 & 17.21 & 7.23 & 11.05 & 5.55 & 8.34 & 11.98 \\
BoS~\cite{yang2024bos} & \underline{14.64} & 13.04 & 16.96 & 13.47 & \underline{3.93} & 14.51 & 7.63 & \underline{6.59} & 4.61 & \underline{7.81} & \underline{10.32} \\
\midrule
PCR & \textbf{10.62} & \textbf{6.53} & \textbf{14.89} & 13.97 & \textbf{1.89} & \textbf{14.23} & \textbf{4.52} & \textbf{0.03} & \underline{3.02} & \textbf{2.58} & \textbf{7.23} \\
\bottomrule
\end{tabular}%
}
\caption{Comparison of AutoEval methods for \textbf{vehicle detection} on the \textbf{corruption-based meta-dataset}.
\textbf{Faster R-CNN with ResNet-50} trained on the ``car''-class subset of COCO is used as the detector, and its true mAP (\%) is reported below the name of each test set.
The best result for each test set is highlighted in \textbf{bold} and the second-best is \underline{underlined}.}\vspace{-15pt}
\label{tab:car_r50_faster_ccc}
\end{table*}

\begin{table*}[ht]
\centering
\renewcommand{\arraystretch}{1.1}
\resizebox{\textwidth}{!}{%
\begin{tabular}{l|cccccccccc|c}
\toprule
\multirow{2}{*}{Method} & COCO & BDD & Cityscapes & DETRAC & ExDark & KITTI & Self-driving & Roboflow & Udacity & Traffic & \multirow{2}{*}{\makecell{Avg. \\ RMSE}} \\
 & 38.7 & 34.4 & 42.6 & 36.4 & 31.6 & 44.4 & 31.4 & 31.9 & 29.2 & 27.3 & \\
\midrule
PS~\cite{hendrycks2018ps} & \textbf{9.18} & 6.82 & \textbf{4.37} & 15.67 & \underline{5.04} & \textbf{6.76} & 1.24 & 7.60 & 6.00 & 0.33 & 6.30 \\
ES~\cite{saito2019semi} & 16.25 & 11.24 & 17.05 & 8.66 & 6.83 & 18.81 & 5.92 & 9.95 & \underline{0.42} & 5.07 & 10.02 \\
AC~\cite{guillory2021predicting} & 11.05 & \underline{6.60} & \underline{7.00} & 15.33 & 6.82 & 12.27 & 0.68 & 8.44 & 2.69 & 1.43 & 7.23 \\
ATC~\cite{arg2022leveragingunlabeleddatapredict} & 10.08 & 6.89 & 8.50 & 10.29 & 5.08 & 11.86 & \underline{0.50} & 6.65 & 2.28 & \textbf{0.72} & \underline{6.29} \\
BoS~\cite{yang2024bos} & 10.63 & 7.58 & 15.20 & \underline{7.97} & 4.54 & 17.46 & 3.12 & \textbf{0.54} & \textbf{0.36} & 1.08 & 6.85 \\
\midrule
PCR  & \underline{9.49} & \textbf{5.93} & 8.11 & \textbf{2.47} & \textbf{2.16} & \underline{11.51} & \textbf{0.10} & \underline{4.52} & 1.52 & \underline{1.22} & \textbf{4.70} \\
\bottomrule
\end{tabular}%
}
\caption{Comparison of AutoEval methods for \textbf{vehicle detection} on the \textbf{corruption-based meta-dataset}.
\textbf{Faster R-CNN with Swin Transformer} trained on the ``car''-class subset of COCO is used as the detector, and its true mAP (\%) is reported below the name of each test set.
The best result for each test set is highlighted in \textbf{bold} and the second-best is \underline{underlined}.}\vspace{-15pt}
\label{tab:car_swin_faster_ccc}
\end{table*}

\begin{table*}[ht]
\centering
\renewcommand{\arraystretch}{1.1}
\resizebox{\textwidth}{!}{%
\begin{tabular}{l|ccccccccc|c}
\toprule
\multirow{2}{*}{\text{Method}} 
  & \text{COCO} & \text{Caltech} & \text{CityPersons} & \text{Cityscapes} & \text{CrowdHuman} 
  & \text{ECP} & \text{ExDark} & \text{KITTI} & \text{Self-driving} 
  & \multirow{2}{*}{\makecell{Avg. \\ RMSE}} \\
  & 28.3 & 12.0 & 17.0 & 21.5 & 34.2  & 24.1 & 27.0  & 15.8  & 17.4  & \\
\midrule
PS~\cite{hendrycks2018ps} & 6.61 & 4.46 & \textbf{0.13} & 4.21 & 35.67 & 7.62 & 5.72 & \underline{0.60} & 2.24 & 7.47 \\
ES~\cite{saito2019semi} & 10.97 & \underline{2.53} & 2.58 & \textbf{0.34} & 11.29 & 6.04 & 9.12 & 0.97 & 2.36 & 5.13 \\
AC~\cite{guillory2021predicting} & 7.68 & 3.34 & 2.35 & \textbf{0.34} & 17.50 & 6.59 & 5.91 & 0.61 & 0.99 & 5.03 \\
ATC~\cite{arg2022leveragingunlabeleddatapredict} & 8.68 & 3.49 & \underline{0.81} & 1.84 & 28.37 & 7.36 & 4.76 & \textbf{0.12} & \textbf{0.46} & 6.21 \\
BoS~\cite{yang2024bos} & \textbf{4.29} & 4.59 & 1.42 & 1.14 & \underline{9.49} & \underline{4.17} & \textbf{0.39} & 4.44 & 3.99 & \underline{3.77} \\
\midrule
PCR  & \underline{6.23} & \textbf{1.49} & 0.93 & 0.46 & \textbf{7.24} & \textbf{4.09} & \underline{3.72} & 1.09 & \underline{0.63} & \textbf{2.88} \\

\bottomrule
\end{tabular}%
}
\caption{Comparison of AutoEval methods for \textbf{pedestrian detection} on the \textbf{augmentation-based meta-dataset}~\cite{yang2024bos}.
\textbf{RetinaNet with ResNet-50} trained on CrowdHuman is used as the detector, and its true mAP (\%) is reported below the name of each test set.
The best result for each test set is highlighted in \textbf{bold} and the second-best is \underline{underlined}.}\vspace{-15pt}
\label{tab:person_r50_retina_bos}
\end{table*}

\begin{table*}[ht]
\centering
\renewcommand{\arraystretch}{1.1}
\resizebox{\textwidth}{!}{%
\begin{tabular}{l|ccccccccc|c}
\toprule
\multirow{2}{*}{\text{Method}} & \text{COCO} & \text{Caltech} & \text{CityPersons} & \text{Cityscapes} & \text{CrowdHuman} & \text{ECP} & \text{ExDark} & \text{KITTI} & \text{Self-driving} & \multirow{2}{*}{\makecell{Avg. \\ RMSE}} \\
 & 24.6 & 13.2 & 17.9 & 21.6 & 31.0 & 24.8 & 22.0 & 16.2 & 17.1 & \\
\midrule
PS~\cite{hendrycks2018ps} & 6.15 & 3.52 & 0.90 & 2.63 & 39.55 & 5.97 & 4.80 & \textbf{0.07} & 2.54 & 7.35 \\
ES~\cite{saito2019semi} & 6.40 & \underline{0.77} & 0.80 & \underline{0.97} & \textbf{5.03} & 8.45 & 4.26 & 1.92 & 3.39 & 3.55 \\
AC~\cite{guillory2021predicting} & 5.42 & 1.76 & \textbf{0.34} & 1.97 & 17.46 & 8.22 & 3.20 & \underline{0.08} & 1.03 & 4.39 \\
ATC~\cite{arg2022leveragingunlabeleddatapredict} & 6.16 & 3.10 & \underline{0.46} & 3.32 & 41.66 & 8.49 & 2.81 & 0.82 & \textbf{0.30} & 7.46 \\
BoS~\cite{yang2024bos} & \textbf{1.49} & 4.03 & 5.29 & 1.98 & \underline{5.78} & \textbf{2.08} & \underline{2.65} & 3.34 & 1.85 & \underline{3.17} \\
\midrule
PCR  & \underline{4.12} & \textbf{0.18} & \underline{0.46} & \textbf{0.44} & 7.78 & \underline{6.37} & \textbf{2.13} & 2.26 & \underline{0.72} & \textbf{2.72} \\
\bottomrule
\end{tabular}%
}
\caption{Comparison of AutoEval methods for \textbf{pedestrian detection} on the \textbf{augmentation-based meta-dataset}~\cite{yang2024bos}.
\textbf{RetinaNet with Swin Transformer} trained on CrowdHuman is used as the detector, and its true mAP (\%) is reported below the name of each test set.
The best result for each test set is highlighted in \textbf{bold} and the second-best is \underline{underlined}.}\vspace{-15pt}
\label{tab:person_swin_retina_bos}
\end{table*}

\begin{table*}[ht]
\centering
\renewcommand{\arraystretch}{1.1}
\resizebox{\textwidth}{!}{%
\begin{tabular}{l|ccccccccc|c}
\toprule
\multirow{2}{*}{\text{Method}} & \text{COCO} & \text{Caltech} & \text{CityPersons} & \text{Cityscapes} & \text{CrowdHuman} & \text{ECP} & \text{ExDark} & \text{KITTI} & \text{Self-driving} & \multirow{2}{*}{\makecell{Avg. \\ RMSE}} \\
 & 28.3 & 11.7 & 17.7 & 22.6 & 36.3 & 26.1 & 26.0 & 18.1 & 15.9 & \\
\midrule
PS~\cite{hendrycks2018ps} & \textbf{4.44} & 6.43 & \underline{1.16} & \underline{2.08} & \underline{11.06} & 8.29 & 1.24 & 10.56 & 10.56 & 6.20 \\
ES~\cite{saito2019semi} & 9.82 & 8.88 & 1.45 & 2.92 & 20.53 & 7.93 & 6.28 & \underline{2.20} & 13.42 & 8.16 \\
AC~\cite{guillory2021predicting} & 7.57 & 8.79 & 1.43 & 2.36 & 17.45 & 8.91 & 2.91 & \textbf{1.22} & 10.55 & 6.80 \\
ATC~\cite{arg2022leveragingunlabeleddatapredict} & 8.44 & 6.97 & \textbf{0.91} & 4.21 & 18.23 & 9.76 & \underline{0.75} & 6.02 & 7.51 & 6.98 \\
BoS~\cite{yang2024bos} & 6.80 & \underline{6.38} & 1.47 & 2.59 & 14.54 & \underline{7.85} & \textbf{0.62} & 7.14 & \underline{5.50} & \underline{5.88} \\
\midrule
PCR  & \underline{5.43} & \textbf{1.71} & 3.48 & \textbf{1.52} & \textbf{5.38} & \textbf{5.35} & 2.44 & 2.32 & \textbf{2.84} & \textbf{3.39} \\

\bottomrule
\end{tabular}%
}
\caption{Comparison of AutoEval methods for \textbf{pedestrian detection} on the \textbf{augmentation-based meta-dataset}~\cite{yang2024bos}.
\textbf{Faster R-CNN with ResNet-50} trained on CrowdHuman is used as the detector, and its true mAP (\%) is reported below the name of each test set.
The best result for each test set is highlighted in \textbf{bold} and the second-best is \underline{underlined}.}\vspace{-15pt}
\label{tab:person_r50_faster_bos}
\end{table*}

\begin{table*}[ht]
\centering
\renewcommand{\arraystretch}{1.1}
\resizebox{\textwidth}{!}{%
\begin{tabular}{l|ccccccccc|c}
\toprule
\multirow{2}{*}{\text{Method}} 
  & \text{COCO} & \text{Caltech} & \text{CityPersons} & \text{Cityscapes} 
  & \text{CrowdHuman} & \text{ECP} & \text{ExDark} & \text{KITTI} 
  & \text{Self-driving} 
  & \multirow{2}{*}{\makecell{Avg. \\ RMSE}} \\
  & 27.6 & 12.9 & 19.7 & 24.0 & 35.2 & 27.3 & 27.7 & 20.2 & 17.4 & \\
\midrule
PS~\cite{hendrycks2018ps} & 3.75 & 4.97 & \textbf{1.26} & 3.63 & 3.73 & 7.03 & \underline{1.28} & 13.60 & 2.52 & 4.64 \\
ES~\cite{saito2019semi} & 9.76 & 9.11 & 1.45 & 5.04 & 16.96 & 8.43 & 7.53 & 1.91 & \underline{1.81} & 6.89 \\
AC~\cite{guillory2021predicting} & \underline{2.82} & 5.60 & \underline{1.34} & 1.54 & 9.08 & 6.98 & 1.89 & \underline{1.60} & 5.80 & 4.07 \\
ATC~\cite{arg2022leveragingunlabeleddatapredict} & 4.08 & 5.70 & 1.59 & 3.99 & \underline{0.82} & 7.88 & \textbf{0.90} & 7.01 & 2.99 & 3.88 \\
BoS~\cite{yang2024bos} & \textbf{2.47} & \underline{4.49} & 2.08 & \underline{1.51} & \textbf{0.75} & \textbf{5.57} & 1.64 & 8.71 & 4.88 & \underline{3.57} \\
\midrule
PCR  & 4.02 & \textbf{1.84} & 1.39 & \textbf{1.13} & 4.11 & \underline{6.49} & 2.46 & \textbf{0.07} & \textbf{1.60} & \textbf{2.57} \\

\bottomrule
\end{tabular}%
}
\caption{Comparison of AutoEval methods for \textbf{pedestrian detection} on the \textbf{augmentation-based meta-dataset}~\cite{yang2024bos}.
\textbf{Faster R-CNN with Swin Transformer} trained on CrowdHuman is used as the detector, and its true mAP (\%) is reported below the name of each test set.
The best result for each test set is highlighted in \textbf{bold} and the second-best is \underline{underlined}.}\vspace{-15pt}
\label{tab:person_swin_faster_bos}
\end{table*}

\begin{table*}[ht]
\centering
\renewcommand{\arraystretch}{1.1}
\resizebox{\textwidth}{!}{%
\begin{tabular}{l|ccccccccc|c}
\toprule
\multirow{2}{*}{\text{Method}} 
  & \text{COCO} & \text{Caltech} & \text{CityPersons} & \text{Cityscapes} & \text{CrowdHuman} 
  & \text{ECP} & \text{ExDark} & \text{KITTI} & \text{Self-driving} 
  & \multirow{2}{*}{\makecell{Avg. \\ RMSE}} \\
  & 28.3 & 12.0 & 17.0 & 21.5 & 34.2  & 24.1 & 27.0  & 15.8  & 17.4  & \\
\midrule
PS~\cite{hendrycks2018ps} & \underline{9.80} & 1.63 & 5.08 & 8.69 & 37.78 & 11.71 & 9.11 & 6.81 & \textbf{1.99} & 10.29 \\
ES~\cite{saito2019semi} & 14.78 & 5.76 & \underline{1.77} & \underline{3.96} & 14.16 & \underline{9.37} & 12.87 & 7.45 & 7.94 & 8.67 \\
AC~\cite{guillory2021predicting} & 12.72 & 3.91 & 3.37 & 5.57 & \underline{11.74} & 10.97 & 10.20 & 7.56 & 4.70 & \underline{7.54} \\
ATC~\cite{arg2022leveragingunlabeleddatapredict} & 13.09 & 3.15 & 4.28 & 6.33 & 10.56 & 11.77 & \underline{8.76} & 6.34 & 4.63 & 7.66 \\
BoS~\cite{yang2024bos} & 16.72 & \textbf{0.10} & 5.10 & 9.72 & 25.11 & 12.32 & 14.50 & \underline{3.60} & 4.94 & 10.23 \\
\midrule
PCR  & \textbf{7.52} & \underline{0.70} & \textbf{0.58} & \textbf{0.36} & \textbf{0.44} & \textbf{6.75} & \textbf{6.18} & \textbf{2.00} & \underline{2.13} & \textbf{2.96} \\

\bottomrule
\end{tabular}%
}
\caption{Comparison of AutoEval methods for \textbf{pedestrian detection} on the \textbf{corruption-based meta-dataset}.
\textbf{RetinaNet with ResNet-50} trained on CrowdHuman is used as the detector, and its true mAP (\%) is reported below the name of each test set.
The best result for each test set is highlighted in \textbf{bold} and the second-best is \underline{underlined}.}\vspace{-15pt}
\label{tab:person_r50_retina_ccc}
\end{table*}

\begin{table*}[ht]
\centering
\renewcommand{\arraystretch}{1.1}
\resizebox{\textwidth}{!}{%
\begin{tabular}{l|ccccccccc|c}
\toprule
\multirow{2}{*}{\text{Method}} & \text{COCO} & \text{Caltech} & \text{CityPersons} & \text{Cityscapes} & \text{CrowdHuman} & \text{ECP} & \text{ExDark} & \text{KITTI} & \text{Self-driving} & \multirow{2}{*}{\makecell{Avg. \\ RMSE}} \\
 & 24.6 & 13.2 & 17.9 & 21.6 & 31.0 & 24.8 & 22.0 & 16.2 & 17.1 & \\
\midrule
PS~\cite{hendrycks2018ps} & \underline{8.23} & \underline{0.57} & \textbf{1.35} & \underline{4.52} & 62.54 & \textbf{7.51} & 7.44 & 3.60 & \textbf{0.65} & 10.71 \\
ES~\cite{saito2019semi} & 9.94 & 4.78 & 3.15 & 5.34 & 12.68 & 11.52 & 7.52 & 5.03 & 6.65 & 7.40 \\
AC~\cite{guillory2021predicting} & 9.35 & 3.62 & 3.88 & 5.97 & \textbf{0.97} & 12.00 & 7.12 & 4.74 & 5.48 & \underline{5.90} \\
ATC~\cite{arg2022leveragingunlabeleddatapredict} & 10.37 & 2.28 & 4.96 & 7.55 & 4.86 & 12.55 & 6.95 & 4.09 & 4.96 & 6.51 \\
BoS~\cite{yang2024bos} & 9.25 & \textbf{0.13} & 2.59 & 6.27 & 16.62 & 9.72 & \textbf{5.51} & \underline{2.00} & \underline{2.68} & 6.08 \\
\midrule
PCR  & \textbf{5.72} & 3.85 & \underline{1.45} & \textbf{1.60} & \underline{4.11} & \underline{8.60} & \underline{3.79} & \textbf{0.31} & 3.45 & \textbf{3.65} \\

\bottomrule
\end{tabular}%
}
\caption{Comparison of AutoEval methods for \textbf{pedestrian detection} on the \textbf{corruption-based meta-dataset}.
\textbf{RetinaNet with Swin Transformer} trained on CrowdHuman is used as the detector, and its true mAP (\%) is reported below the name of each test set.
The best result for each test set is highlighted in \textbf{bold} and the second-best is \underline{underlined}.}\vspace{-15pt}
\label{tab:person_swin_retina_ccc}
\end{table*}

\begin{table*}[ht]
\centering
\renewcommand{\arraystretch}{1.1}
\resizebox{\textwidth}{!}{%
\begin{tabular}{l|ccccccccc|c}
\toprule
\multirow{2}{*}{\text{Method}} & \text{COCO} & \text{Caltech} & \text{CityPersons} & \text{Cityscapes} & \text{CrowdHuman} & \text{ECP} & \text{ExDark} & \text{KITTI} & \text{Self-driving} & \multirow{2}{*}{\makecell{Avg. \\ RMSE}} \\
 & 28.3 & 11.7 & 17.7 & 22.6 & 36.3 & 26.1 & 26.0 & 18.1 & 15.9 & \\
\midrule
PS~\cite{hendrycks2018ps} & \textbf{9.66} & 1.46 & \underline{2.74} & \underline{6.24} & \textbf{15.07} & \underline{11.96} & \textbf{4.82} & 4.62 & 5.00 & \underline{6.84} \\
ES~\cite{saito2019semi} & 16.97 & 1.63 & 5.64 & 10.73 & 26.25 & 14.59 & 13.95 & 5.82 & \underline{0.37} & 10.66 \\
AC~\cite{guillory2021predicting} & 14.47 & 2.05 & 4.48 & 8.60 & 20.47 & 14.24 & 10.65 & 5.70 & 1.71 & 9.15 \\
ATC~\cite{arg2022leveragingunlabeleddatapredict} & 14.30 & \textbf{0.14} & 6.19 & 9.57 & 20.29 & 14.64 & 7.84 & \underline{0.76} & 1.24 & 8.33 \\
BoS~\cite{yang2024bos} & 15.88 & \textbf{0.14} & 6.86 & 10.30 & 24.67 & 14.46 & 11.39 & 4.15 & 2.94 & 10.09 \\
\midrule
PCR  & \underline{10.56} & 1.66 & \textbf{1.70} & \textbf{4.51} & \underline{15.29} & \textbf{9.93} & \underline{6.17} & \textbf{0.02} & \textbf{0.05} & \textbf{5.54} \\

\bottomrule
\end{tabular}%
}
\caption{Comparison of AutoEval methods for \textbf{pedestrian detection} on the \textbf{corruption-based meta-dataset}.
\textbf{Faster R-CNN with ResNet-50} trained on CrowdHuman is used as the detector, and its true mAP (\%) is reported below the name of each test set.
The best result for each test set is highlighted in \textbf{bold} and the second-best is \underline{underlined}.}\vspace{-15pt}
\label{tab:person_r50_faster_ccc}
\end{table*}

\begin{table*}[ht]
\centering
\renewcommand{\arraystretch}{1.1}
\resizebox{\textwidth}{!}{%
\begin{tabular}{l|ccccccccc|c}
\toprule
\multirow{2}{*}{\text{Method}} 
  & \text{COCO} & \text{Caltech} & \text{CityPersons} & \text{Cityscapes} 
  & \text{CrowdHuman} & \text{ECP} & \text{ExDark} & \text{KITTI} 
  & \text{Self-driving} 
  & \multirow{2}{*}{\makecell{Avg. \\ RMSE}} \\
  & 27.6 & 12.9 & 19.7 & 24.0 & 35.2 & 27.3 & 27.7 & 20.2 & 17.4 & \\
\midrule
PS~\cite{hendrycks2018ps} & \underline{6.80} & 0.91 & 5.08 & 6.93 & \textbf{6.13} & \underline{10.11} & \textbf{4.00} & 11.81 & 1.56 & 5.93 \\
ES~\cite{saito2019semi} & 14.01 & 2.05 & 5.24 & 10.05 & 22.81 & 13.31 & 13.27 & 6.33 & 2.92 & 10.00 \\
AC~\cite{guillory2021predicting} & 7.92 & \underline{0.26} & \underline{3.23} & \underline{6.08} & \underline{8.36} & 11.30 & 7.01 & 3.29 & \textbf{0.85} & \underline{5.37} \\
ATC~\cite{arg2022leveragingunlabeleddatapredict} & 8.49 & \textbf{0.02} & 6.04 & 8.23 & 9.24 & 11.95 & 5.24 & 2.63 & 1.71 & 5.95 \\
BoS~\cite{yang2024bos} & 10.49 & 0.58 & 4.25 & 8.24 & 17.48 & 11.86 & 7.91 & \underline{1.14} & \underline{0.91} & 6.98 \\
\midrule
PCR  & \textbf{6.66} & 3.17 & \textbf{3.17} & \textbf{4.45} & 8.46 & \textbf{9.51} & \underline{4.69} & \textbf{0.83} & 2.94 & \textbf{4.88} \\
\bottomrule
\end{tabular}%
}
\caption{Comparison of AutoEval methods for \textbf{pedestrian detection} on the \textbf{corruption-based meta-dataset}.
\textbf{Faster R-CNN with Swin Transformer} trained on CrowdHuman is used as the detector, and its true mAP (\%) is reported below the name of each test set.
The best result for each test set is highlighted in \textbf{bold} and the second-best is \underline{underlined}.}\vspace{-15pt}
\label{tab:person_swin_faster_ccc}
\end{table*}

\begin{figure*}[t!]
    \centering
    \begin{minipage}{\textwidth}
        \makebox[\textwidth]{
            \includegraphics[width=0.12\textwidth]{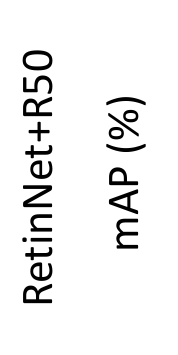}
            \includegraphics[width=0.24\textwidth]{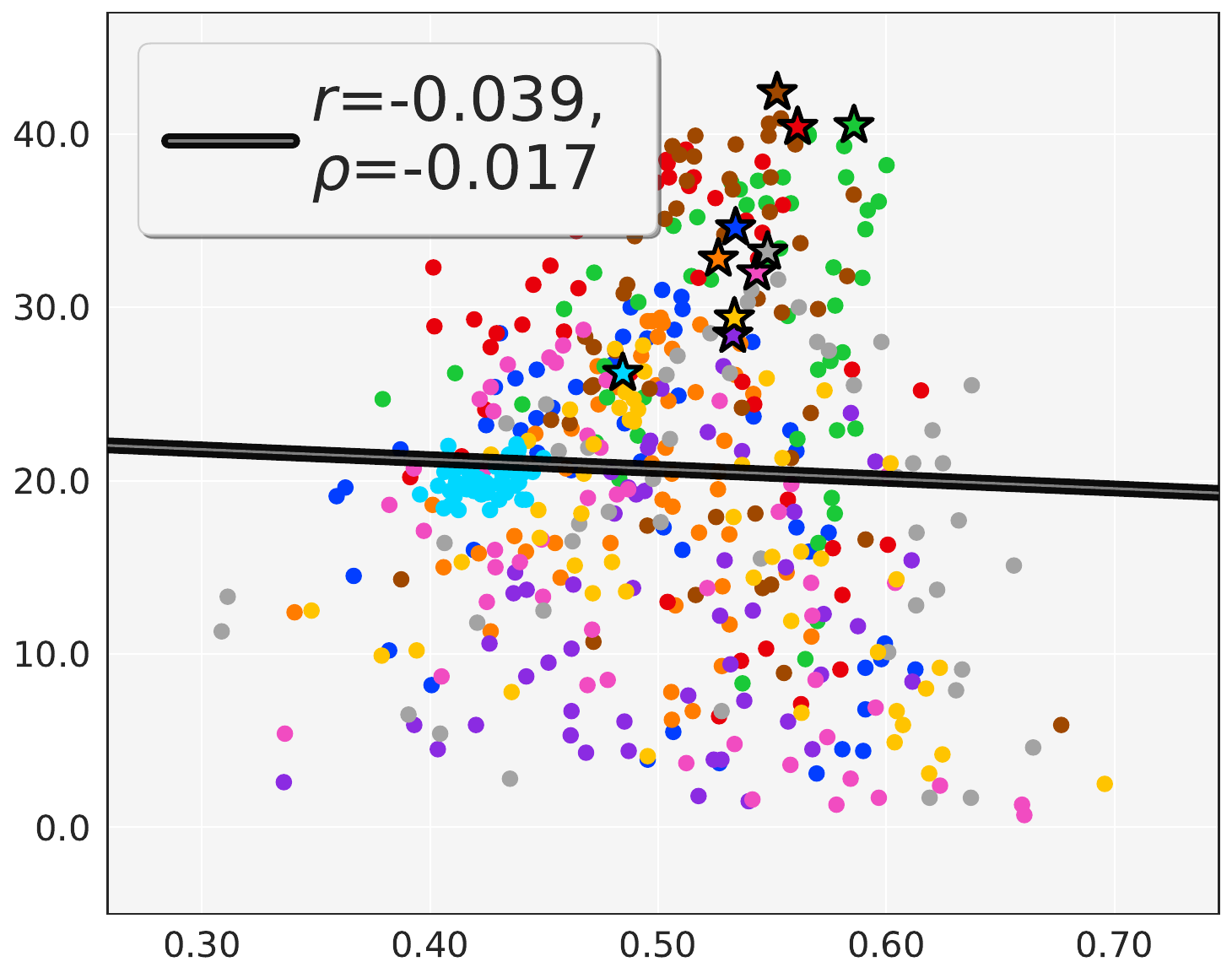}
            \includegraphics[width=0.24\textwidth]{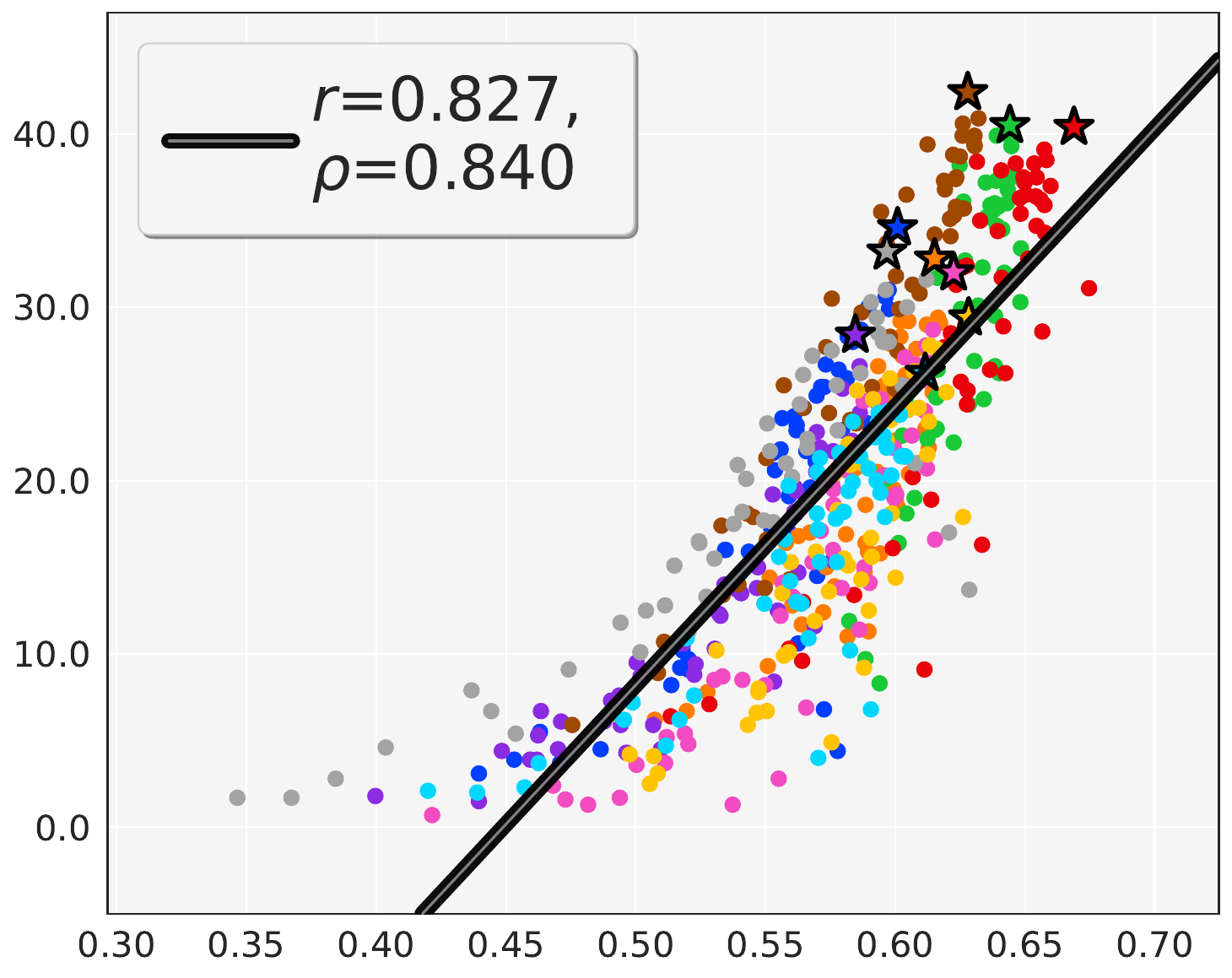}
            \includegraphics[width=0.24\textwidth]{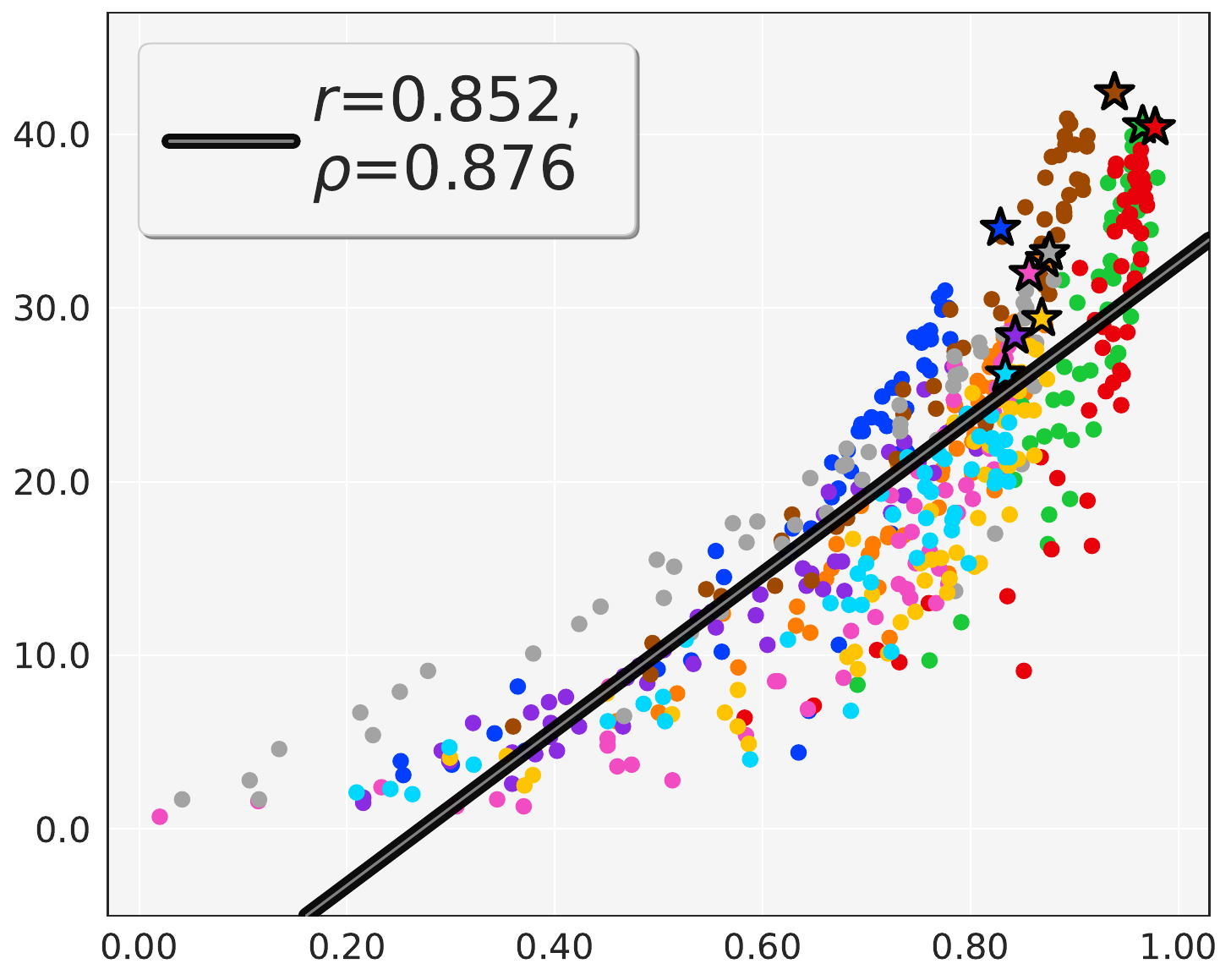}
            \includegraphics[width=0.12\textwidth]{figs/car_legend2.pdf}
            \\
        }
        \makebox[\textwidth]{
            \includegraphics[width=0.12\textwidth]{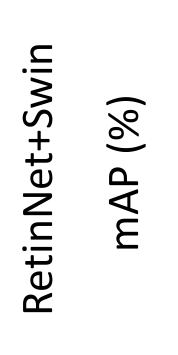}
            \includegraphics[width=0.24\textwidth]{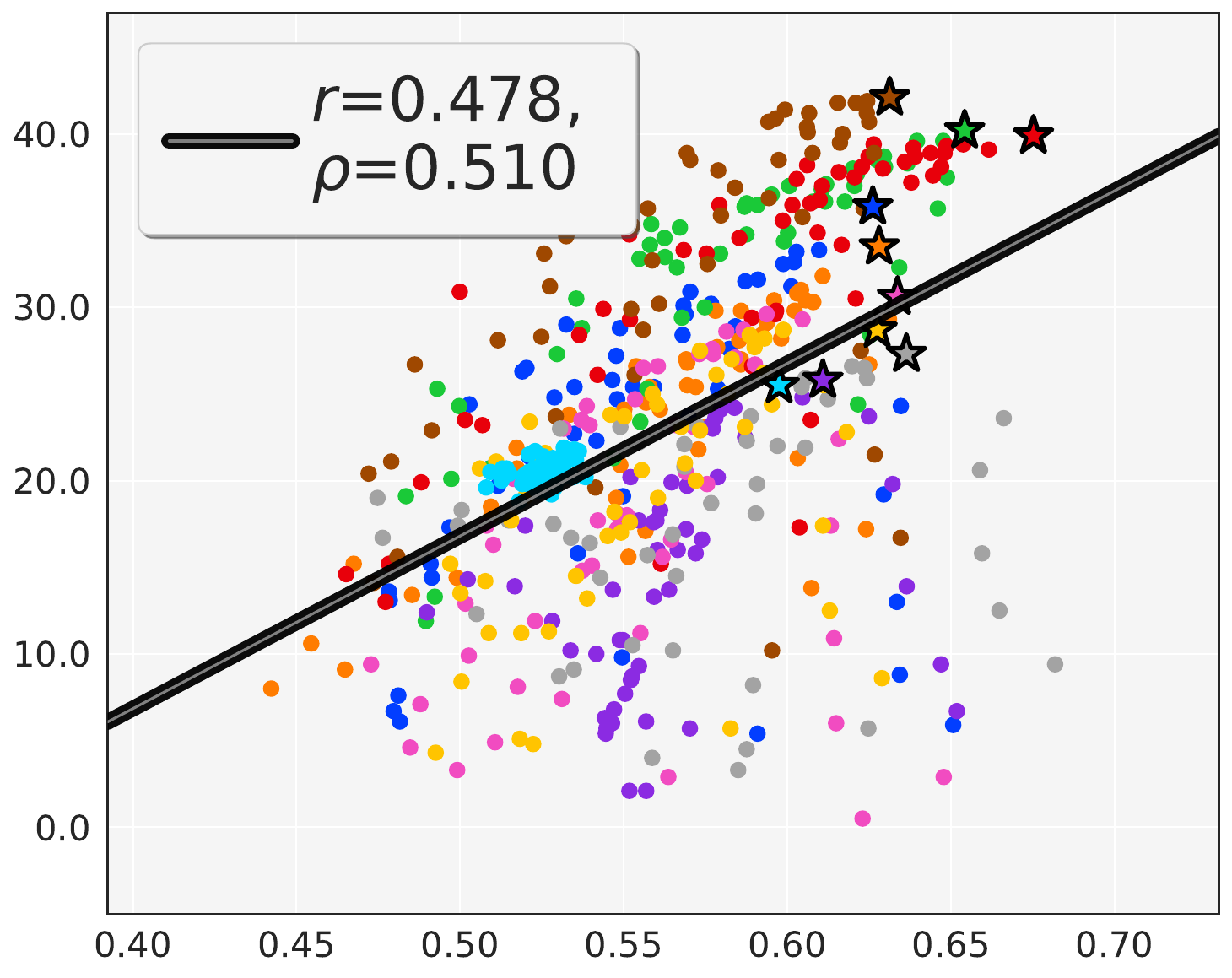}
            \includegraphics[width=0.24\textwidth]{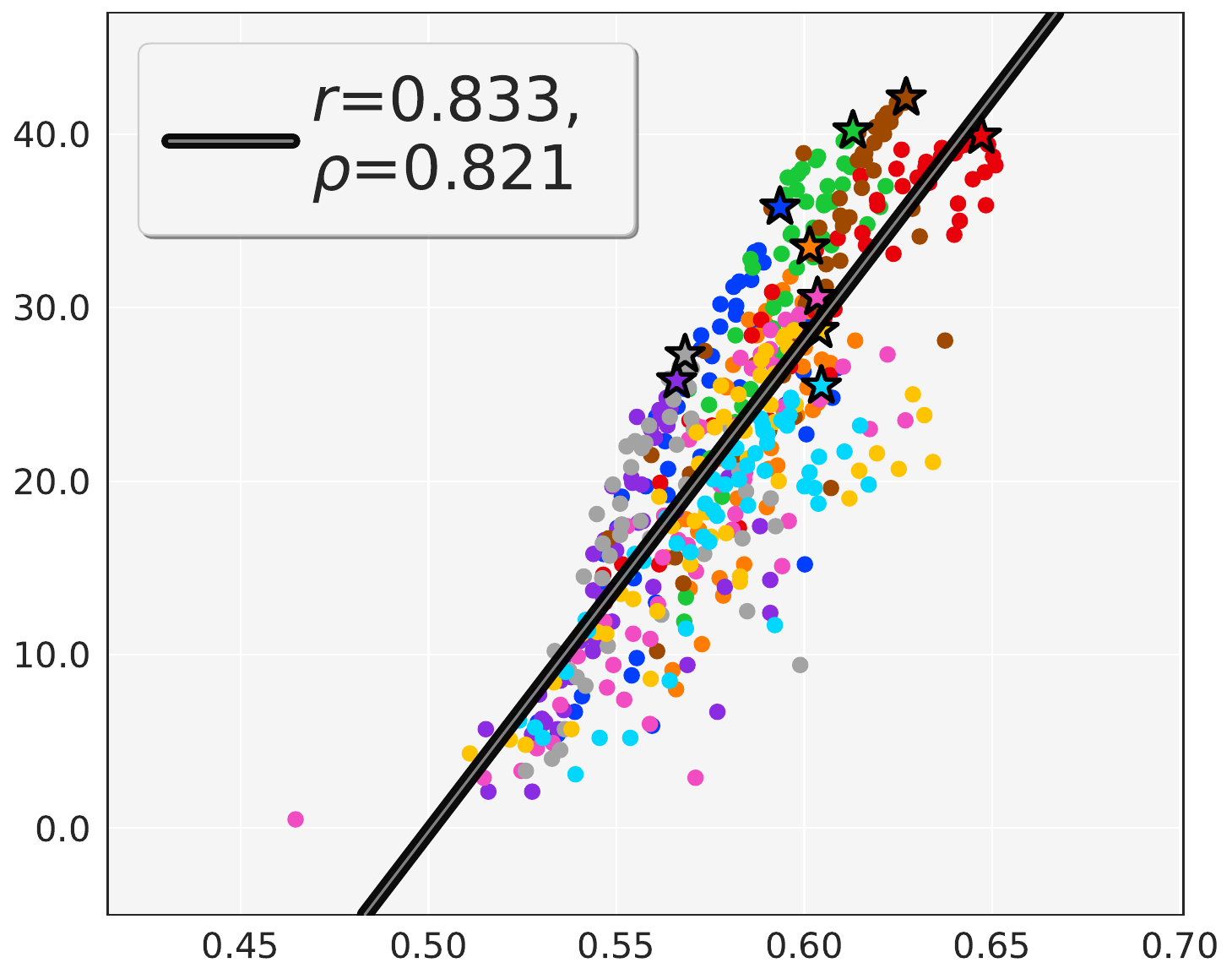}
            \includegraphics[width=0.24\textwidth]{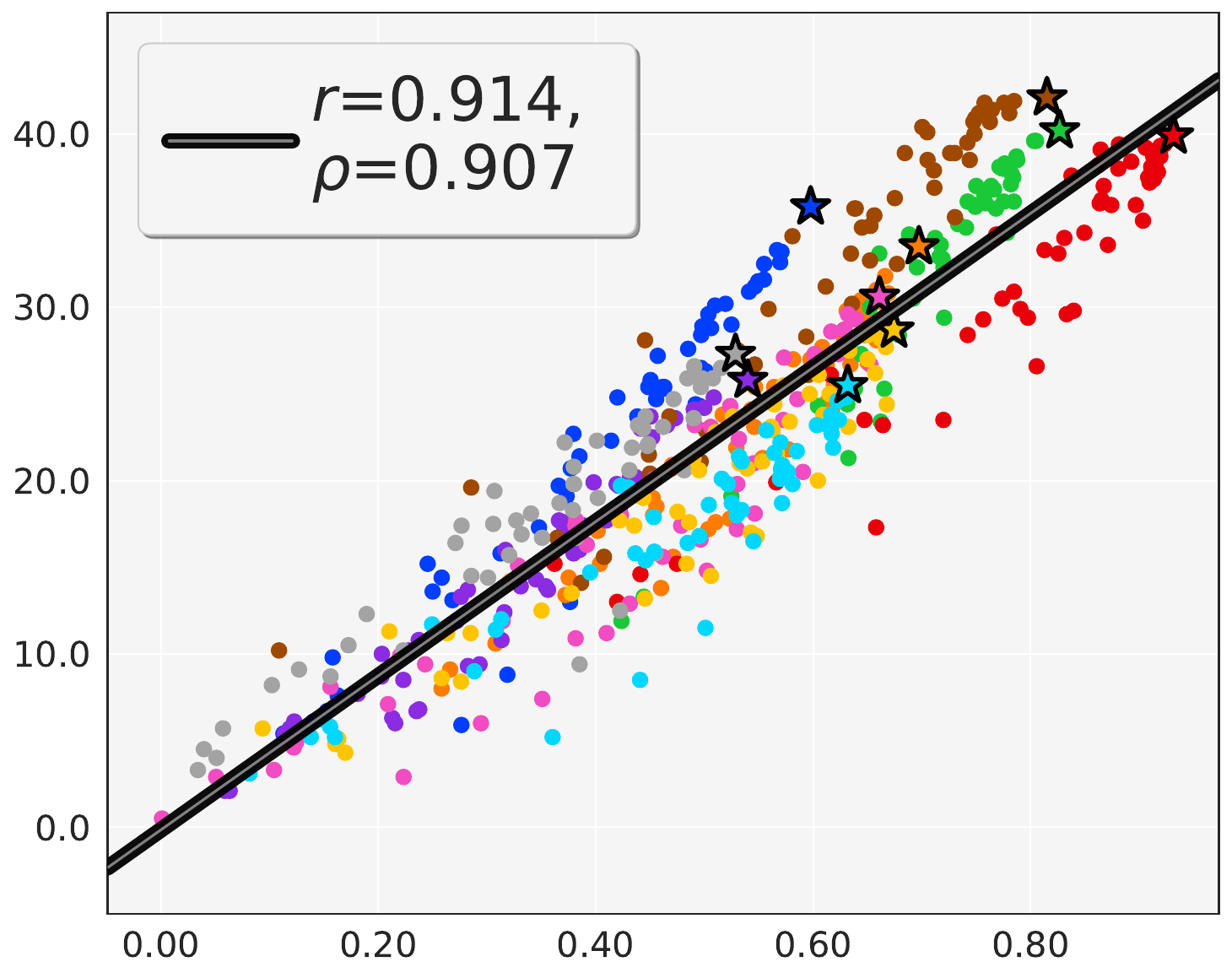}
            \includegraphics[width=0.12\textwidth]{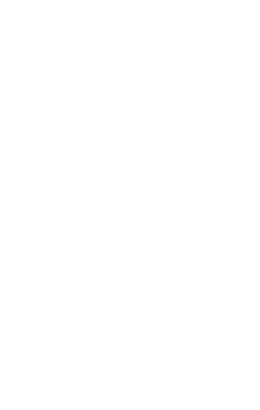}
            \\
        }
        \makebox[\textwidth]{
            \includegraphics[width=0.12\textwidth]{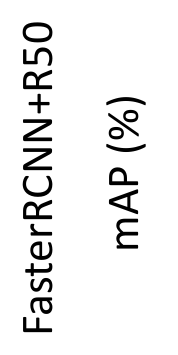}
            \includegraphics[width=0.24\textwidth]{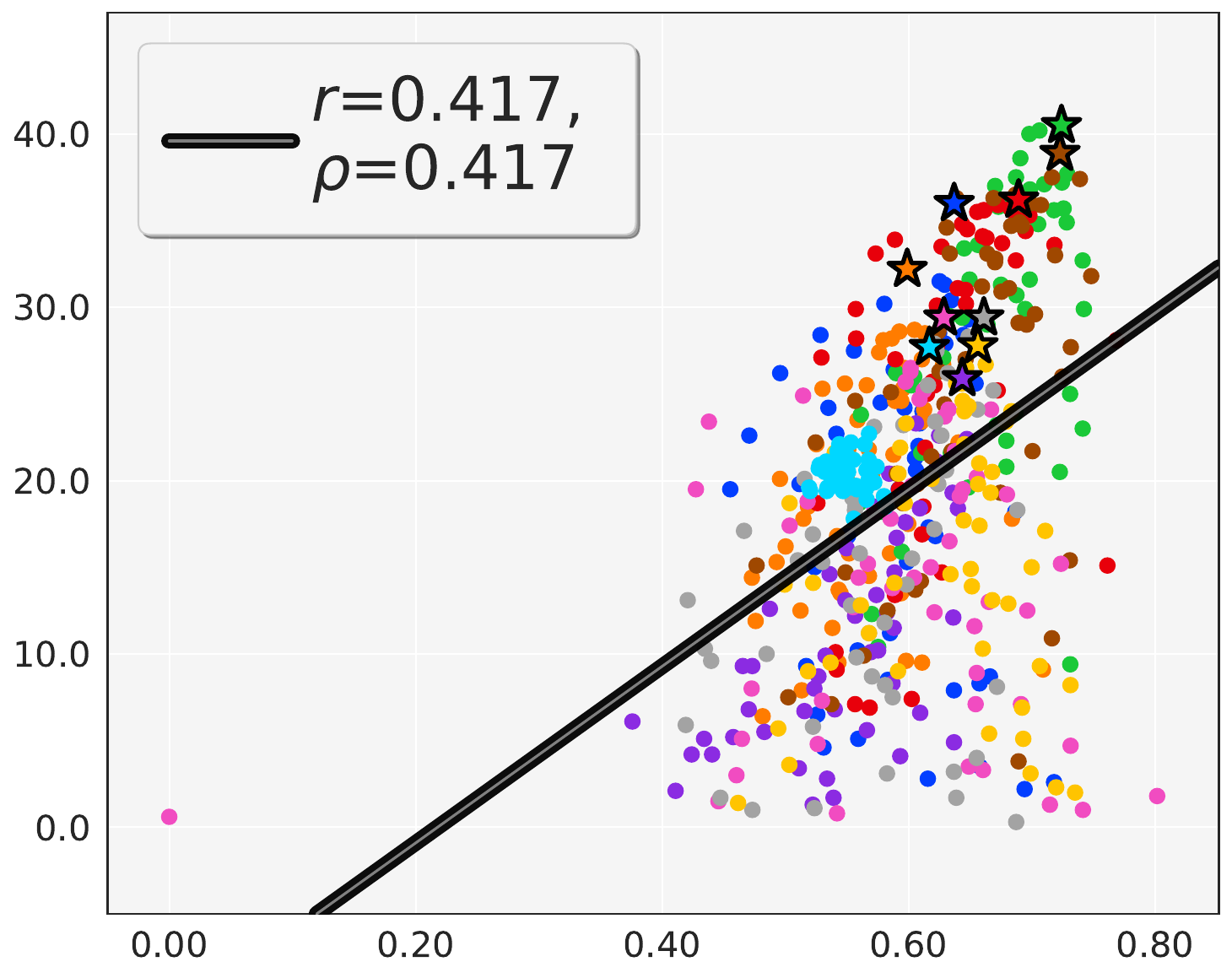}
            \includegraphics[width=0.24\textwidth]{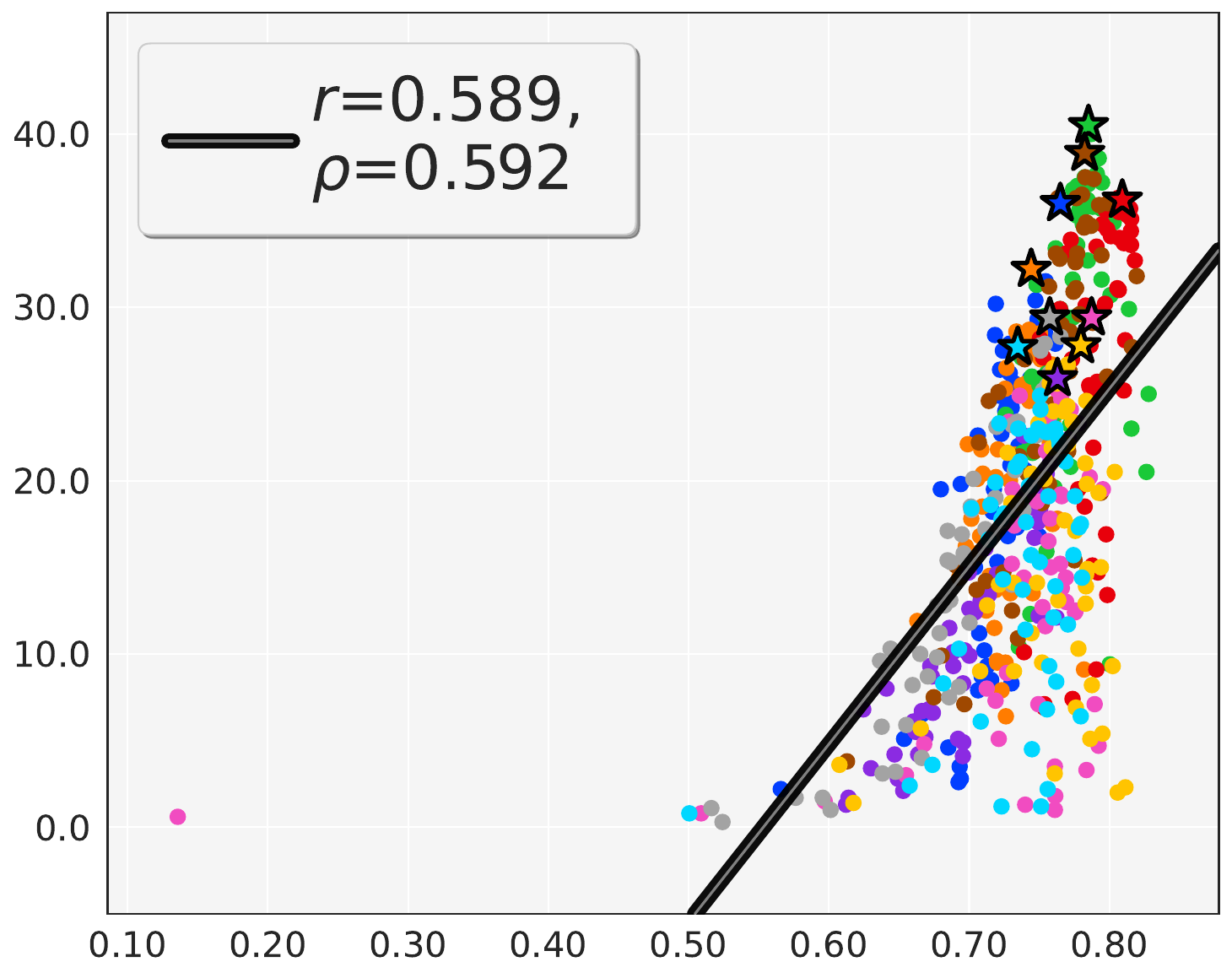}
            \includegraphics[width=0.24\textwidth]{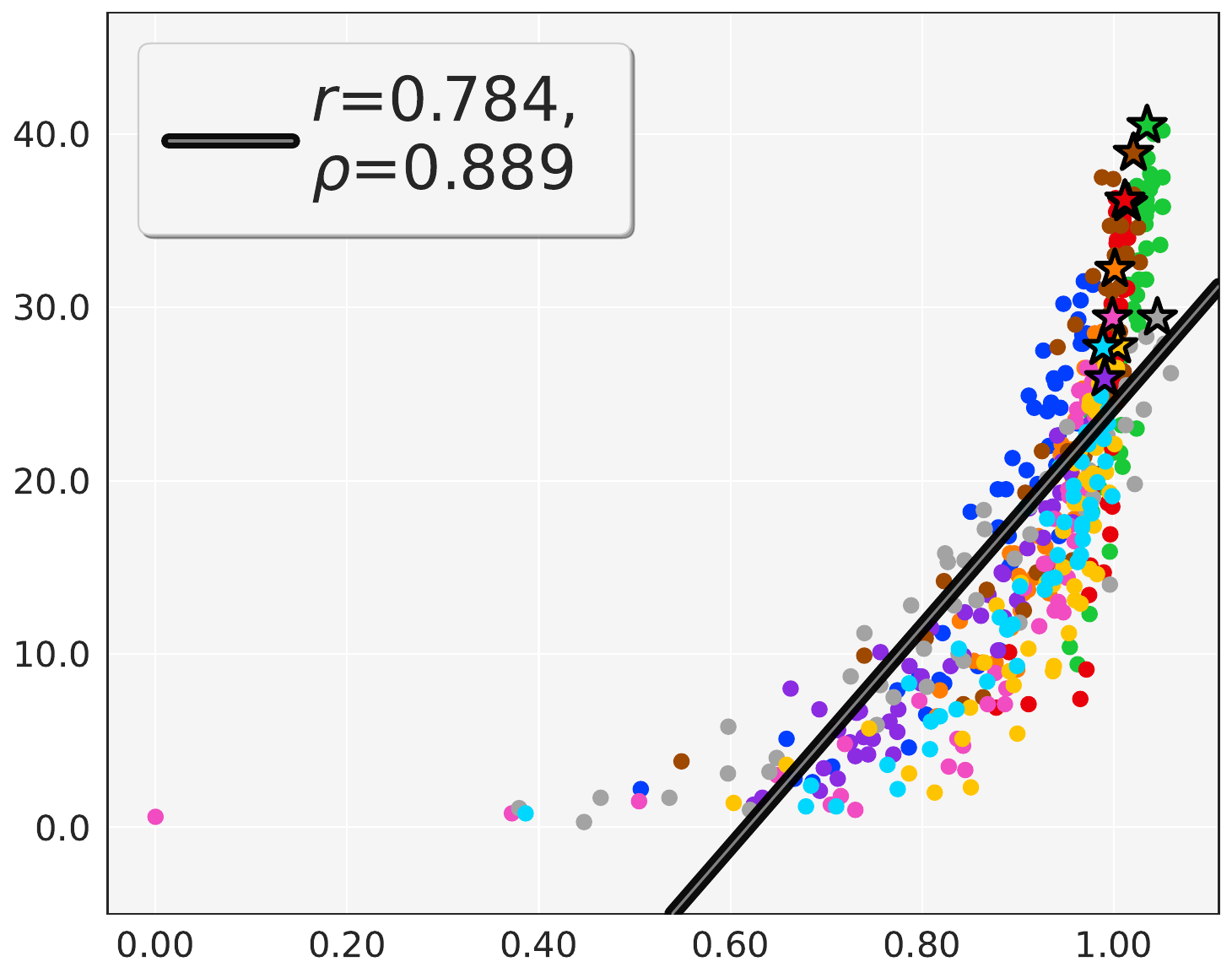}
            \includegraphics[width=0.12\textwidth]{figs/empty.pdf}
            \\
        } 
        \makebox[\textwidth]{
            \includegraphics[width=0.12\textwidth]{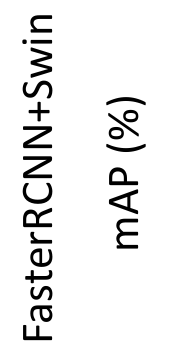}
            \includegraphics[width=0.24\textwidth]{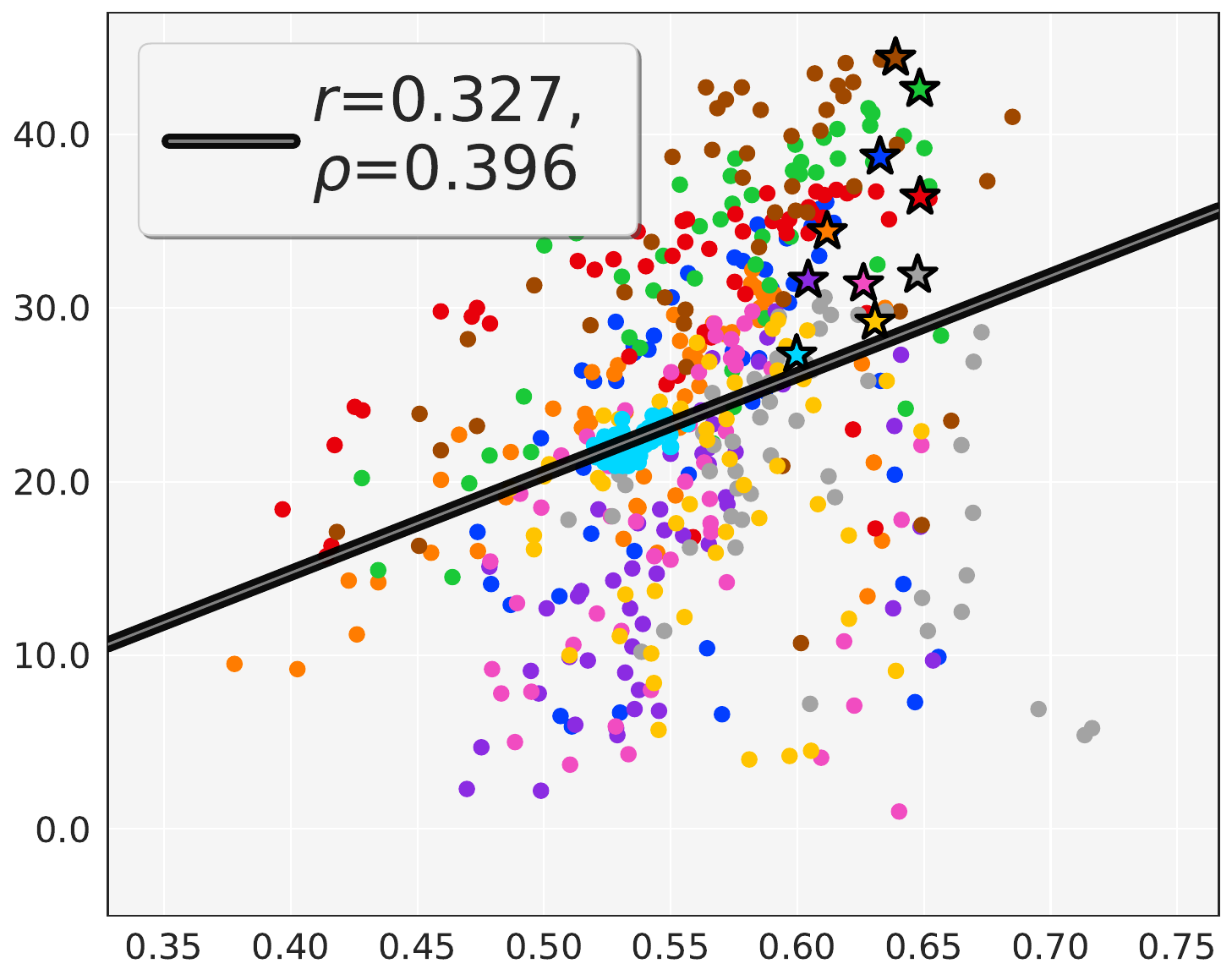}
            \includegraphics[width=0.24\textwidth]{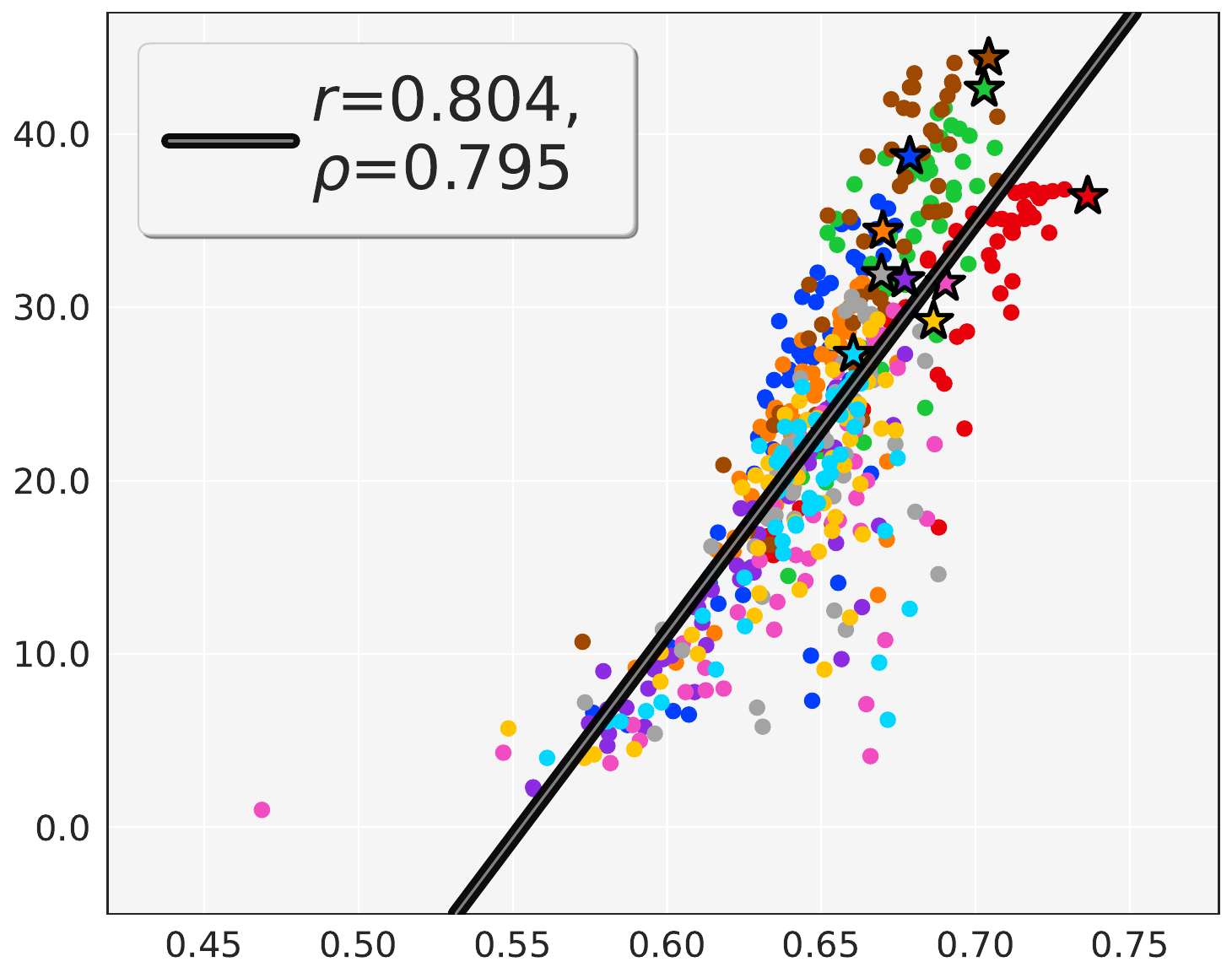}
            \includegraphics[width=0.24\textwidth]{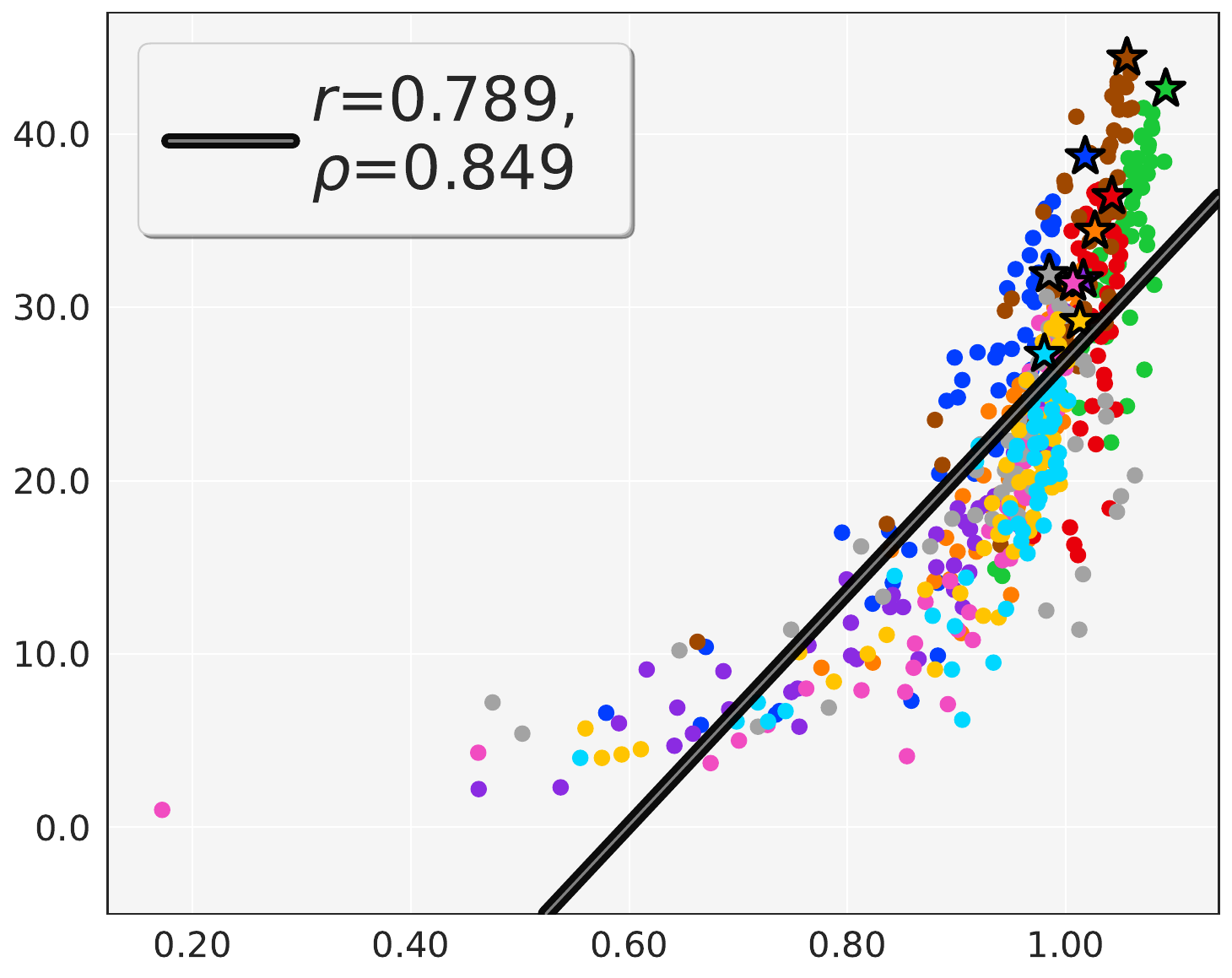}
            \includegraphics[width=0.12\textwidth]{figs/empty.pdf}
            \\
        }
        \hspace*{3cm}\includegraphics[width=0.66\textwidth]{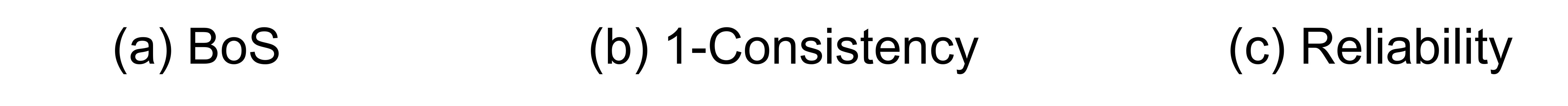}
    \end{minipage}
    \caption{Correlation analysis for vehicle detection, comparing the BoS score~\cite{yang2024bos} and the consistency and reliability scores used in our proposed PCR.
    Test sets (\(\star\)) and datasets in the meta-dataset (\(\circ\)) are indicated by different markers, where each color corresponds to a distinct source dataset.
    The blue line represents the linear regression of mAP on each AutoEval score, where correlation is measured by Pearson's correlation coefficient (\(r\)) and Spearman's rank correlation coefficient (\(\rho\)).}
    \label{fig:vehicle_corr}
\end{figure*}

\begin{figure*}[t!]
    \centering
    \begin{minipage}{\textwidth}
        \makebox[\textwidth]{
            \includegraphics[width=0.12\textwidth]{figs/retina_r50.pdf}
            \includegraphics[width=0.24\textwidth]{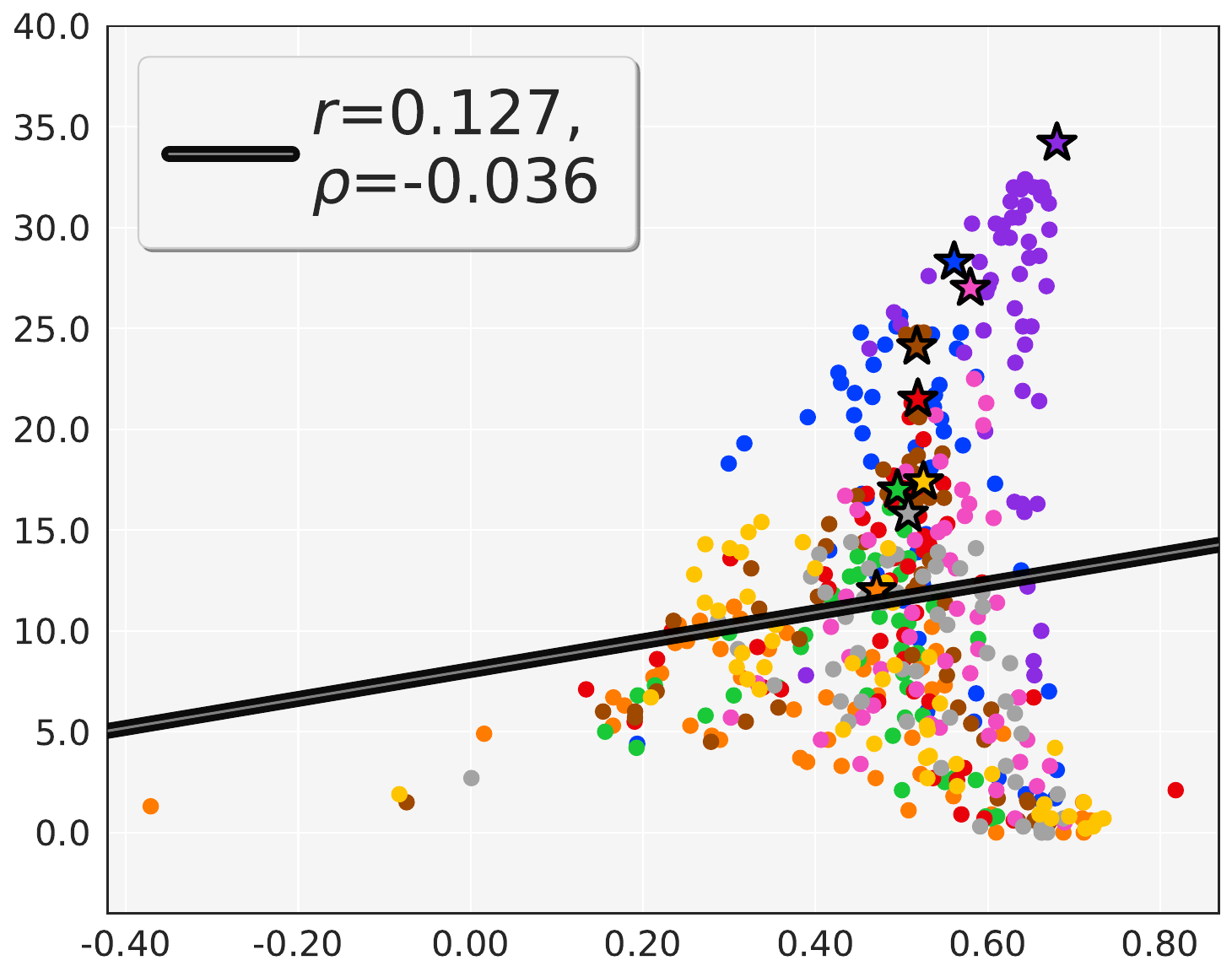}
            \includegraphics[width=0.24\textwidth]{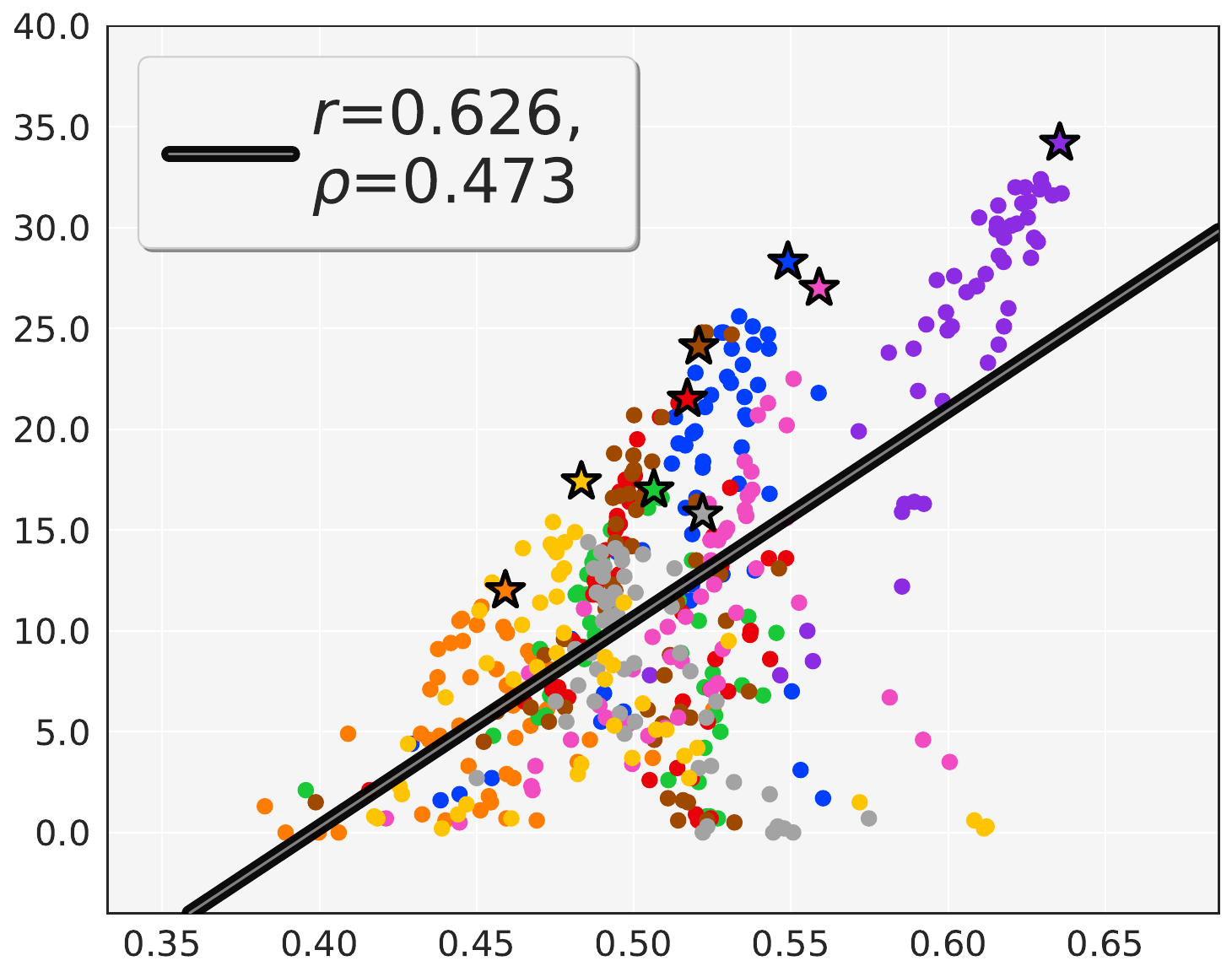}
            \includegraphics[width=0.24\textwidth]{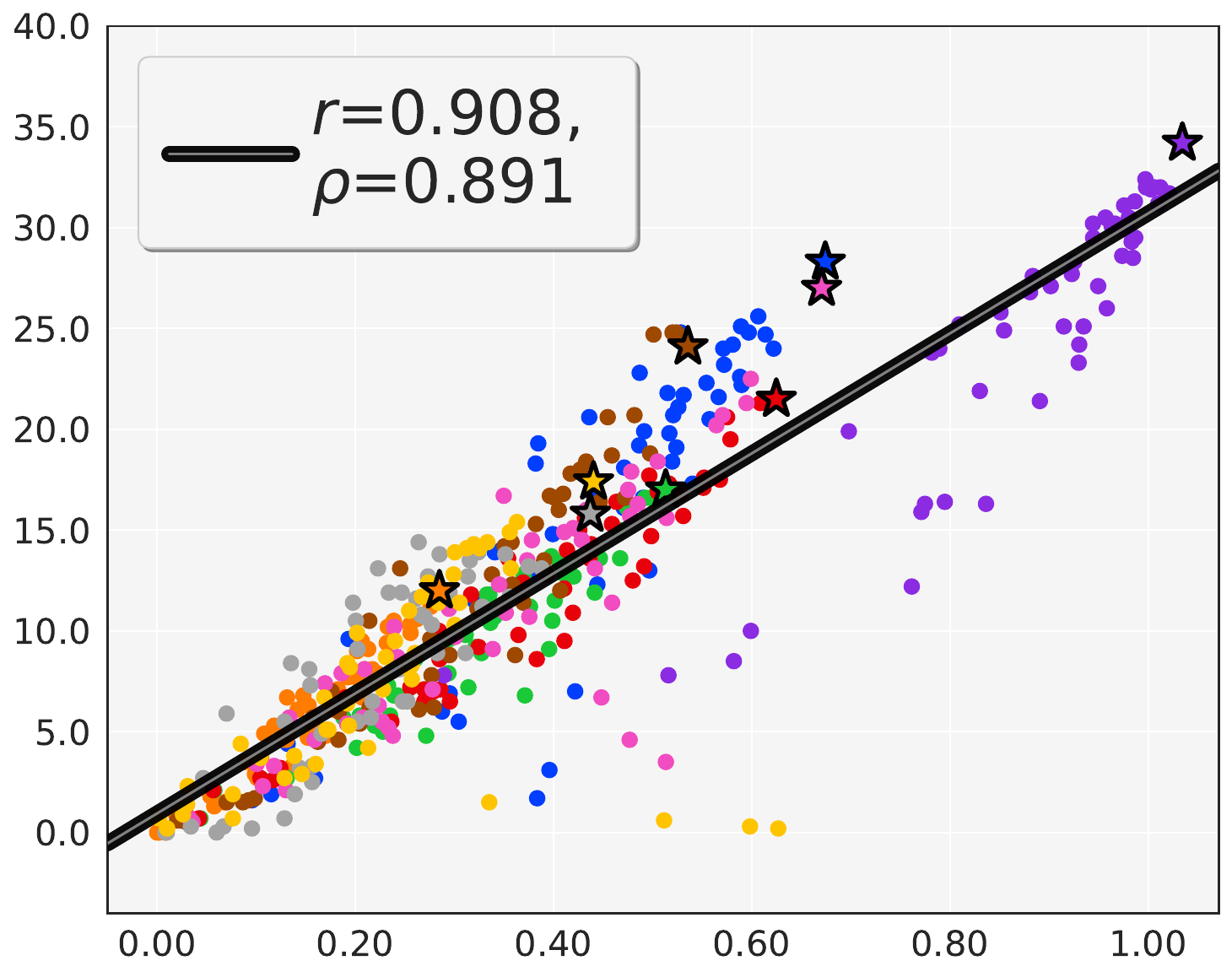}
            \includegraphics[width=0.12\textwidth]{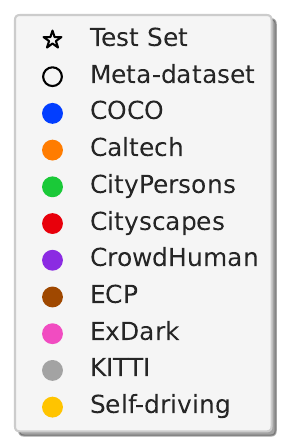}
            \\
        }
        \makebox[\textwidth]{
            \includegraphics[width=0.12\textwidth]{figs/retina_swin.pdf}
            \includegraphics[width=0.24\textwidth]{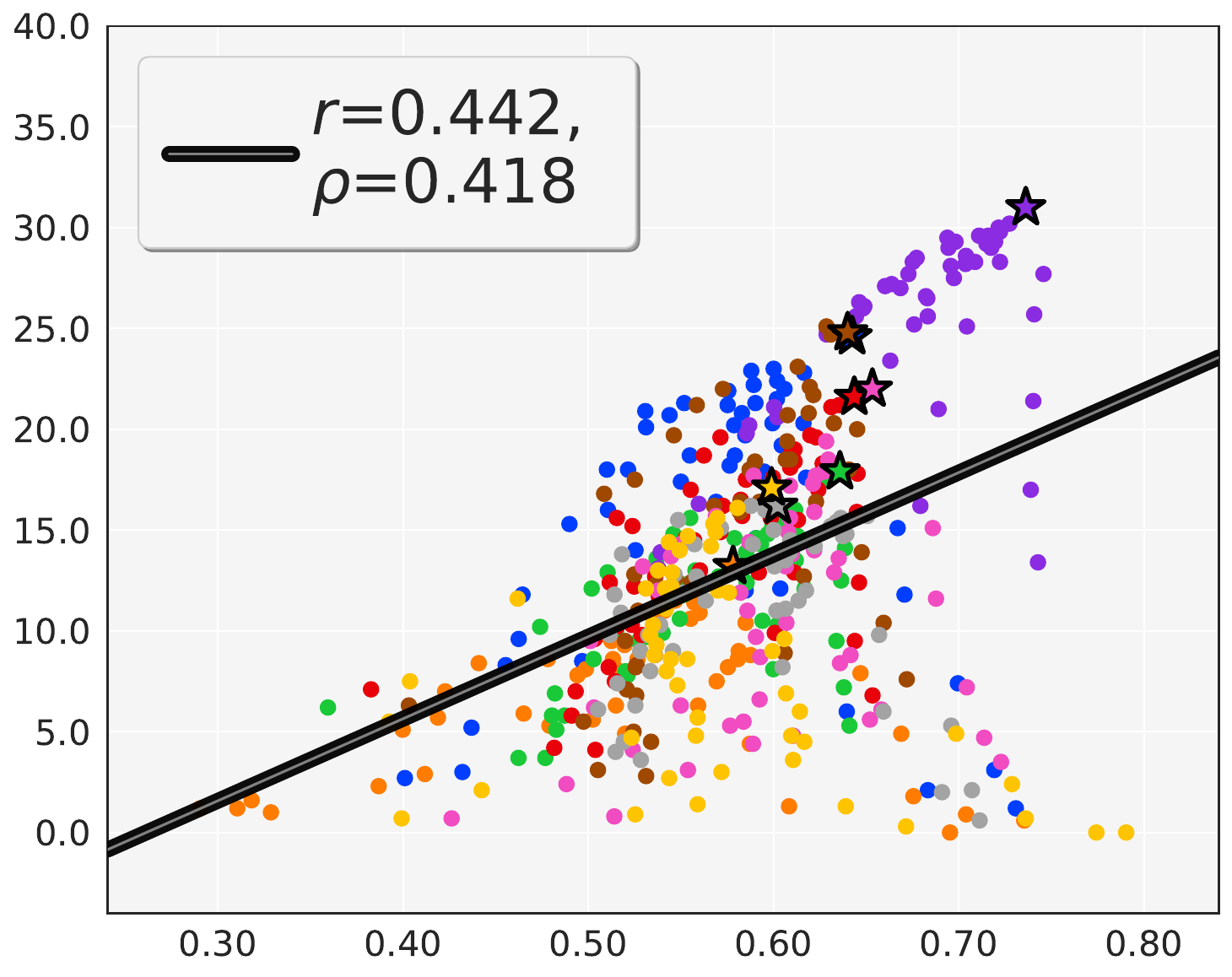}
            \includegraphics[width=0.24\textwidth]{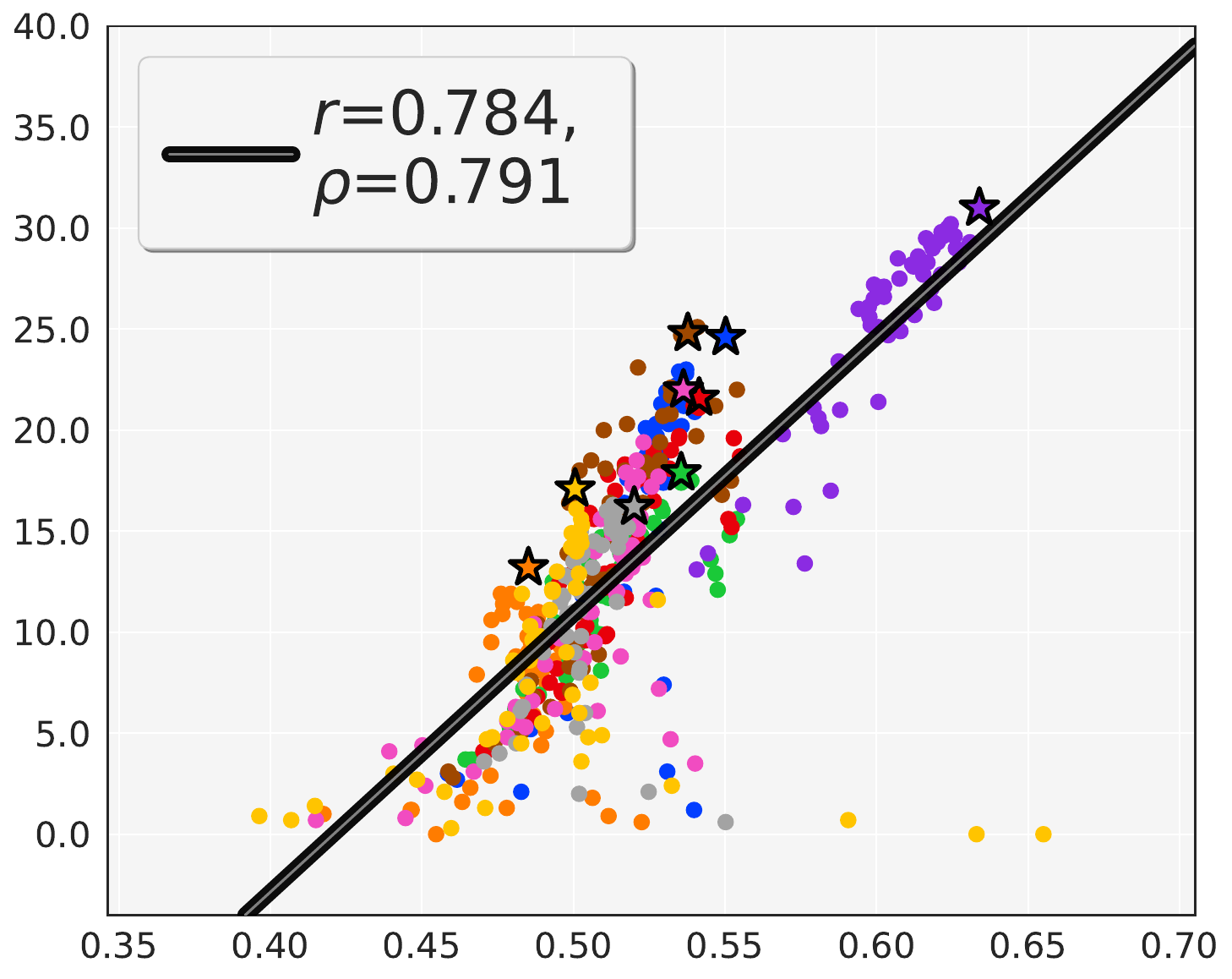}
            \includegraphics[width=0.24\textwidth]{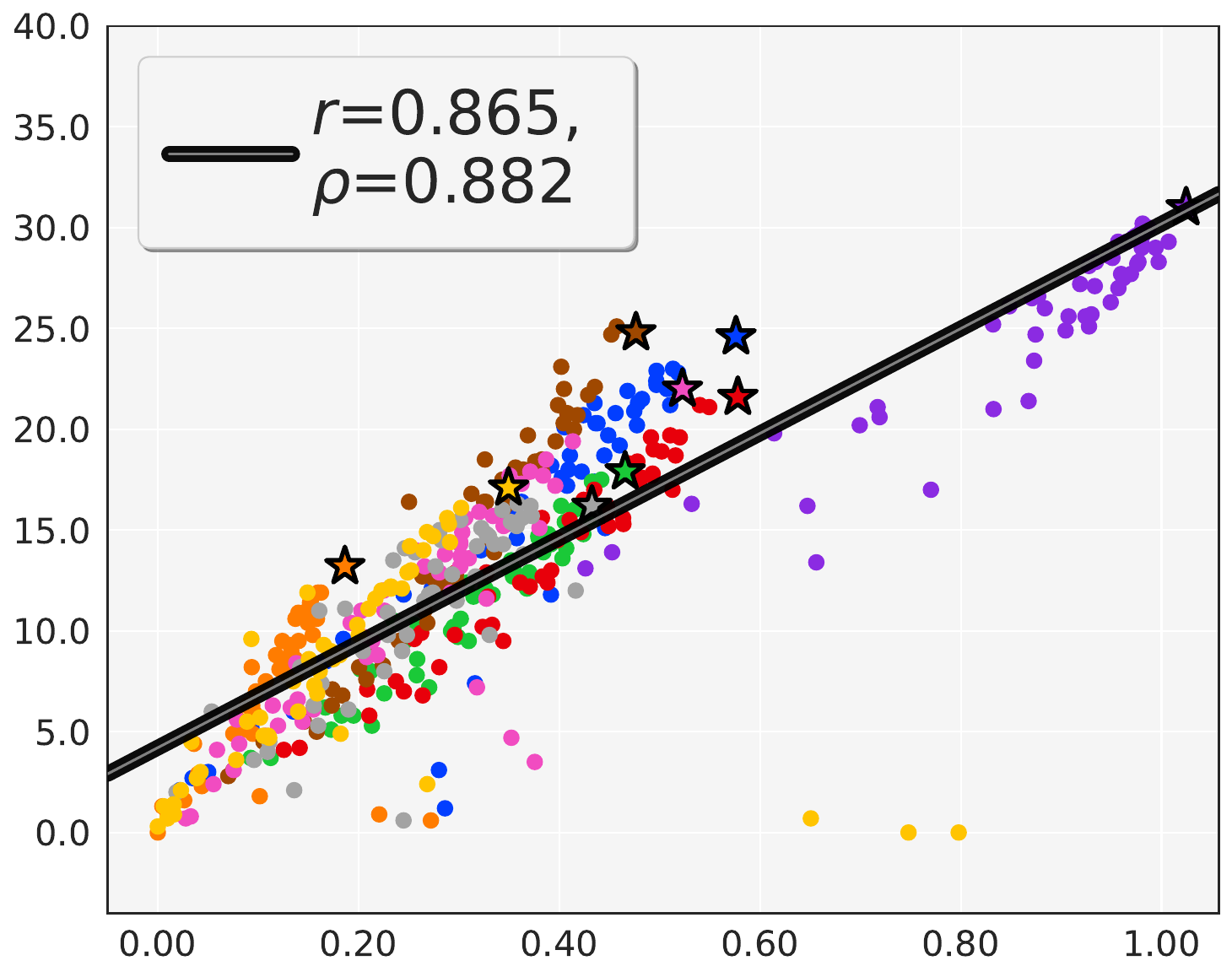}
            \includegraphics[width=0.12\textwidth]{figs/empty.pdf}
            \\
        }
        \makebox[\textwidth]{
            \includegraphics[width=0.12\textwidth]{figs/faster_r50.pdf}
            \includegraphics[width=0.24\textwidth]{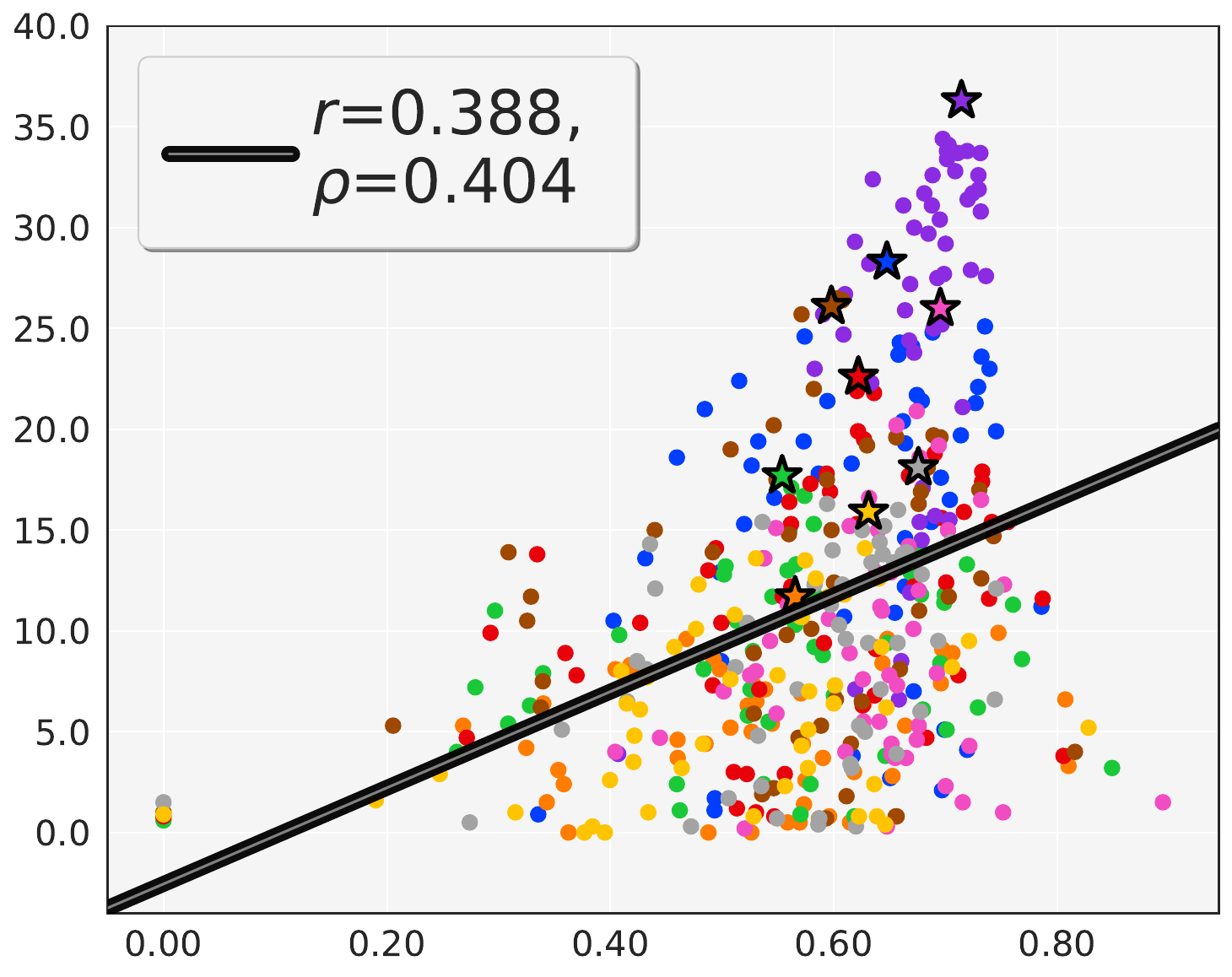}
            \includegraphics[width=0.24\textwidth]{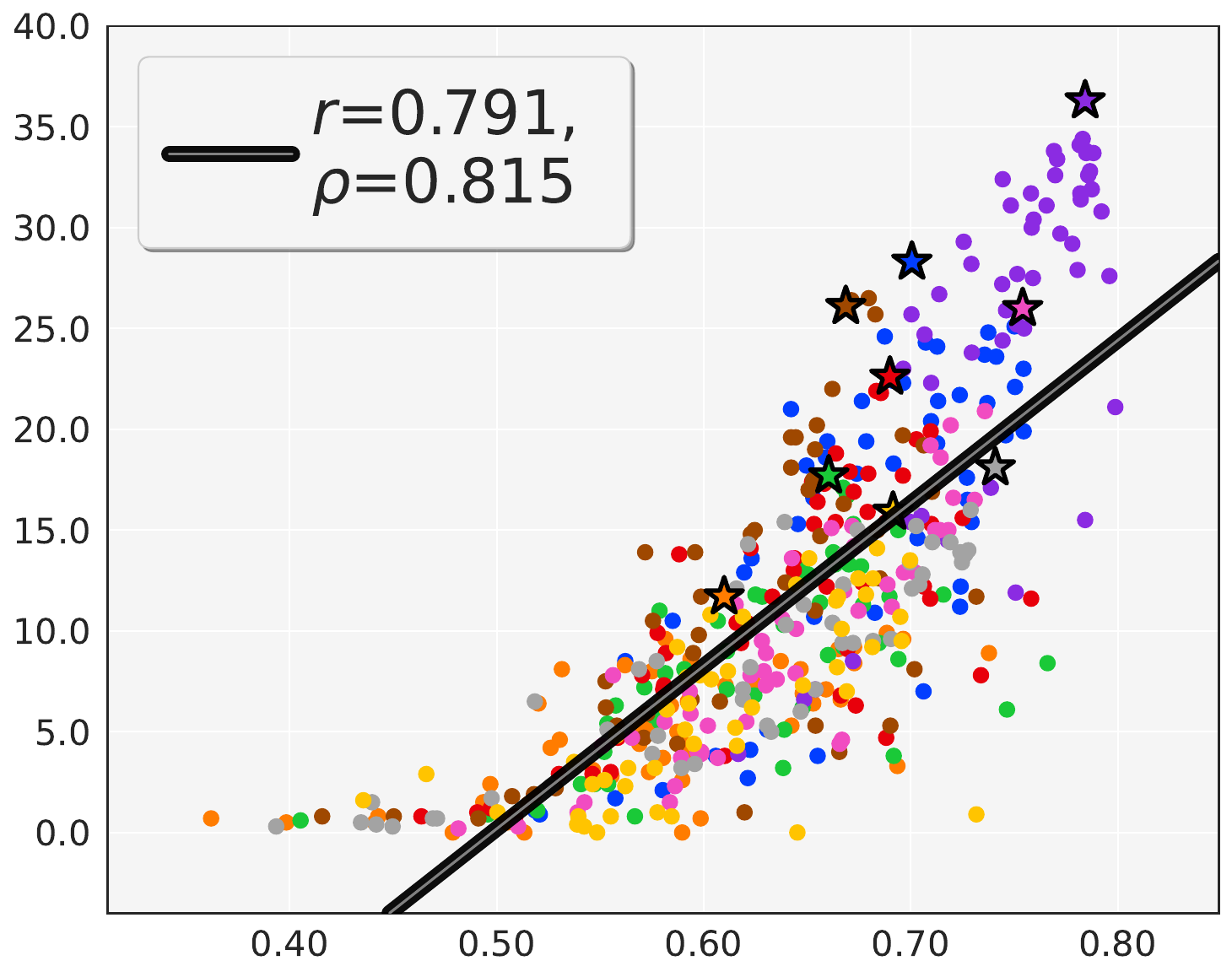}
            \includegraphics[width=0.24\textwidth]{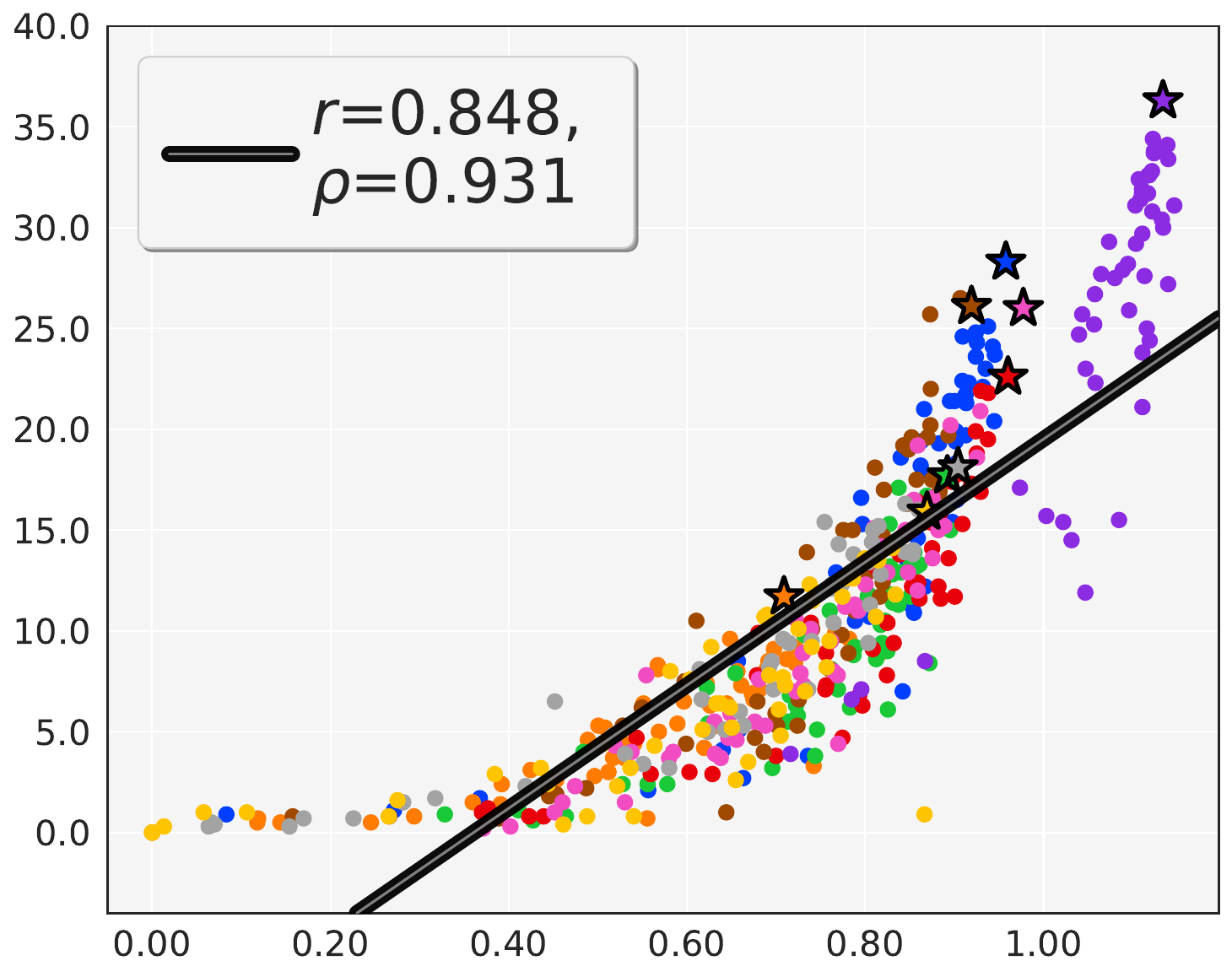}
            \includegraphics[width=0.12\textwidth]{figs/empty.pdf}
            \\
        } 
        \makebox[\textwidth]{
            \includegraphics[width=0.12\textwidth]{figs/faster_swin.pdf}
            \includegraphics[width=0.24\textwidth]{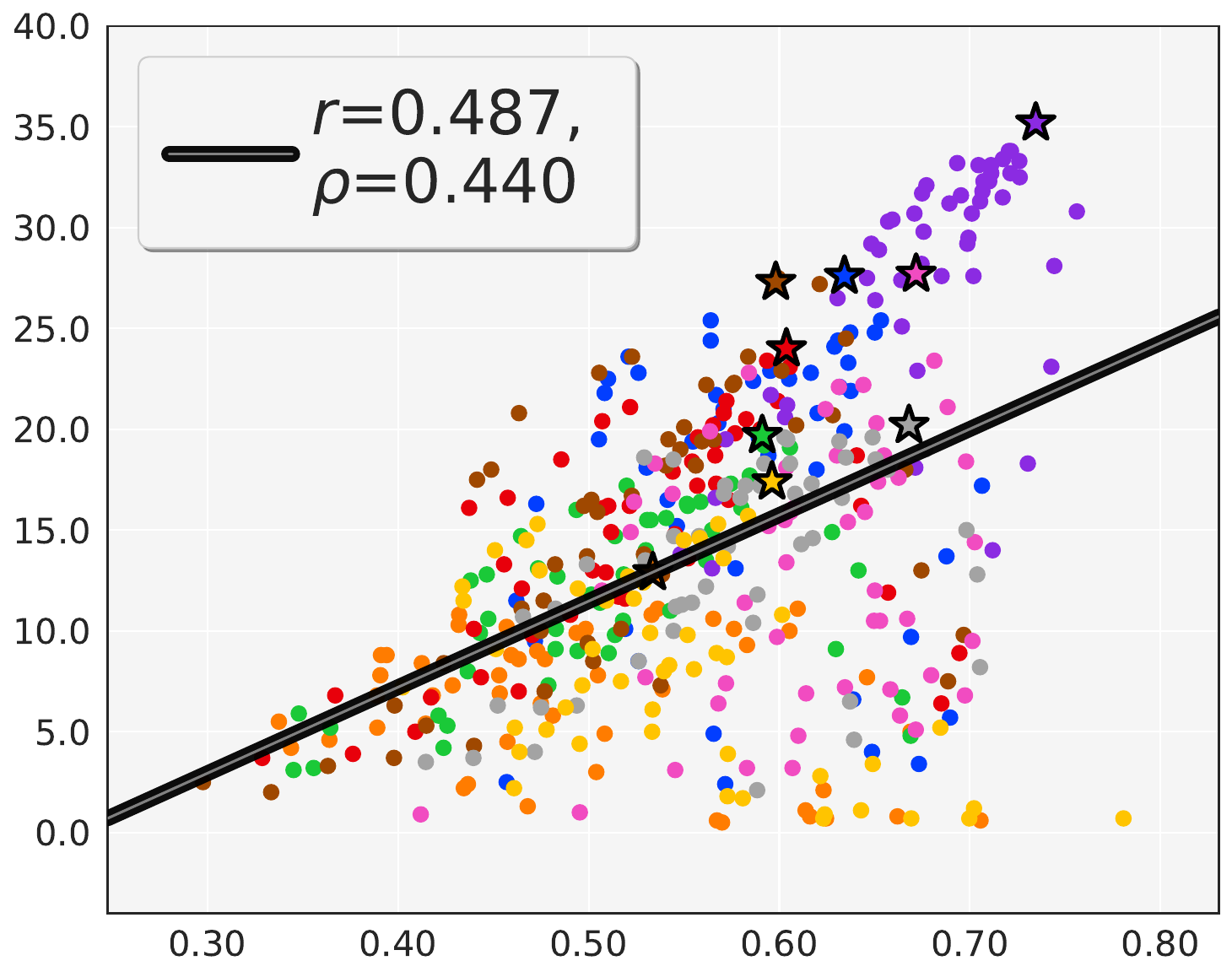}
            \includegraphics[width=0.24\textwidth]{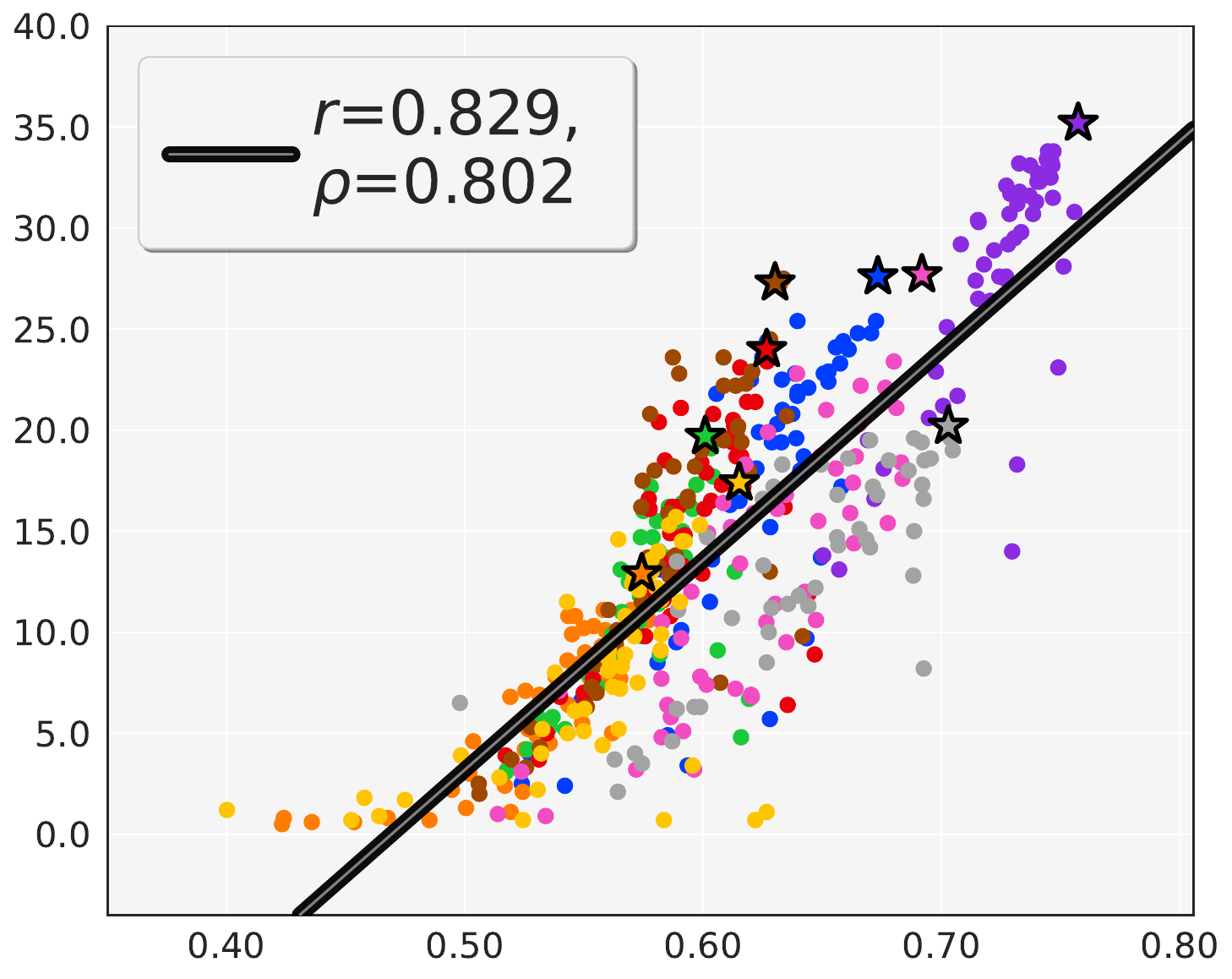}
            \includegraphics[width=0.24\textwidth]{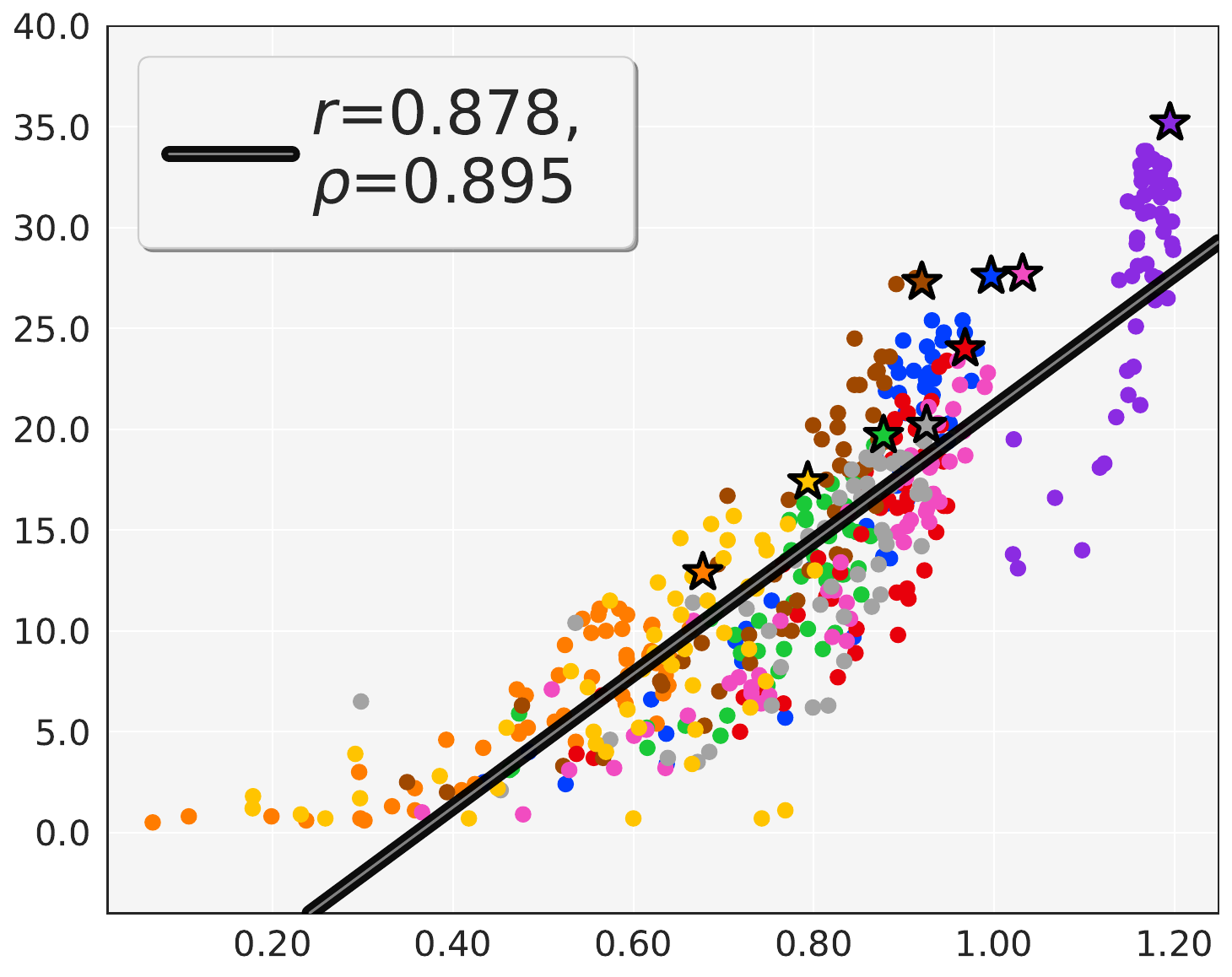}
            \includegraphics[width=0.12\textwidth]{figs/empty.pdf}
            \\
        }
        \hspace*{3cm}\includegraphics[width=0.66\textwidth]{figs/method_names_final.pdf}
    \end{minipage}
    \caption{Correlation analysis for pedestrian detection, comparing the BoS score~\cite{yang2024bos} and the consistency and reliability scores used in our proposed PCR.
    Test sets (\(\star\)) and datasets in the meta-dataset (\(\circ\)) are indicated by different markers, where each color corresponds to a distinct source dataset.
    The blue line represents the linear regression of mAP on each AutoEval score, where correlation is measured by Pearson's correlation coefficient (\(r\)) and Spearman's rank correlation coefficient (\(\rho\)).}
    \label{fig:Pedestrian_corr}
\end{figure*}

\end{document}